\documentclass{article}
\usepackage{amsfonts}
\usepackage{amsmath}
\usepackage{amssymb}

\usepackage{bm}
\usepackage{graphicx}
\usepackage{tabularx}
\usepackage{multirow}
\usepackage[ruled]{algorithm2e}

\newtheorem{problem}{Problem}
\DeclareMathOperator*{\argmin}{arg\,min}

\begin{document}

\title{Automatically Redundant Features Removal for Unsupervised Feature Selection via Sparse Feature Graph}
\author{Shuchu Han\\
shuchu.han@gmail.com\\
Department of Computer Science, \\
Stony Brook University, 
\\Stony Brook, NY 11794, United States\\
Hao Huang\\
haohuanghw@gmail.com\\
Machine Learning Laboratory,\\
 General Electric Global Research, \\
 San Ramon, CA 94853, United States\\
Hong Qin\\
qin@cs.stonybrook.edu\\
Department of Computer Science, \\
Stony Brook University, \\
Stony Brook, NY 11794, United States}


\begin{abstract}
The redundant features existing in high dimensional datasets always affect the performance of learning and mining algorithms. How to detect and remove them is an important research topic in machine learning and data mining research. In this paper, we propose a graph based approach to find and remove those redundant features automatically for high dimensional data. Based on sparse learning based unsupervised feature selection framework, Sparse Feature Graph (SFG) is introduced not only to model the redundancy between two features, but also to disclose the group redundancy between two groups of features. With SFG, we can divide the whole features into different groups, and improve the intrinsic structure of data by removing detected redundant features. With accurate data structure, quality indicator vectors can be obtained to improve the learning performance of existing unsupervised feature selection algorithms such as multi-cluster feature selection (MCFS). Our experimental results on benchmark datasets show that the proposed SFG and feature redundancy remove algorithm can improve the performance of unsupervised feature selection algorithms consistently. 

    
\end{abstract}

\date{\today}
\maketitle

For unsupervised feature selection algorithms, the structure of data is used to generate indication vectors for selecting informative features. The structure of data could be local manifold structure~\cite{he2011variance}~\cite{hou2014joint}, global structure~\cite{liu2014global}~\cite{zhao2013similarity}, discriminative information~\cite{yang2011l2}~\cite{li2012unsupervised} and etc. To model the structure of data, methods like Gaussian similarity graph, or $k$-nearest neighbor similarity graph are very popular in machine learning research. All these similarity graphs are built based on the pairwise distance like Euclidean distance ($\mathcal{L}_2$ norm) or Manhattan distance ($\mathcal{L}_1$ norm) defined between two data samples (vectors). As we can see, the pairwise distance is crucial to the quality of indication vectors, and the success of unsupervised feature selection depends on the accuracy of these indication vectors.

When the dimensional size of data becomes high, or say, for high dimensional datasets, we will meet the curse of high dimensionality issue~\cite{beyer1999nearest}. That means the differentiating ability of pairwise distance will degraded rapidly when the dimension of data goes higher, and the nearest neighbor indexing will give inaccurate results~\cite{weber1998quantitative}~\cite{aggarwal2001surprising}. As a result, the description of data structure by using similarity graphs will be not precise and even wrong. This create an embarrassing \textit{chicken-and-egg} problem~\cite{du2015unsupervised} for unsupervised feature selection algorithms: ``the success of feature selection depends on the quality of indication vectors which are related to the structure of data. But the purpose of feature selection is to giving more accurate data structure.'' 

Most existing unsupervised feature selection algorithms use all \textit{original features}~\cite{du2015unsupervised} to build the similarity graph. As a result, the obtained data structure information will not as accurate as the intrinsic one it should be. To remedy this problem, dimensionality reduction techniques are required. For example, Principal Component Analysis (PCA) and Random Projection (RP) are popular methods in machine learning research. However, most of them will project the data matrix into another (lower dimensional) space with the constraint to approximate the original pairwise similarities. As a result, we lose the physical meaning or original features and the meaning of projected features are unknown.

In this study, we proposed a graph-based approach to reduce the data dimension by removing redundant features. Without lose of generality, we categorize features into three groups~\cite{dash1997feature}: \textit{relevant feature},\textit{irrelevant feature} and \textit{redundant feature}. A feature $\bm{f_i}$ is relevant or irrelevant based on it's correlation with indication vectors (or target vectors named in other articles) $\bm{Y}=\{\bm{y_i}, i \in [1,k]\}$.	For supervised feature selection algorithms~\cite{robnik2003theoretical}~\cite{tibshirani1996regression}~\cite{peng2005feature}, these indication vectors usually relate to class labels. For unsupervised scenario~\cite{dy2004feature}~\cite{cai2010unsupervised}, as we mentioned early, they follow the structure of data. Redundant features are features that highly correlated to other features, and have no contribution or trivial contribution to the target learning task. The formal definition of redundant feature is by~\cite{yu2004efficient} based on the Markov blanket given by~\cite{koller1996toward}.

Based on the philosophy of sparse learning based MCFS algorithm, a feature could be redundant to another single feature, or to a subset of features. In this work, we propose a graph based approach to identify these two kind of redundancy at the same time. The first step is to build a Sparse Feature Graph (SFG) at feature side based on sparse representation concept from subspace clustering theory~\cite{elhamifar2013sparse}. Secondly, we review the quality of sparse representation of each single feature vector and filtered out those failed ones. In the last, we defined Local Compressible Subgraphs (LCS) to represent those local feature groups that are very redundant. Moreover, a greedy local search algorithm is designed to discover all those LCSs. Once we have all LCSs, we pick the feature which has the highest node in-degree as the representative feature and treat all other as redundant features. With this approach, we obtain a new data matrix with reduced size and alleviate the curse of dimensional issues.

To be specific, the contribution of our study can be highlighted as:
\begin{itemize}
	\item We propose sparse feature graph to model the feature redundancy existing in high dimensional datasets. The sparse feature graph inherits the philosophy of sparse learning based unsupervised feature selection framework. The sparse feature graph not only records the redundancy between two features but also show the redundancy between one feature and a subset of features.
	\item We propose local compressible subgraph to represent redundant feature groups. And also design a local greedy search algorithm to find all those subgraphs.
	\item We reduce the dimensionality of input data and alleviate the curse of dimensional issue through redundant features removal. With a more accurate data structure, the \textit{chicken-and-egg} problem for unsupervised feature selection algorithms are remedied in certain level. One elegant part of our proposed approach is to reduce the data dimension without any pairwise distance calculation.
	\item Abundant experiments and analysis over twelve high dimensional datasets from three different domains are also presented in this study. The experiment results show that our method can obtain better data structure with reduced size of dimensionality, and proof the effectiveness of our proposed approach.
\end{itemize}

The rest of paper is organized as follows. The first section describe the math notation used in our work. The Section 2 introduces the background , motivation and preliminaries of our problem. In Section 3, we define the problem we are going to solve. In Section 4, we present our proposed sparse feature graph algorithm and discuss the sparse representation error problem. We also introduce the local compressible subgraph and related algorithm. The experiment results are reported in Section 5, and a briefly reviewing of related works is given in Section 6. Finally, we conclude our study in last Section 7.
\section{Math Notation}
Throughout this paper, matrices are written as boldface capital letters and vectors are represented as boldface lowercase letters.
Let the data matrix be represented as $\bm{X}\in\mathbb{R}^{n\times d}$, while each row is a sample (or instance), and each column means a feature. If we view the data matrix $\bm{X} = \bm{[x_1,x_2,\cdots,x_n]}^T, \bm{x_i} \in \mathbb{R}^{d\times 1}$ from feature side, it can be seen as $\bm{F} = \bm{X}^T = \bm{[f_1,f_2,\cdots,f_d]}, \bm{f_i}\in \mathbb{R}^{n\times1} (1\leq i\leq d)$. 

\section{Background and Preliminaries}
\subsection{Unsupervised Feature Selection}
\begin{figure}
	\centering
	\includegraphics[width=0.5\linewidth,bb=0 3 643 201]{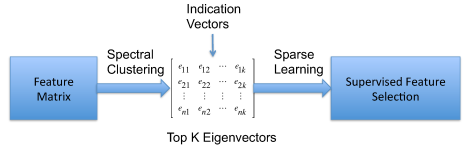}
	\caption{The framework of sparse learning based unsupervised feature selection.}
	\label{fig:17tkdd_unsupervised_fs}
\end{figure}
In unsupervised feature selection framework, we don't have label information to determine the feature relevance. Instead, the data similarity or manifold structure constructed from the whole feature space are used as criteria to select features. Among all those algorithms of unsupervised feature selection, the most famous one is MCFS. The MCFS algorithm is a sparse learning based unsupervised feature selection method which can be illustrated as figure~\ref{fig:17tkdd_unsupervised_fs}. 
The core idea of MCFS is to use the eigenvectors of graph Lapalcian over similarity graph as indication vectors. And then find set of features that can approximate these eigenvectors through sparse linear regression. Let us assume the input data has number $K$ clusters that is known beforehand (or an estimated $K$ value by the expert's domain knowledge). The top $K$ non-trivial eigenvectors, $\bm{Y} = [\bm{y_1},\cdots,\bm{y_k}]$, form the spectral embedding $\bm{Y}$ of the data. Each row of $\bm{Y}$ is the new coordinate in the embedding space. To select the relevant features, MCFS solves $K$ sparse linear regression problems between $\bm{F}$ and $\bm{Y}$ as:
\begin{equation}
\min\limits_{\alpha_i} \| \bm{y}_i - \bm{F}\bm{\alpha}_i \|^2 + \beta \|\bm{\alpha}_i\|_1,
\end{equation} 
where $\bm{\alpha}_i$ is a $n$-dimensional vector and it contains the combination coefficients for different features $\bm{f}_i$ in approximating $\bm{y}_i$. Once all coefficients $\bm{\alpha}_i$ are collected, features will be ranked by the absolute value of these coefficients and top features are selected. This can be show by a weighted directed bipartite graph as following:
\begin{figure}[h!]
	\centering
	\includegraphics[width=0.4\linewidth,bb=2 0 350 330]{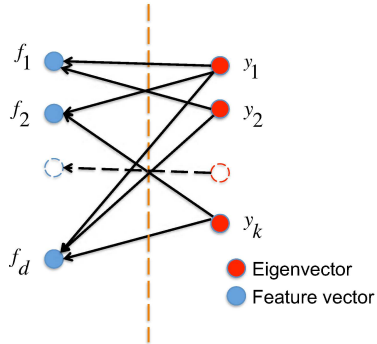}
	\caption{Sparse learning bipartite graph for MCFS.}
	\label{fig::17tkdd_mcfs_bipartite}
\end{figure}

\subsection{Adaptive Structure Learning for High Dimensional Data}
As we can seen, the MCFS uses whole features to model the structure of data. That means the similarity graph such as Gaussian similarity graph is built from all features. This is problematic when the dimension of data vector goes higher. To be specific, the pairwise distance between any two data vectors becomes almost the same, and as a consequence of that, the obtained structural information of data is not accuracy. This observation is the motivation of unsupervised Feature Selection with Adaptive Structure Learning (FSASL) algorithm which is proposed by Du et al.~\cite{du2015unsupervised}. The idea of FSASL is to repeat MCFS iteratively with updating selected feature sets. It can be illustrated as following:
\begin{figure}
	\centering
	\includegraphics[width=0.4\linewidth,bb=0 0 643 315]{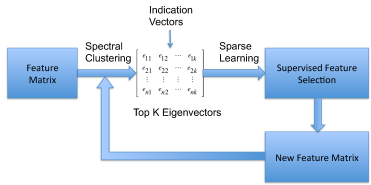}
	\caption{Unsupervised Feature Selection with Adaptive Structure Learning.}
	\label{fig:17tkdd_fsasl}
\end{figure}
FASAL is an iterative algorithms which keeps pruning irrelevant and noisy features to obtain better manifold structure while improved structural info can help to search better relevant features. FASAL shows better performance in normalized mutual information and accuracy than MCFS generally. However, it's very time consuming since it is an iterative algorithm includes many eigen-decompositions.

\subsection{Redundant Features}
For high dimensional data $\bm{X} \in \mathbb{R}^{n \times d}$, it exists information redundancy among features since $d \ll n$. Those redundant features can not provide further performance improvement for ongoing learning task. Instead, they impair the efficiency of learning algorithm to find intrinsic data structure. 

In this section, we describe our definition of feature redundancy. Unlike the feature redundancy defined bt Markov blanket~\cite{yu2004efficient} which is popular in existing research works, our definition of feature redundancy is based on the linear correlation between two vectors (the ``vector'' we used here could be a feature vector or a linear combination of several feature vectors.) To measure the redundancy between two vectors $\bm{f}_i$ and $\bm{f}_j$, squared cosine similarity\cite{xufeature} is used:
\begin{equation}
R_{ij} = cos^2(\bm{f}_i,\bm{f}_j).
\end{equation}
By the math definition of cosine similarity, it is straightforward to know that a higher value of $R_{i,j}$ means high redundancy existing between $\bm{f}_i$ and $\bm{f}_j$. For example, feature vector $\bm{f}_i$ and its duplication $\bm{f}_i$ will have $R_{ii}$ value equals to one. And two orthogonal feature vectors will have redundancy value equals to zero. 

\section{Problem Statement}
In this work, our goal is to detect those redundant features existing in high dimensional data and obtain a more accurate intrinsic data structure. To be specific: 
\begin{problem}
	Given a high dimensional data represented in the form of feature matrix $\bm{X}$, how to remove those redundant features $f_{(\cdot)} \in \bm{X}^T$ for unsupervised feature selection algorithms such as MCFS?
\end{problem}

Technically, the MCFS algorithm does not involve redundant features. However, the performance of MCFS depends on the quality of indication vectors which are used to select features via sparse learning. And those indication vectors are highly related to the intrinsic structure of data which is described by the selected features and given distance metric. For example, the MCFS algorithm uses all features and Gaussian similarity to represent the intrinsic structure. This is the discussed `chicken-and-egg" problem~\cite{du2015unsupervised} between structure characterization and feature selection. The redundant and noise features will lead to an inaccurate estimation of data structure. As a result, it's very demanding to remove those redundant (and noise) features before the calculation of indication vectors.

\section{Algorithm}
In this section, we present our graph-based algorithm to detect and remove  redundant features existing in high dimensional data. First, the sparse feature graph that modeling the redundancy among feature vectors will be introduced. Secondly, the sparse representation error will be discussed. In the last, the local compressible subgraph is proposed to extract redundant feature groups.

\subsection{Sparse Feature Graph (SFG)}
The most popular way to model the redundancy among feature vectors is correlation such as Pearson Correlation Coefficient (PCC). The correlation value is defined over two feature vectors, and it's a pairwise measurement. However, there also exiting redundancy between one feature vector and a set of feature vectors according to the philosophy of MCFS algorithm. In this section, we present SFG, which model the redundancy not only between two feature vectors but also one feature vector and a set of feature vectors.  

The basic idea of sparse feature graph is to looking for a sparse linear representation for each feature vector while using all other feature vectors as dictionary. For each feature vector $\bm{f}_i$ in features set $\bm{F} = \bm{[f_1,f_2,\cdots,f_d]}$, SFG solves the following optimization problem:
\begin{equation}
\label{eqn:sfg}
\min\limits_{\bm{\alpha}\in\mathbb{R}^{d-1}}\|\bm{f_i} - \bm{\Phi^i\alpha_i}\|_2^2,\quad\mathrm{s.t.}~\|\bm{\alpha_i}\|_{0} < L,
\end{equation}
where $\bm{\Phi}^i = \bm{[f_1,f_2,\cdots,f_{i-1},f_{i+1},\cdots,f_d]}$ is the dictionary of $f_i$ and each column of $\bm{\Phi}^i$ is a selected feature from data matrix $\bm{X}$. $L$ is a constraint to limit the number of nonzero coefficients. In SFG, we set it to the number of features $d$. The $\bm{\alpha}_i$ is the coefficient of each atom of dictionary $\bm{\Phi}^i$. This coefficient vector not only decides the edge link to $\bm{f_i}$ but also indicates the weight of that connection. The resulted SFG is a weighted directed graph and may have multiple components. 
\begin{figure}[h!]
	\centering
	\includegraphics[width=0.5\linewidth,bb=0 0 459 389]{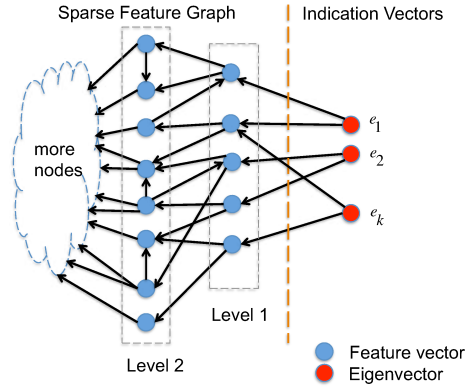}
	\caption{Sparse feature graph and its relation with indication vectors. The level 1 features are direct sparse representation of those calculated indication vectors. The level 2 features only have representation relationship with level 1 features but not with indication vectors.}
	\label{fig::tkdd_sfg}
\end{figure}

To solve the optimization problem~\ref{eqn:sfg}, we use Orthogonal Matching Pursuit (OMP) solver~\cite{you2016scalable} here since the number of features in our datasets is larger than 1,000. We modify the stop criterion of OMP by checking the value change of residual instead of residual itself or the maximum number of supports. The reason is that we want the number of supports (or say, the number of edge connections) to follow the raw data property. Real world datasets are always noisy and messy. It's highly possible that several feature vectors may fail to find a correct sparse linear representation through OMP. If we set residual or maximum of supports as criteria, we can not differentiate the successful representations and the failed ones.

The OMP solver and SFG algorithm can be described as following.
\begin{algorithm}
	\SetAlgoLined
	\SetKwInOut{Input}{Input}\SetKwInOut{Output}{Output}
	\Input{$\bm{\Phi}=[\bm{f_1,f_2,\cdots,f_{i-1},f_{i+1}, \cdots, f_d}] \in \mathbb{R}^{n\times (d-1)}, \bm{f_i} \in \mathbb{R}^{n},\epsilon$.}
	\Output{Coefficient $\bm{\alpha}_i$.}
	Initialize residual difference threshold $r_0 = 1.0$, residual $\bm{q}_0 = \bm{f}_i$, support set $\Gamma_0 = \emptyset$, $k = 1$ \;
	\While{$ k \leq d-1$ and $ |r_k - r_{k-1}| > \epsilon$}{
		Search the atom which most reduces the objective:\\
			$\quad j^* = \argmin\limits_{j\in\Gamma^C}\left\{\min\limits_{\alpha}\|\bm{f}_i - \bm{\Phi}_{\Gamma\cup\{j\}}\bm{\alpha}\|_2^2 \right\}$\;
		Update the active set:\\ $\quad\Gamma_k = \Gamma_{k-1} \cup \{j^*\}$\;
		Update the residual (orthogonal projection):\\
		$\quad\bm{q}_k = (I - \bm{\Phi}_{\Gamma_k}(\bm{\Phi}_{\Gamma_k}^T\bm{\Phi}_{\Gamma_k})^{-1}\bm{\Phi}_{\Gamma_k}^T)\bm{f}_i$\;
		Update the coefficients:\\
		$\quad \bm{\alpha}_{\Gamma_k} = (\bm{\Phi}_{\Gamma_k}^T\bm{\Phi}_{\Gamma_k})^{-1}\bm{\Phi}_{\Gamma_k}^T\bm{f}_i$\;
		$r_k = \|\bm{q}_k\|_2^2$\;
		$k \leftarrow k + 1$\;
}
	\caption{Orthogonal Matching Pursuit (OMP)}
	\label{alg:tkdd_omp}
\end{algorithm}%
\hfill
 \begin{algorithm}
 	\SetAlgoLined
 	\SetKwInOut{Input}{Input}\SetKwInOut{Output}{Output}
 	\Input{Data matrix $\bm{F} = [\bm{f_1,f_2,\cdots,f_d}] \in \mathbb{R}^{n\times d}$;}
 	\Output{Adjacent matrix $\bm{W}$ of Graph $\bm{G} \in \mathbb{R}^{d\times d}$;}
 	\BlankLine
 	Normalize each feature vector $\bm{f_i}$ with $\bm{\|f_i\|_2^2 = 1}$\;
 	\For{$i = 1,\cdots,d$}{
	 	Compute $\bm{\alpha}_i$ from OMP($\bm{F}_{-i}$,$\bm{f}_i$) using algorithm~\ref{alg:tkdd_omp}\;
 	}
	Set adjacent matrix $W_{ij} = \bm{\alpha_i(j)}$ if $i>j$, $W_{ij} = \bm{\alpha_i(j-1)}$, if $i<j$ and $W_{ij}=0$ if $i==j$\;
 	\BlankLine
 	\caption{Sparse Feature Graph}
 \end{algorithm}

\subsection{Sparse Representation Error} 
In our modified OMP algorithm~\ref{alg:tkdd_omp}, we set a new stop criterion of searching sparse representation solution for each feature vector $\bm{f}_i$. Instead of keep searching until arriving a minimization error, we stop running while the solver could not reduce the length of residual vector anymore. To be specific, the 2-norm of residual vector is monitored and the solver will stop once the change of this value small than a user specified threshold. 

The reason we use this new stop criterion is that several feature vectors may not find correct sparse representation in current dataset, and the ordinary OMP solver will return a meaningless sparse representation when the maximum iteration threshold arrived. Since the goal of SFG is not to find a correct sparse representation for every feature vectors, we utilize the new stop criterion and add a filter process in our algorithm to identify those failed sparse representation.

To identify those failed sparse representation, we check the angle between the original vector and the linear combination of its sparse representation. In the language of SFG, we check the angle between a node (a feature vector) and the weighted combination of its one-ring neighbor. Only the neighbors of  out edges will be considered. This can be illustrated by following figure~\ref{fig:tkdd_sparse_rep_err}.
\begin{figure}
	\centering
	\includegraphics[width=0.4\textwidth,bb=1 0 323 242]{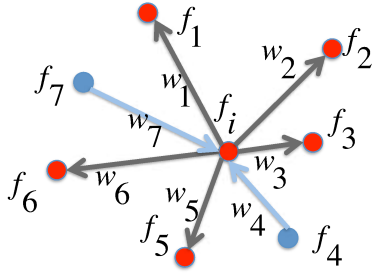}
	\caption{Illustration of sparse representation error. SFG is a weighted directed graph.}
	\label{fig:tkdd_sparse_rep_err}
\end{figure}
As the example in Figure~\ref{fig:tkdd_sparse_rep_err}, node $\bm{f}_i$ has seven one-ring neighbors. But only $\\bm{f}_1,bm{f}_2,\bm{f}_3,\bm{f}_5,\bm{f}_6$ are its sparse representation and $\bm{f}_4$ and $\bm{f}_7$ are not. Then the sparse representation error $\zeta$ is calculated by:
\begin{align*}
\bm{f}_i^*& = w_1\bm{f}_1+w_2\bm{f}_2+w_3\bm{f}_3+w_5\bm{f}_5+w_6\bm{f}_6,\\
\zeta &= \arccos (\bm{f}_i, \bm{f}_i^*).
\end{align*}
Once we have the SFG, we calculate the sparse representation errors for all nodes. A sparse representation is treated as fail if the angle $\zeta$ less than a user specified value. We will filter out these node which has failed representation by removing its out-edges.

\subsection{Local Compressible Subgraph}
We group high correlated features through \textit{local compressible subgraphs}. The SFG $\bm{G}$ is a weighted directed graph. With this graph, we need to find all feature subsets that has very high redundancy. To archive this goal, we propose a local search algorithm with seed nodes to group those highly correlated features into many subgraphs which are named as \textit{local compressible subgraphs} in this study. Our local search algorithm involves two steps, the first step is to sort all nodes by the in-degree. By the definition of SFG, the node with higher in-degree means it appears more frequently in other nodes' sparse representation. The second step is a local bread-first search approach which finds all nodes that has higher weight connections (in and out) to the growing subgraph. The detail subgraph searching algorithm can be described by:
\begin{algorithm}
	\SetAlgoLined
	\SetKwInOut{Input}{Input}\SetKwInOut{Output}{Output}
	\Input{Weighted directed graph $G=(V,E)$, edge weight threshold $\theta$;}
	\Output{Local compressible subgraphs $C$ .}
	\BlankLine
	Tag all nodes with initial label $0$\;
	Sort the nodes by its in-degree decreasingly\;
	$current\_label=1$\; 
	\For{$n = 1:|V|$}{
		\If{$label(n)\  != 0$}{continue;}
		set label of node $n$ to $current\_label$\;
		$BFS(n,\theta,current\_label)$\;
		$current\_label\  += 1$\;
	}
	\tcc{$current\_label$ now has the maximum value of labels.}
	\For{$i = 1:current\_label$}{
		Extract subgraph $c_i$ which all nodes have label $i$\;
		\If{$|c_i| > 1$}{
		add $c_i$ to $C$\;
		}
	}	
	\caption{Local Compressible Subgraphs.}
	\label{alg:tkdd_lcs}
\end{algorithm}
In Alg.~\ref{alg:tkdd_lcs}, function $label(n)$ check the current label of node $n$, and $BFS(n,\theta,current\_label)$ function runs a local Breadth-First search for subgraph that has edge weight large than $\theta$. 

\subsection{Redundant Feature Removal}
The last step of our algorithm is to remove the redundant features. For each local compressible subgraph we found, we pick up the node which has the highest in-degree as the representative node of that local compressible subgraph. So the number of final feature vectors equals to the number of local compressible subgraphs.

\section{Experiments}
In this section, we present experimental results to demonstrate the effectiveness of our proposed algorithm. We first evaluate the spectral clustering performance before and after applying our algorithms. Secondly, we show the performance of MCFS with or without our algorithm. In the last, the properties of generated sparse graphs and sensitivity of parameters are discussed.  

\subsection{Experiment Setup}
\paragraph{Datasets.} We select twelve real-world high dimensional datasets~\cite{Li-etal16} from three different domains: Image, Text and Biomedical. The detail of each dataset is listed in Table~\ref{tab:datasets}. The datasets have sample size different from 96 to 8293 and feature size ranging from 1,024 to 18,933. Also, the datasets have class labels from 2 to 64. The purpose of this selection is to let the evaluation results be more general by applying datasets with various characteristics.
\begin{table}[h!]
	\centering
    \begin{tabular}{|l|c|c|c|l|}
    \hline
    Name & \#Sample & \#Feature & \#Class & Type \\
    \hline\hline
	ORL & 400 & 1024 & 40 & Image \\ \hline
	Yale & 165 & 1024 & 15 & Image \\ \hline
	PIE10P & 210 & 2420 & 10 & Image \\ \hline
	ORL10P & 100 & 10304& 10 & Image \\ \hline
	BASEHOCK & 1993 & 4862 & 2 & Text \\ \hline
	RELATHE & 1427 & 4322 & 2 & Text \\ \hline
	PCMAC & 1943 & 3289 & 2 & Text \\ \hline
	Reuters & 8293 & 18933 & 65 & Text \\ \hline
	lymphoma & 96 & 4026 & 9 & Biomedical \\ \hline
    LUNG & 203 & 3312 & 5 & Biomedical\\ \hline
    Carcinom & 174 & 9182 &  11 & Biomedical\\ \hline
    CLL-SUB-111 & 111 & 11340 & 3 & Biomedical\\ 
    \hline    
    \end{tabular}
    \caption{High dimensional datasets.}
    \label{tab:datasets}
\end{table}

\paragraph{Normalization.} The features of each dataset are normalized to have unit length, which means $\|\bm{f_i}\|_2=1$ for all datasets. 

\paragraph{Evaluation Metric.} Our proposed algorithm is under the framework of unsupervised learning. Without loss of generality, the cluster structure of data is used for evaluation. To be specific, we measure the  spectral clustering performance with Normalized Mutual Information (NMI) and Accuracy (ACC). NMI value ranges from $0.0$ to $1.0$, with higher value means better clustering performance. ACC is another metric to evaluate the clustering performance by measuring the fraction of its clustering result that are correct. Similar to NMI, its values range from $0$ to $1$ and higher value indicates better algorithm performance.

Suppose $A$ is the clustering result and $B$ is the known sample label vector. Let $p(a)$ and $p(b)$ denote the marginal probability mass function of $A$ and $B$, and let $p(a,b)$ be the joint probability mass function of $A$ and $B$. Suppose $H(A), H(B)$ and $H(A,B)$ denote the entropy of $p(a),p(b)$ and $p(a,b)$ respectively. Then the normalized mutual information NMI is defined as:
\begin{equation}
NMI(A,B) = \frac{H(A)+H(B)-H(A,B)}{max(H(A),H(B))}
\end{equation}

Assume $A$ is the clustering result label vector, and $B$ is the known ground truth label vector, ACC is defined as:
\begin{equation}
ACC = \frac{\sum\limits_{i=1}^{N}\delta (B(i),Map_{(A,B)}(i))}{N}
\end{equation}
where $N$ denotes the length of label vector, $\delta(a,b)$ equals to 1 if only if $a$ and $b$ are equal. $Map_{A,B}$ is the best mapping function that permutes $A$ to match $B$. 

\subsection{Effectiveness of Redundant Features Removal} 
Our proposed algorithm removes many features to reduce the dimension size of all data vectors. As a consequence, the pairwise Euclidean distance is changed and the cluster structure will be affected. To measure the effectiveness of our proposed algorithm, we check the spectral clustering performance before and after redundant feature removal. If the NMI and ACC values are not changed to much and stay in the same level, the experiment results show that our proposed algorithm is correct and effective.  

The spectral clustering algorithm we used in our experiments is the Ng-Jordan-Weiss (NJW) algorithm~\cite{ng2001spectral}. The Gaussian similarity graph is applied here as the input and parameter $\sigma$ is set to the mean value of pairwise Euclidean distance among all vectors. 

Our proposed LCS algorithm includes a parameter $\theta$ which is the threshold of redundancy. It decides the number of redundant features implicitly, and affects the cluster structure of data consequently. In our experiment design, we test different $\theta$ values ranging from $90\%$ to $10\%$ with step size equal to $10\%$: $\theta = [0.9,0.8,0.7,\cdots,0.1]$.

\begin{figure}
	\centering
\includegraphics[width=0.225\textwidth]{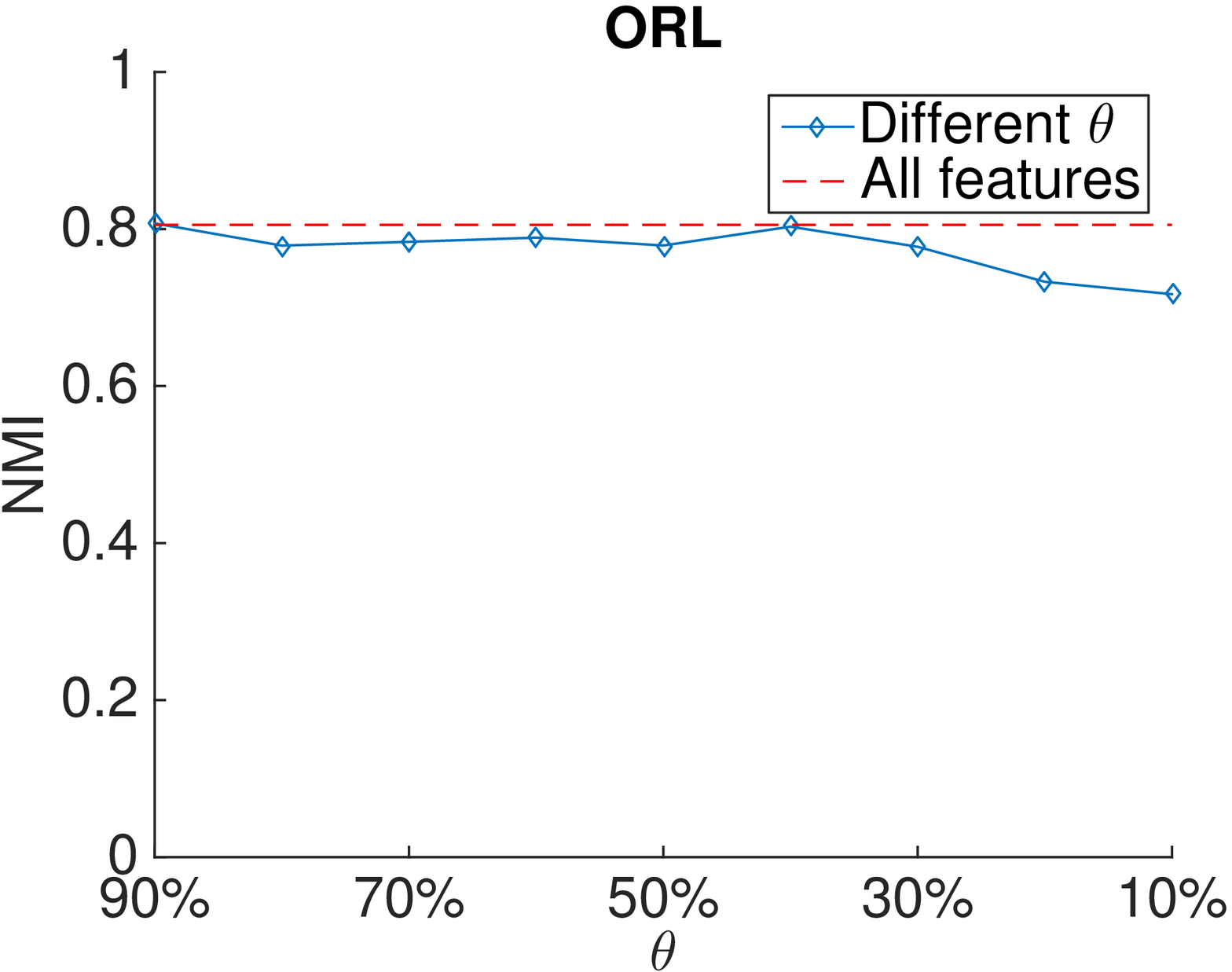} 	
\includegraphics[width=0.225\textwidth]{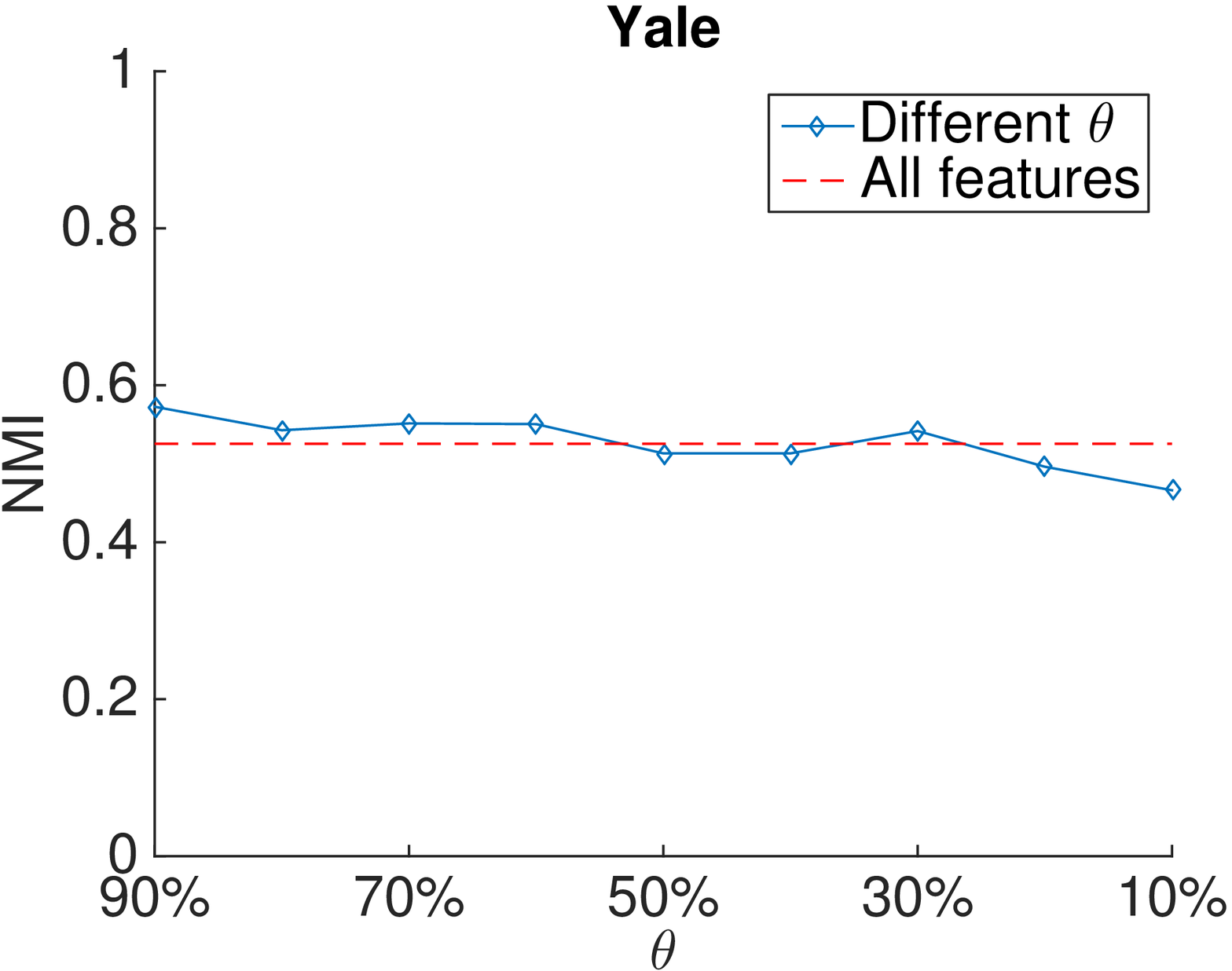} 
\includegraphics[width=0.225\textwidth]{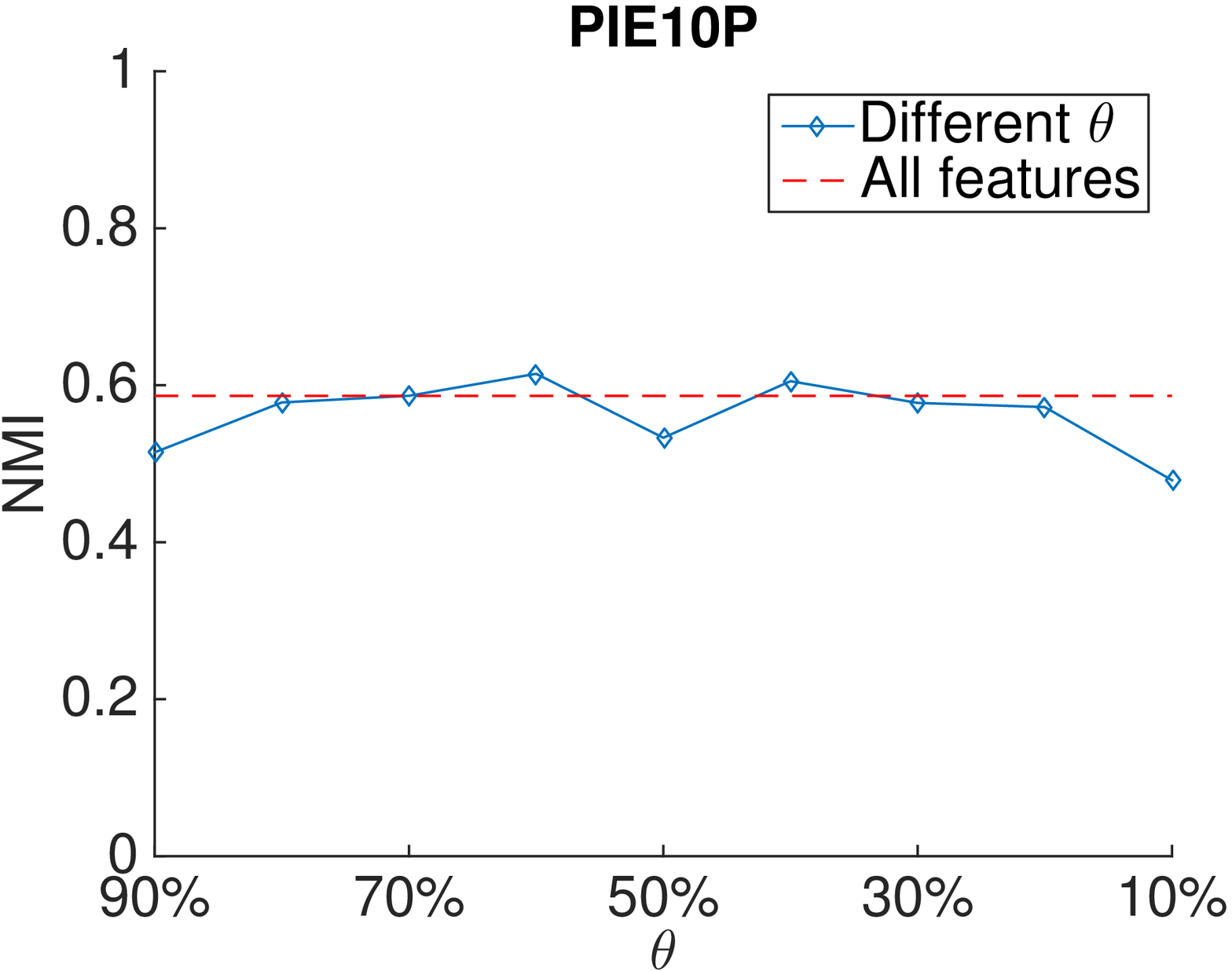}
\includegraphics[width=0.225\textwidth]{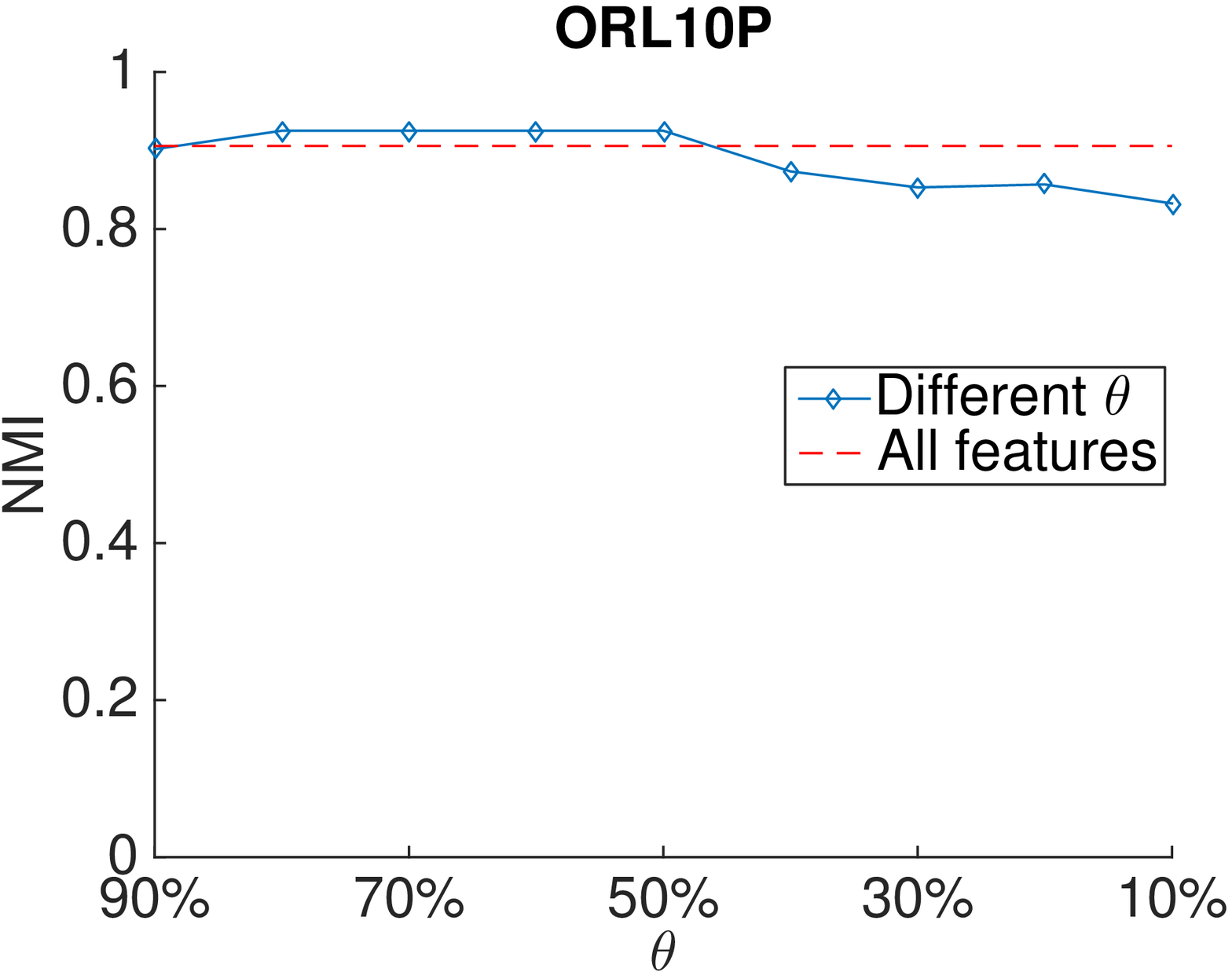} \\
\includegraphics[width=0.225\textwidth]{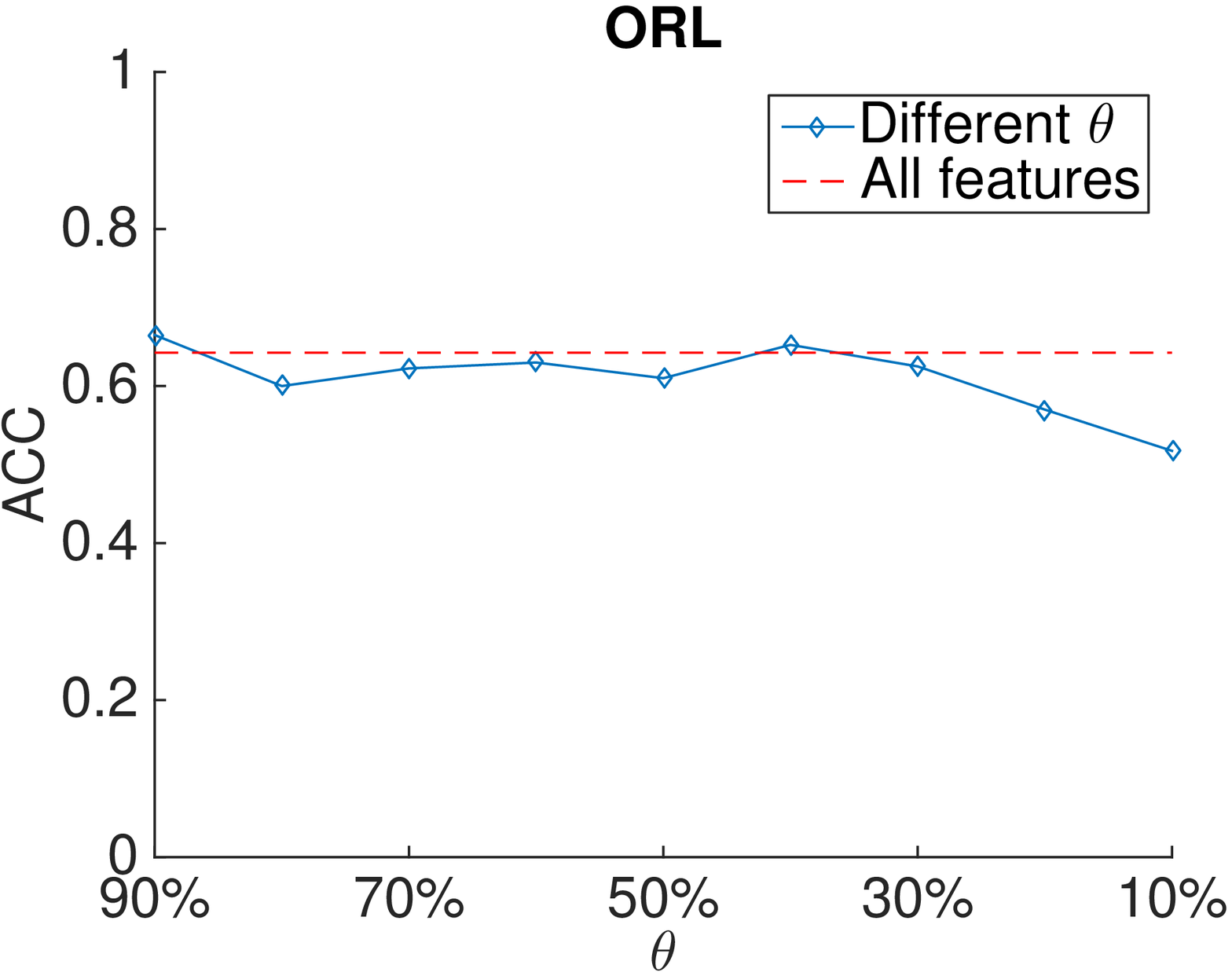} 	
\includegraphics[width=0.225\textwidth]{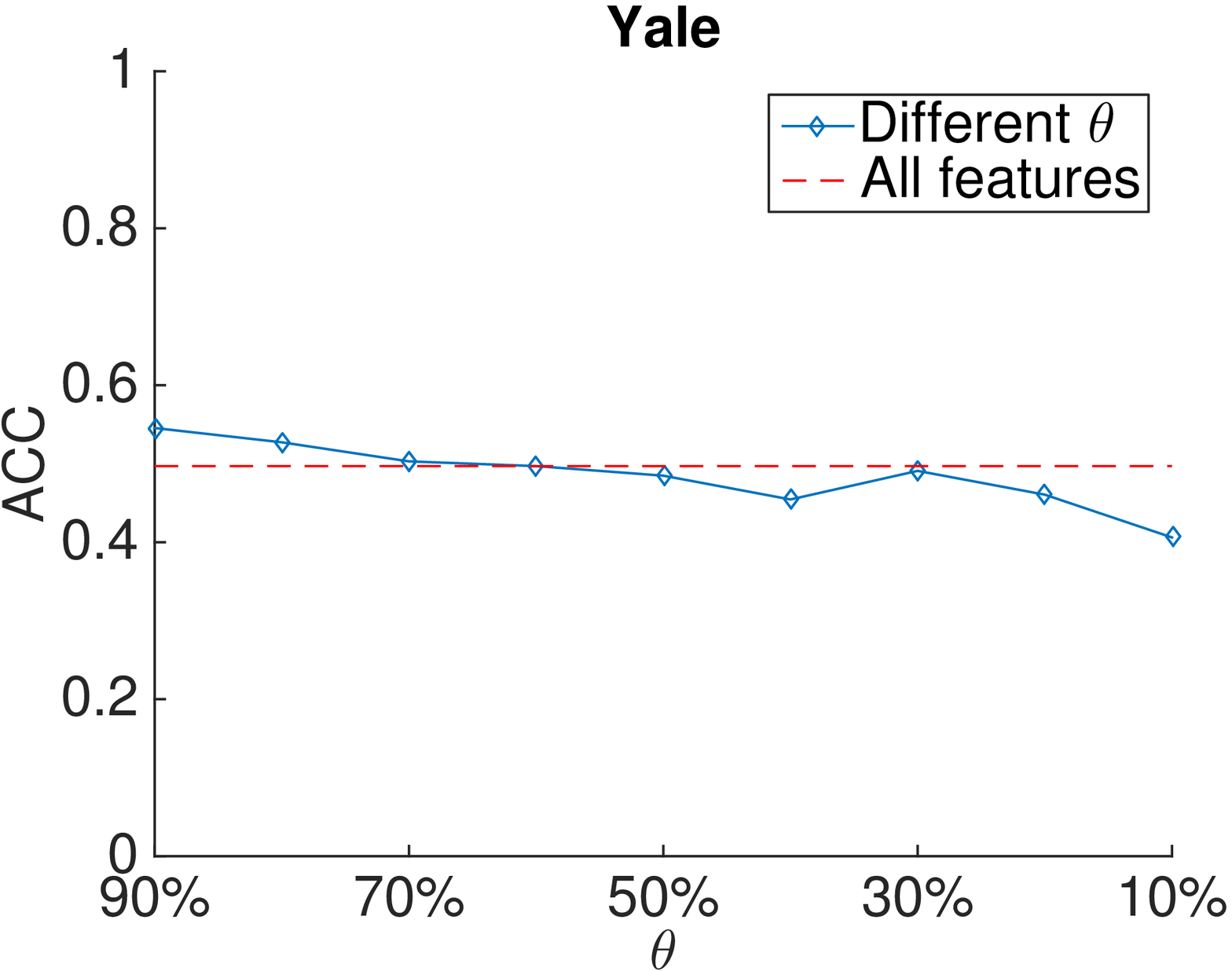} 
\includegraphics[width=0.225\textwidth]{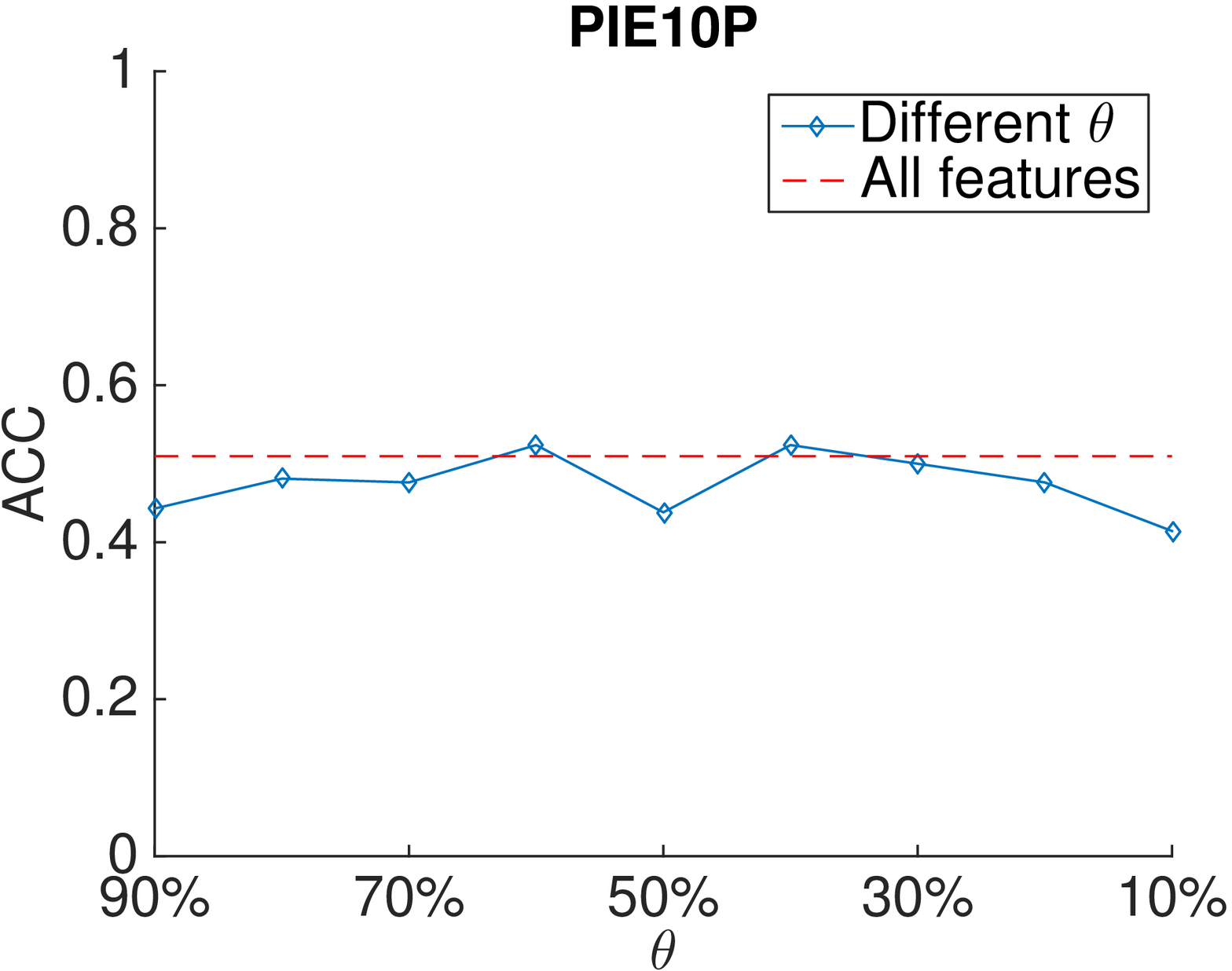}
\includegraphics[width=0.225\textwidth]{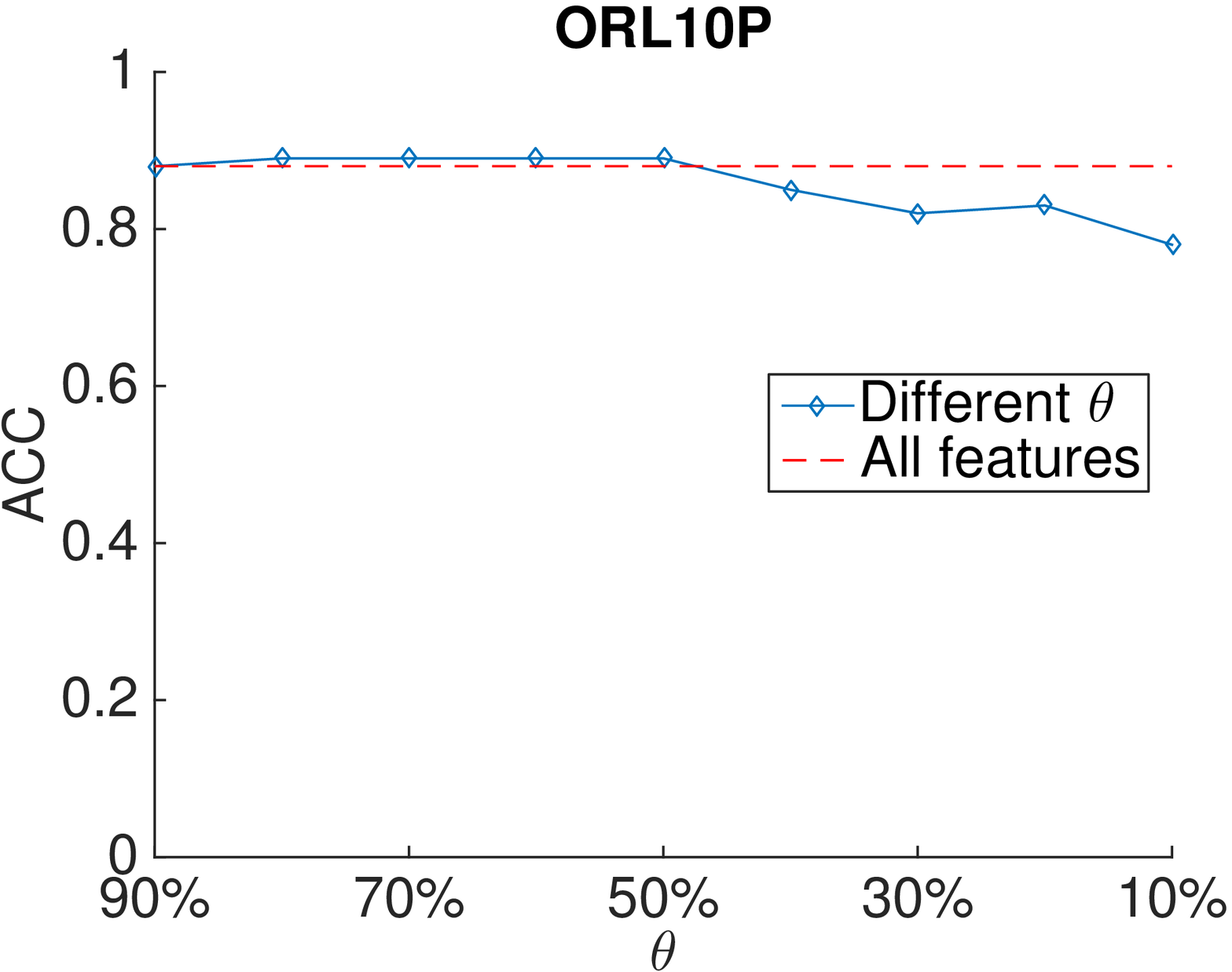} \\
\includegraphics[width=0.225\textwidth]{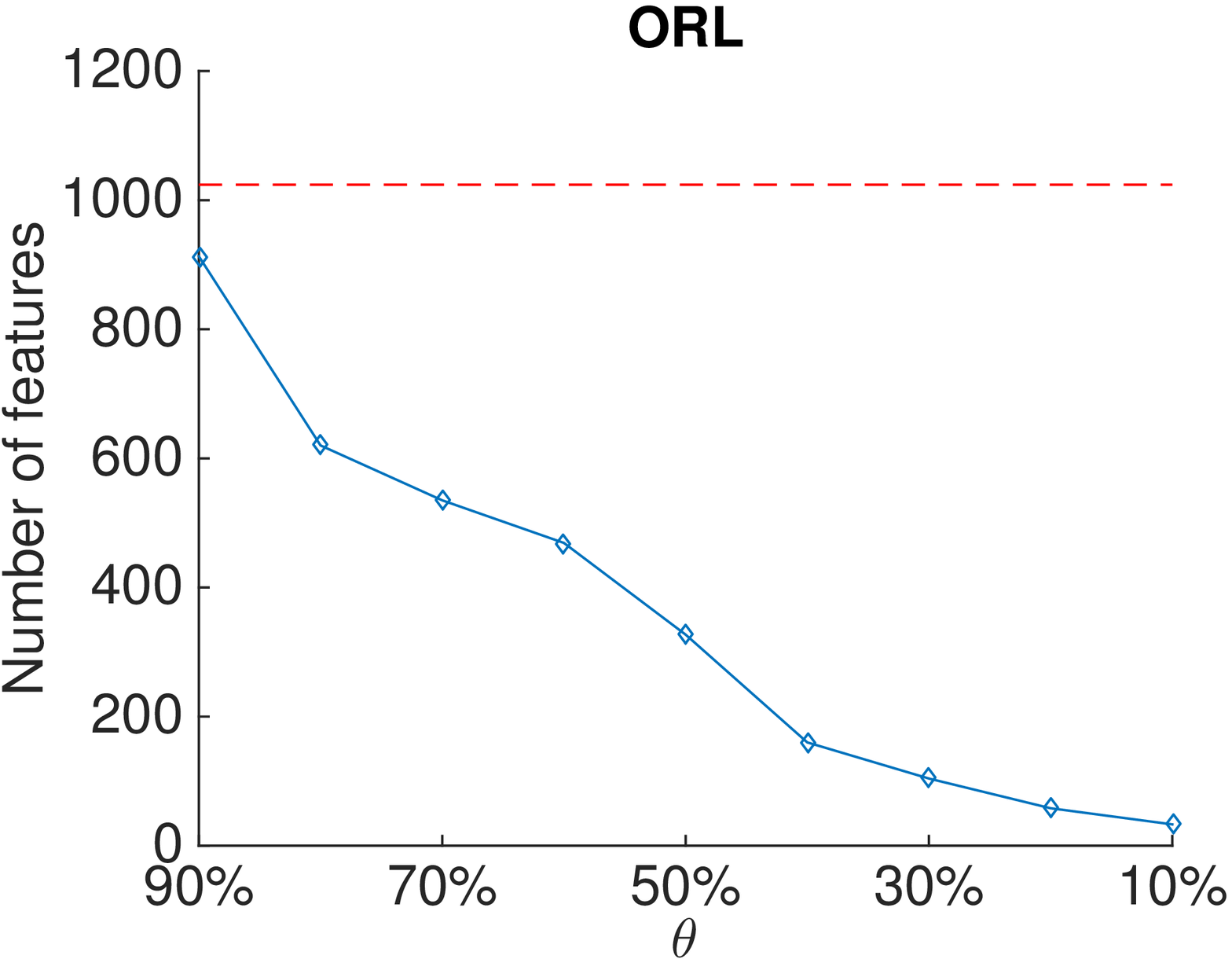} 	
\includegraphics[width=0.225\textwidth]{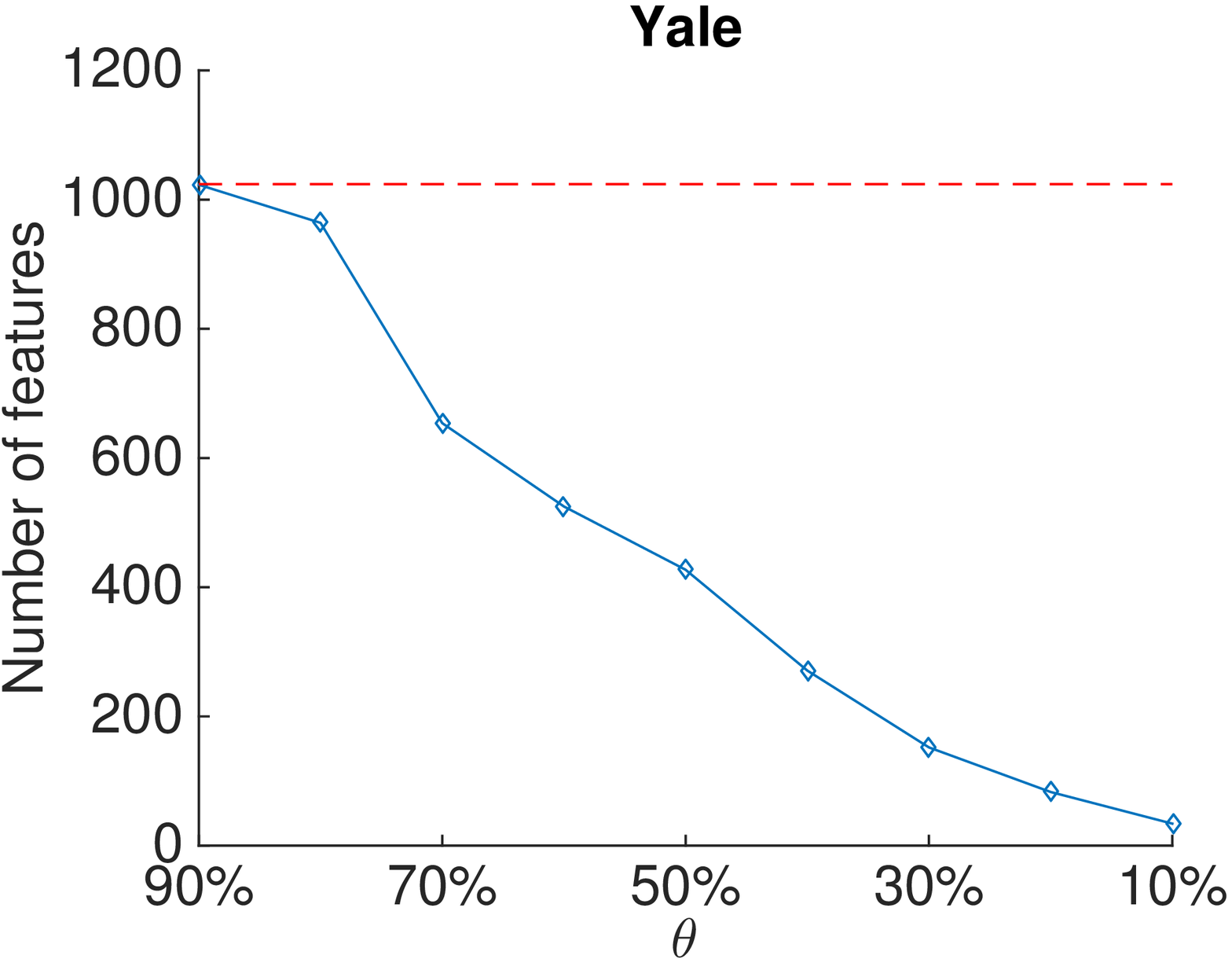} 
\includegraphics[width=0.225\textwidth]{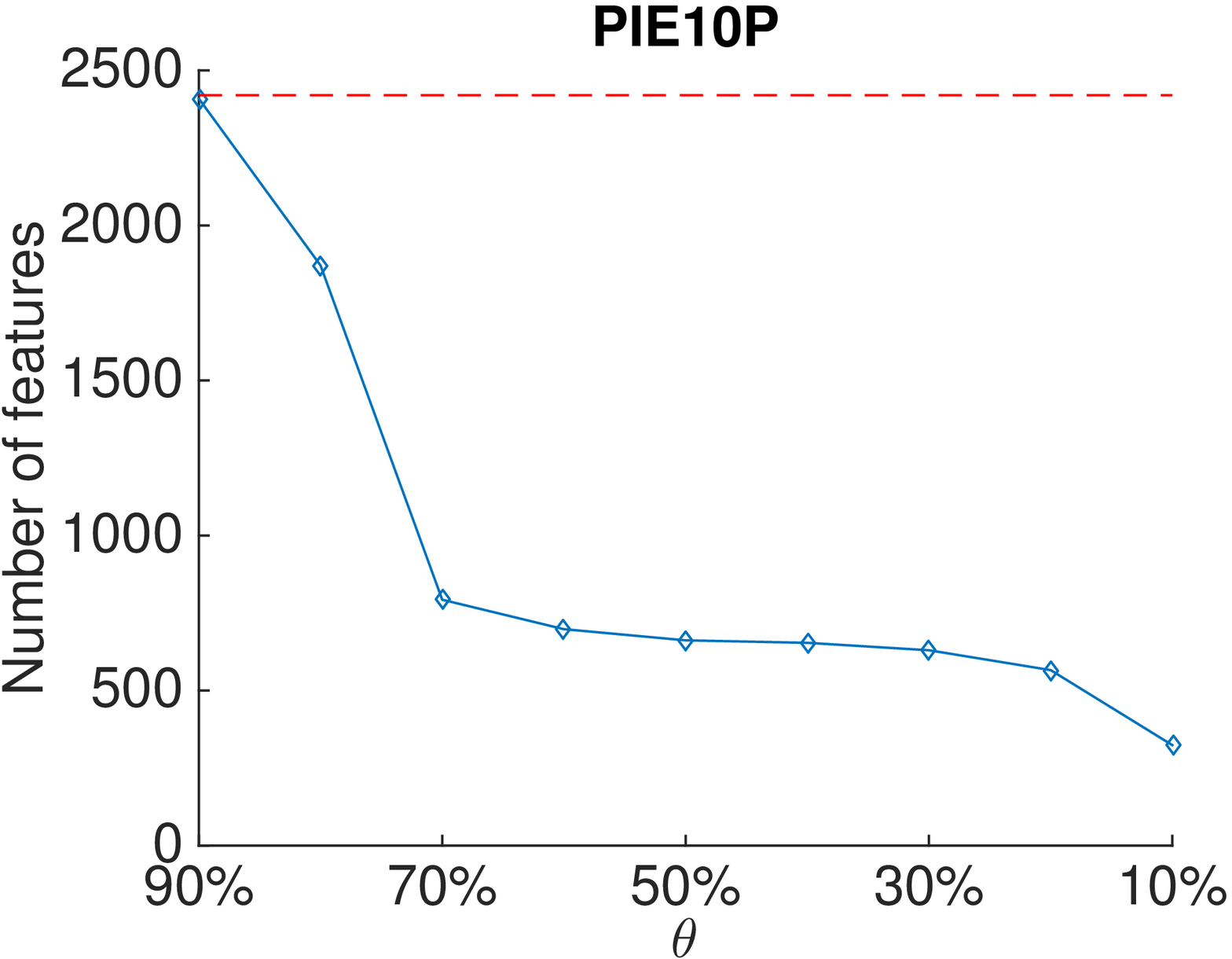}
\includegraphics[width=0.225\textwidth]{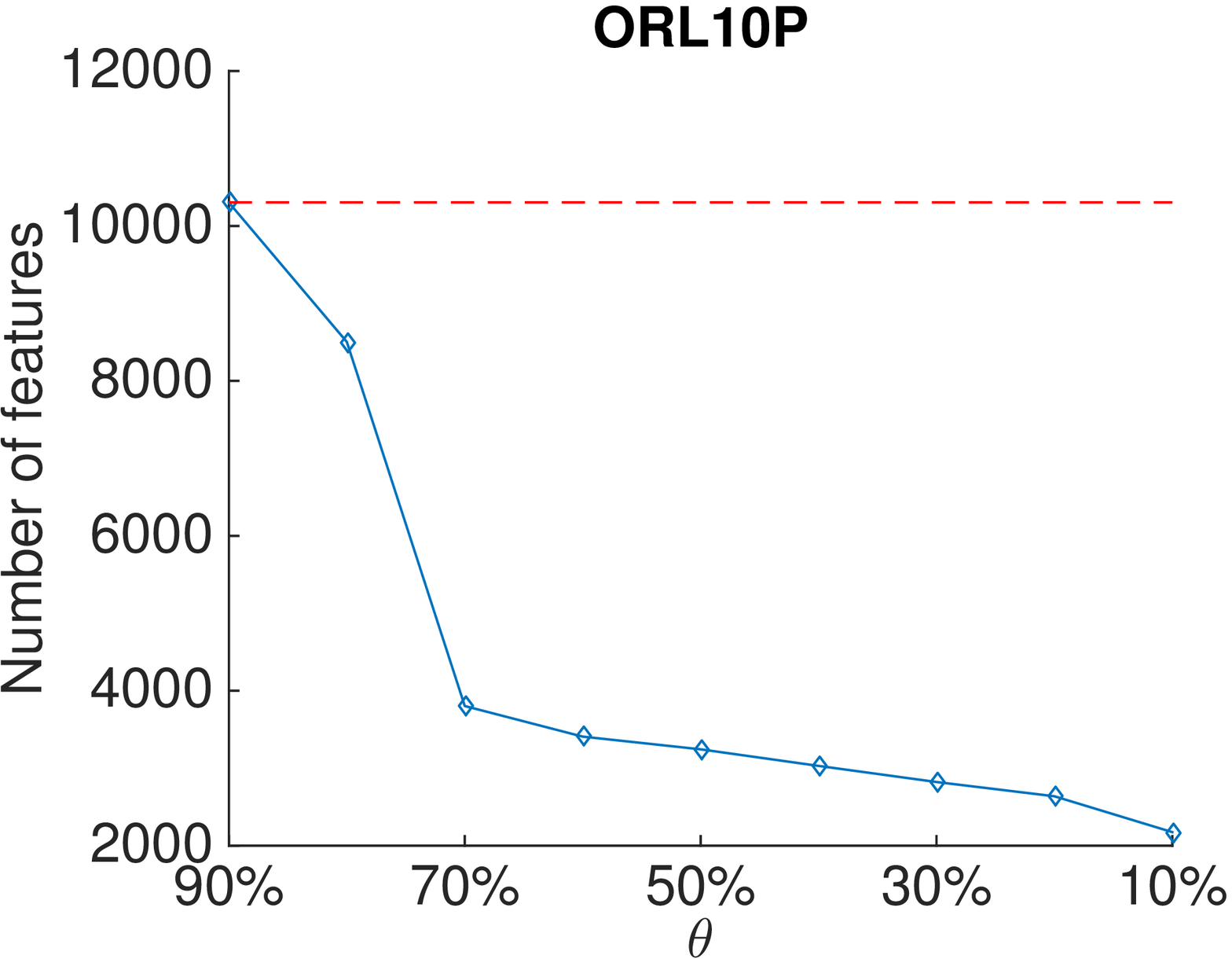} \\
\caption{Spectral clustering performance of Image datasets with different parameter $\theta$. Top row: NMI; Middle row: ACC; Bottom row: number of features, the red dash line means the size of raw dataset.}
\label{fig:tkdd_sc_image}
\end{figure}
\begin{figure}
	\centering
	\includegraphics[width=0.225\textwidth]{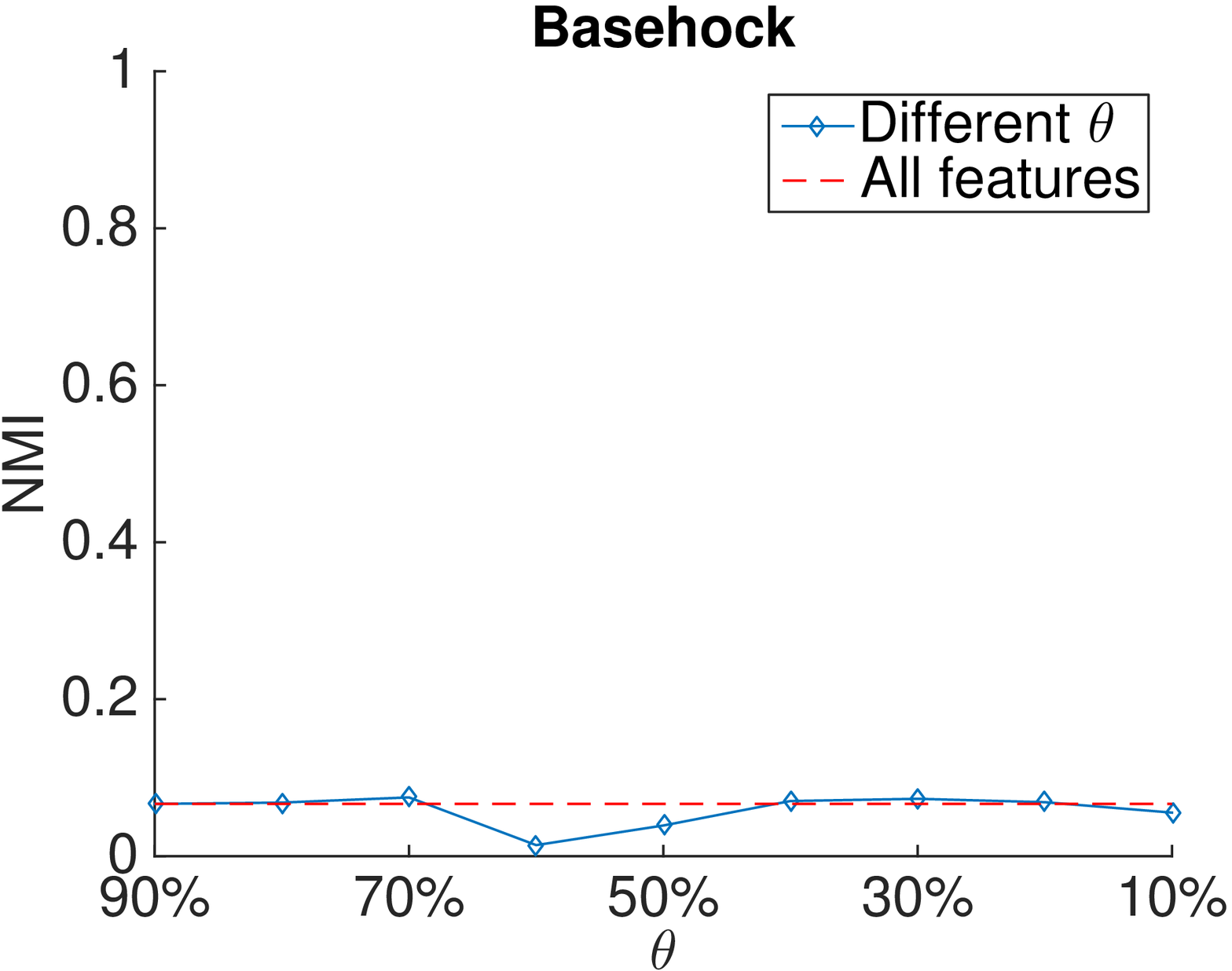} 	
	\includegraphics[width=0.225\textwidth]{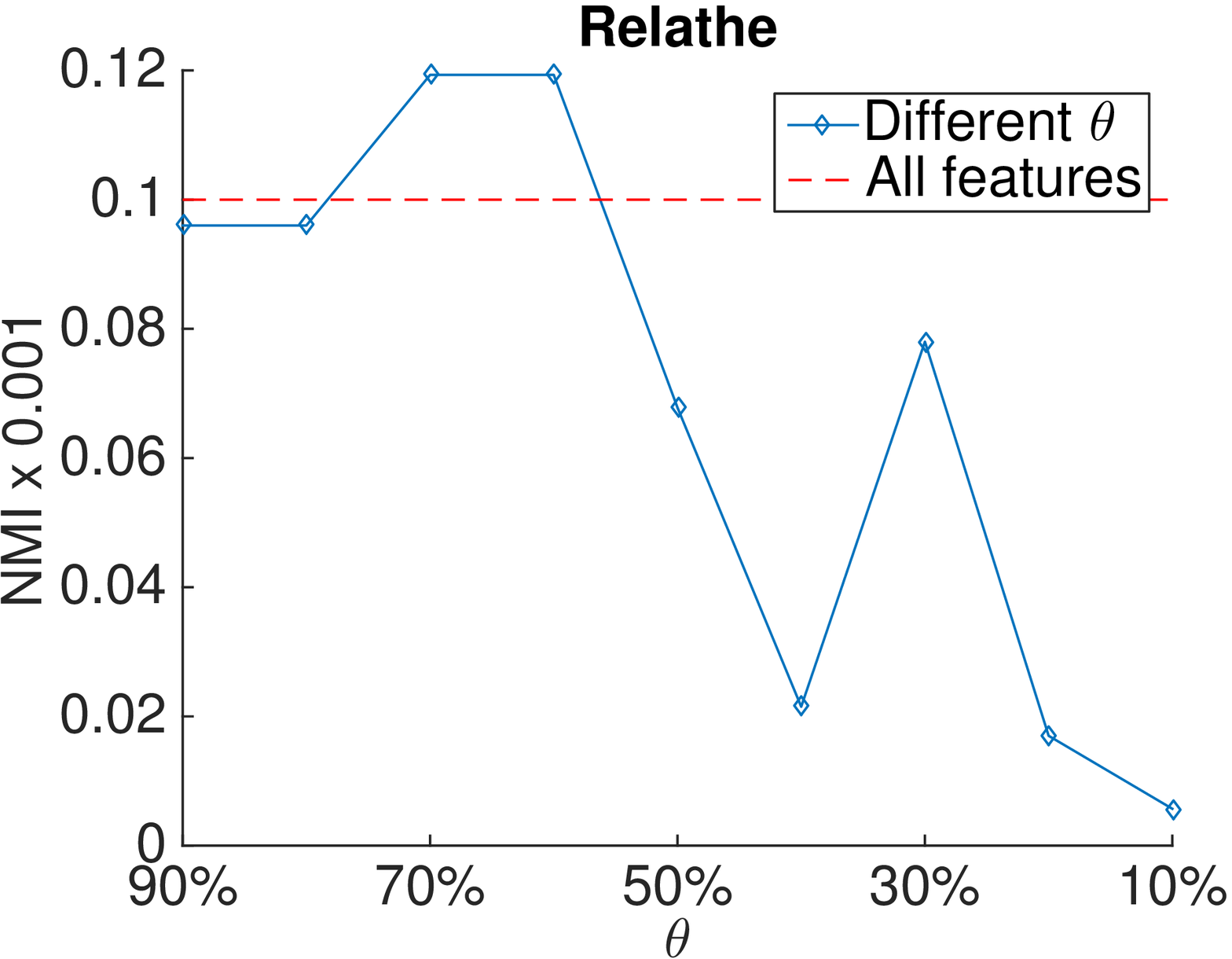} 
	\includegraphics[width=0.225\textwidth]{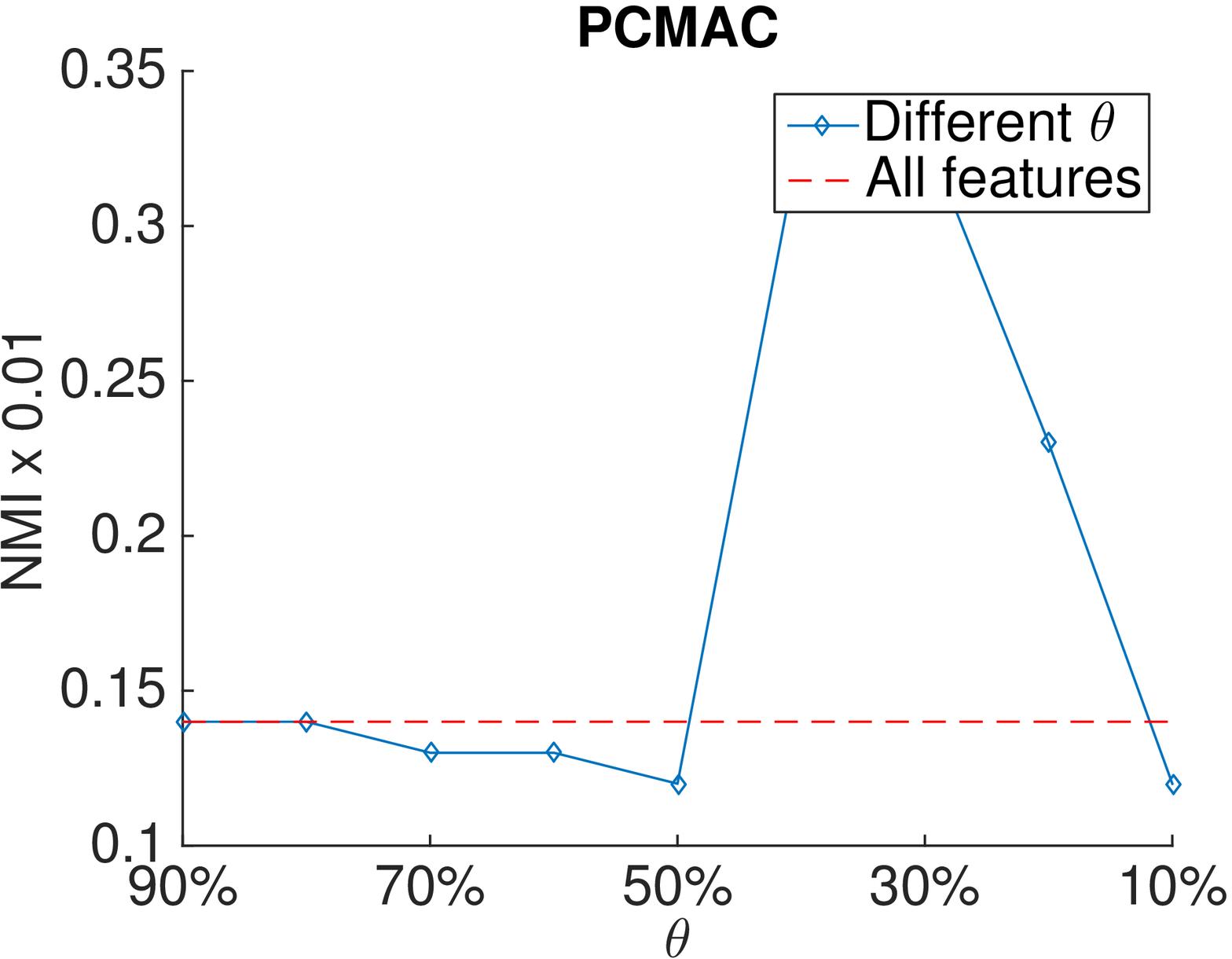}
	\includegraphics[width=0.225\textwidth]{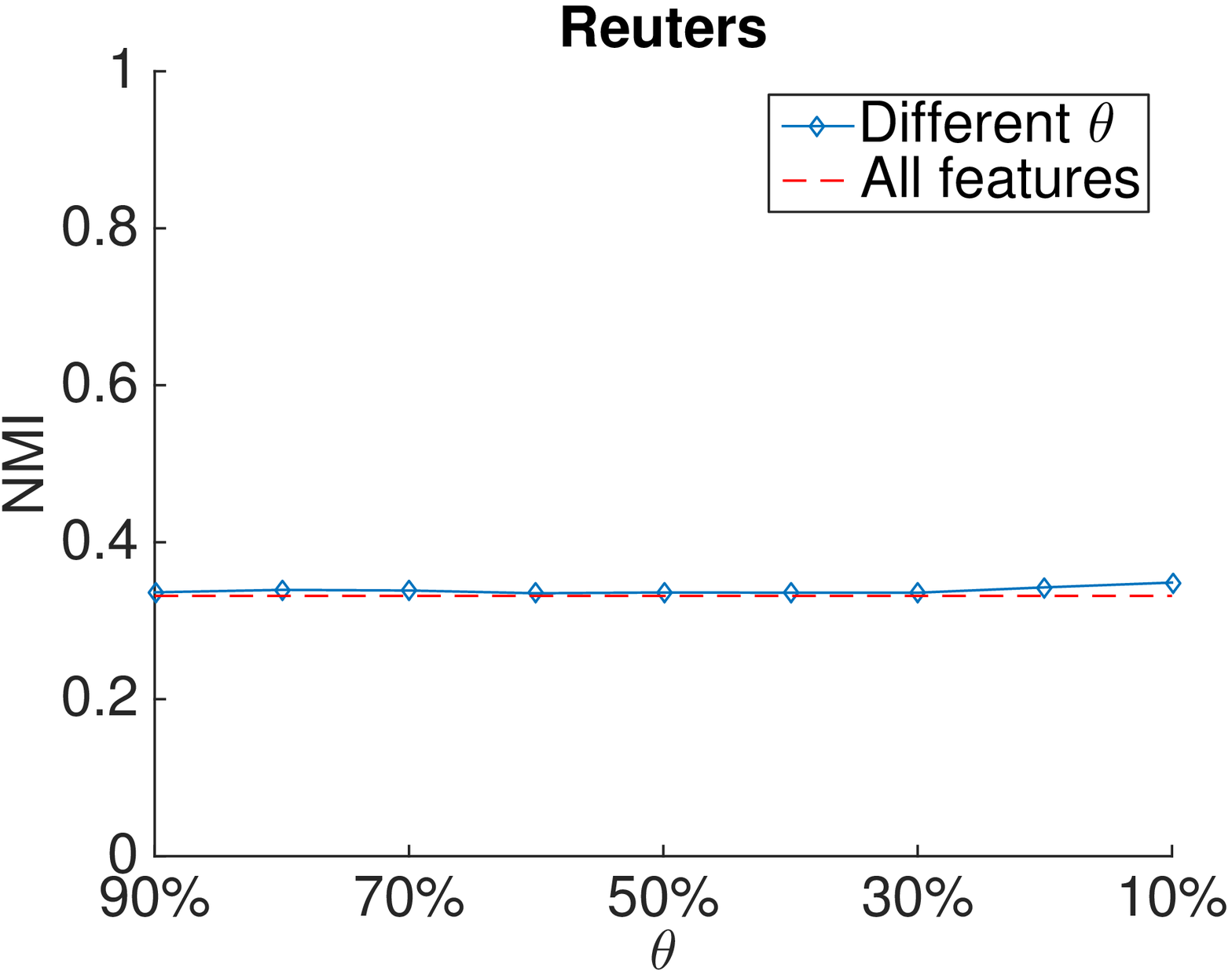} \\
	\includegraphics[width=0.225\textwidth]{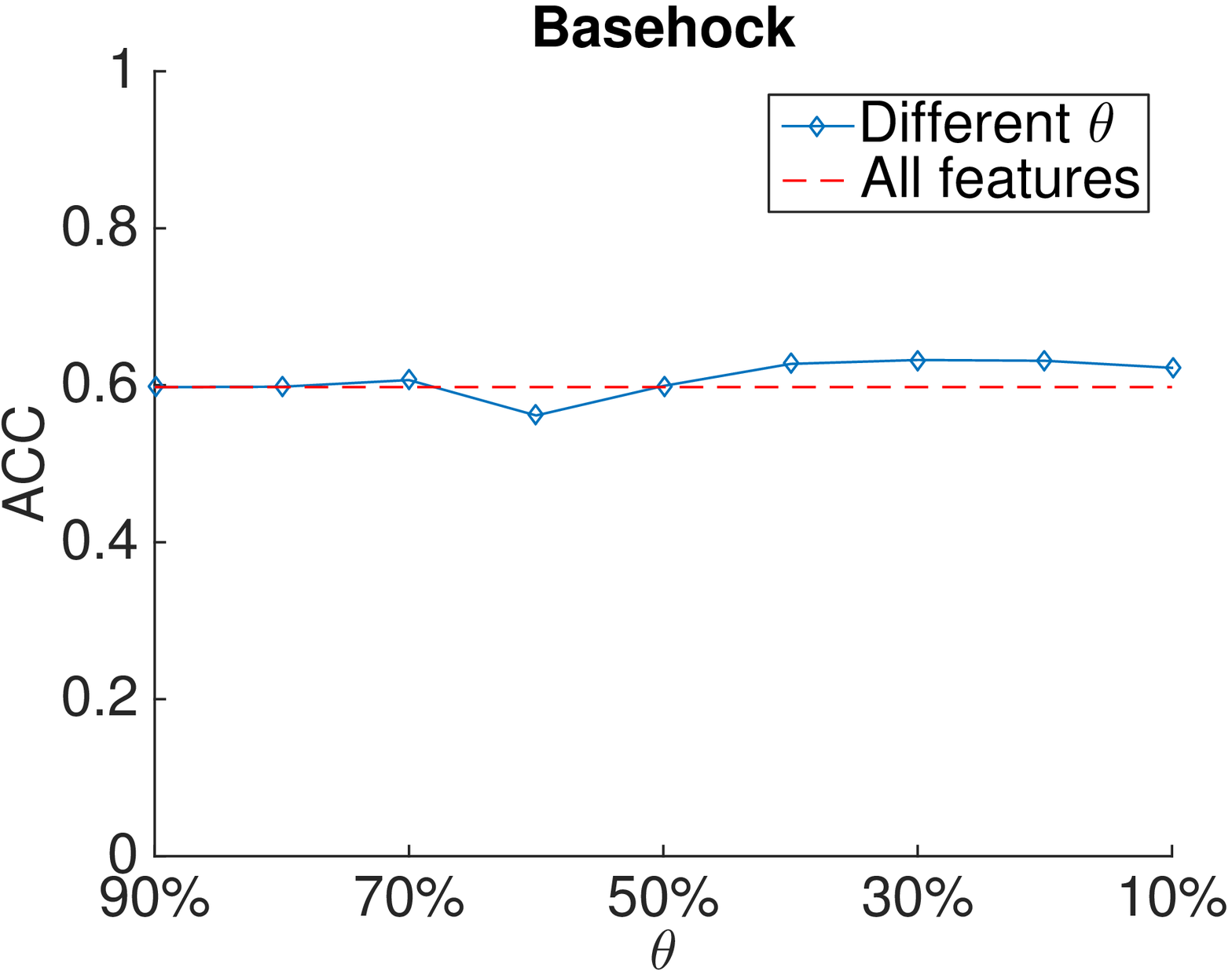} 	
	\includegraphics[width=0.225\textwidth]{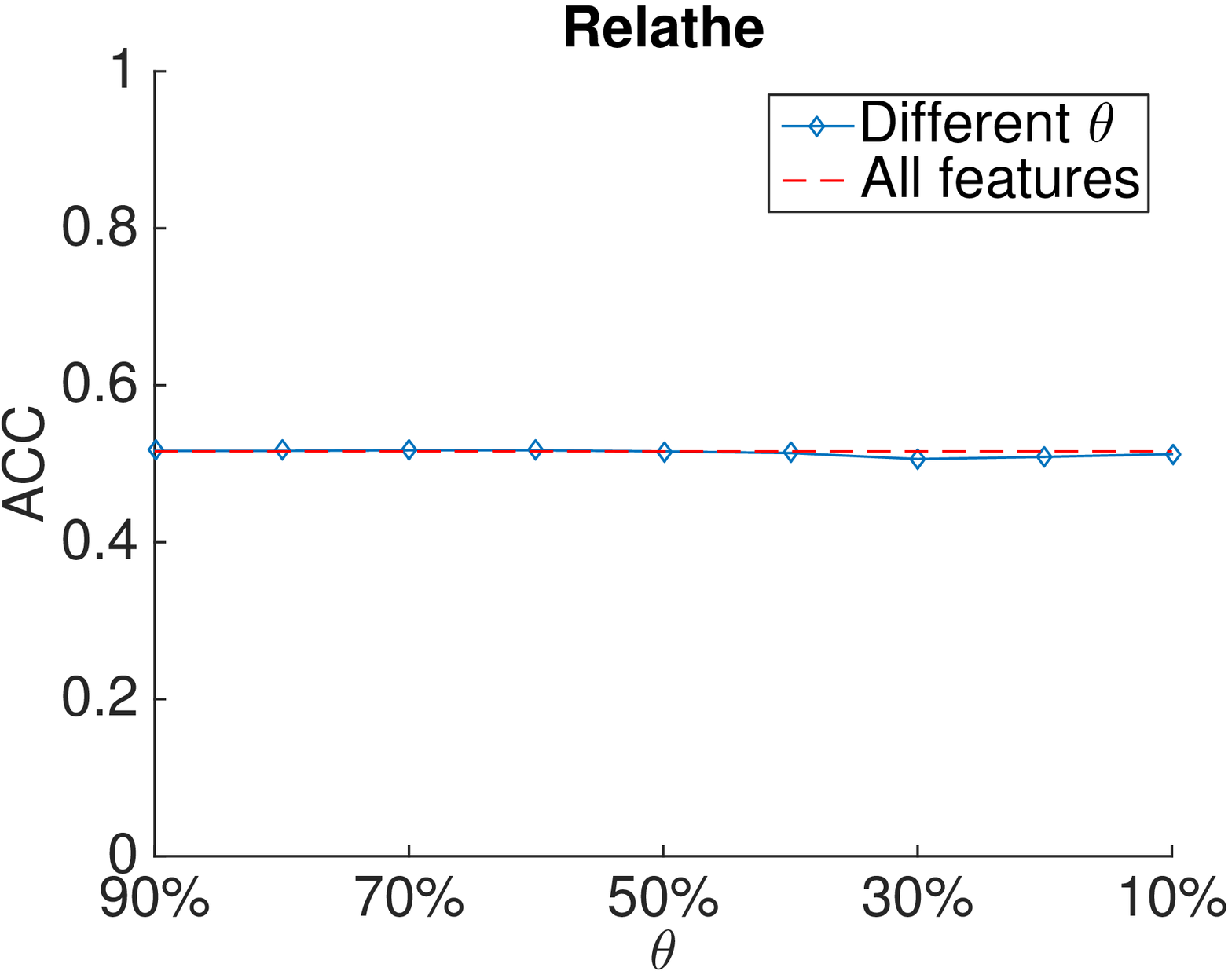} 
	\includegraphics[width=0.225\textwidth]{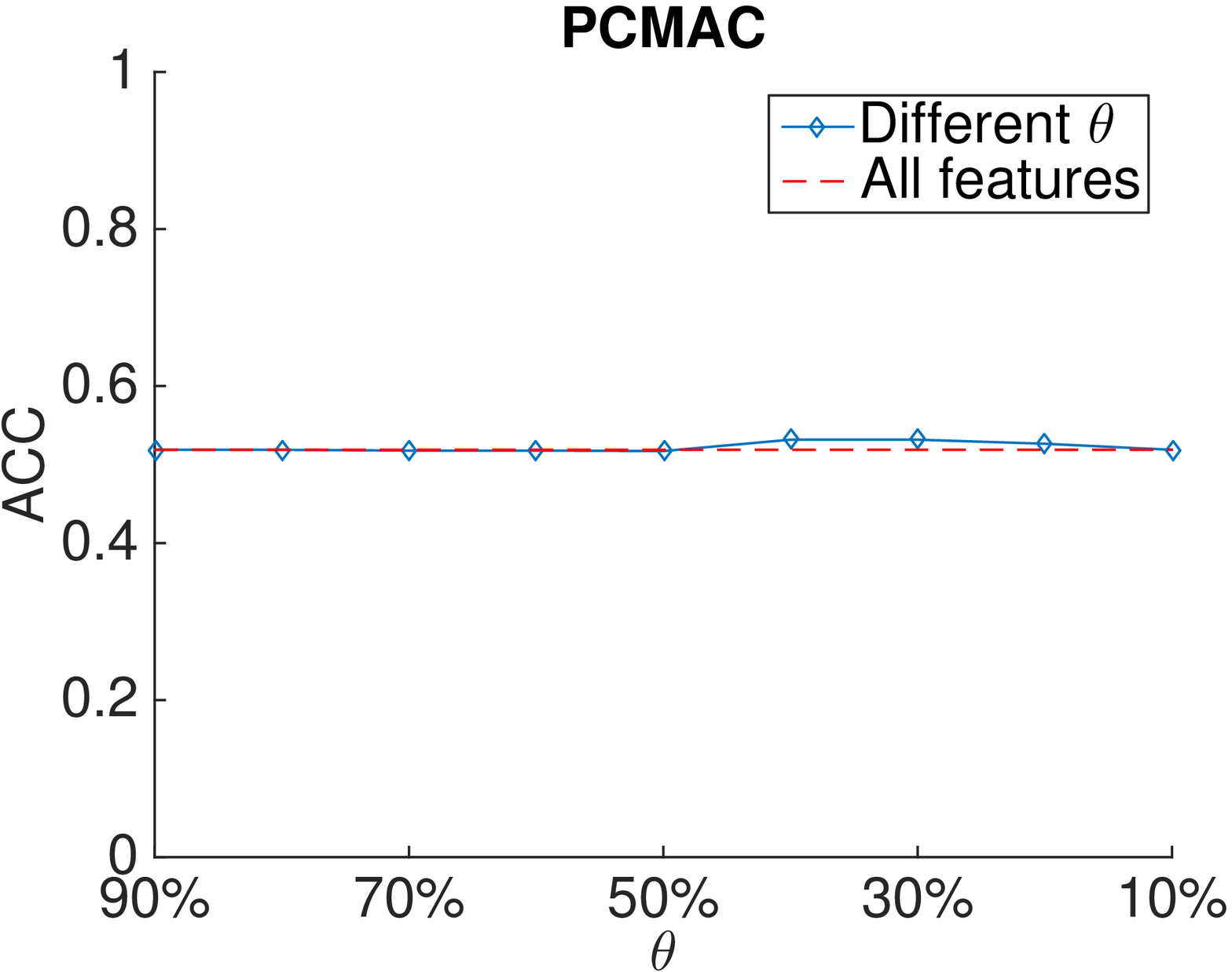}
	\includegraphics[width=0.225\textwidth]{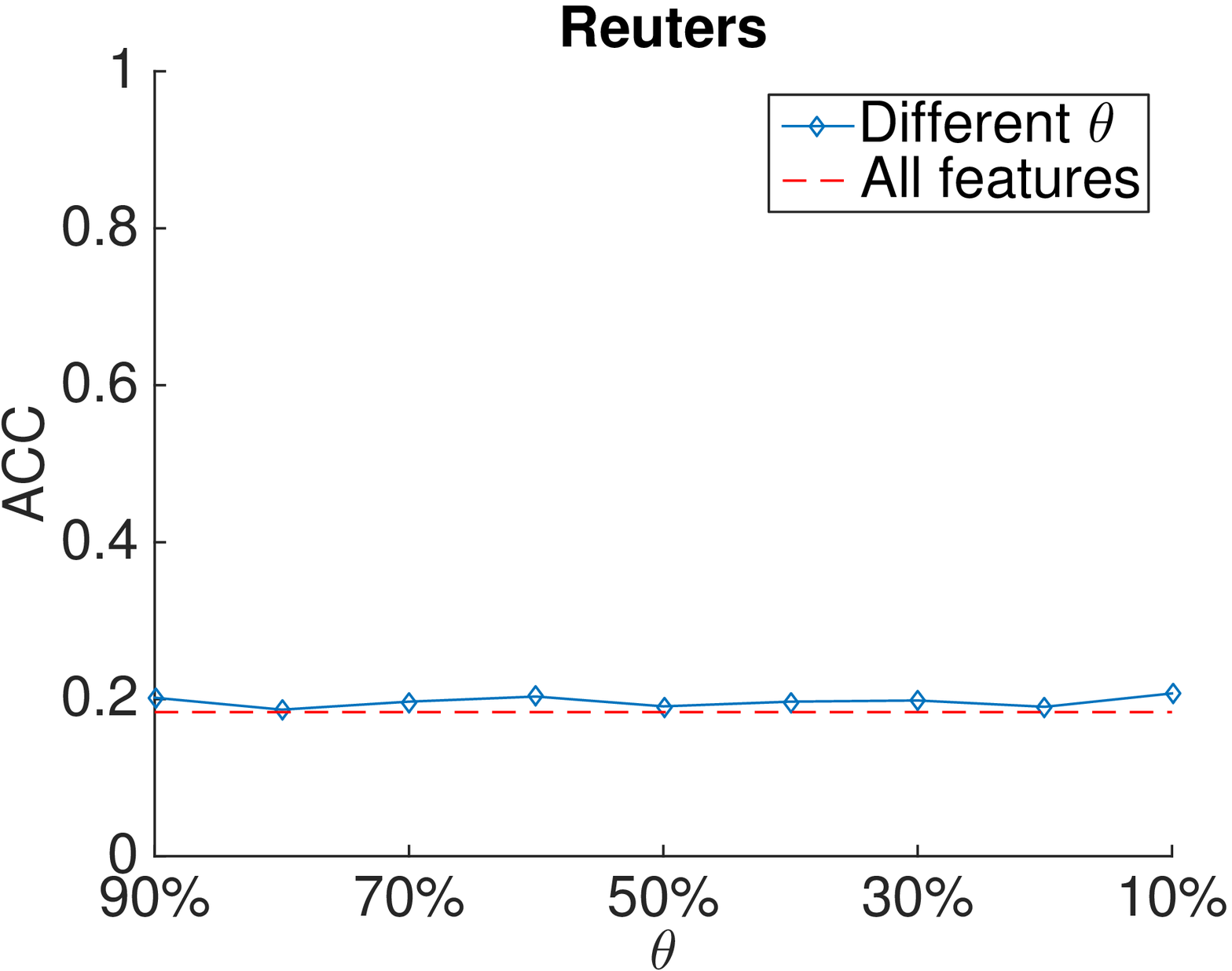} \\
	\includegraphics[width=0.225\textwidth]{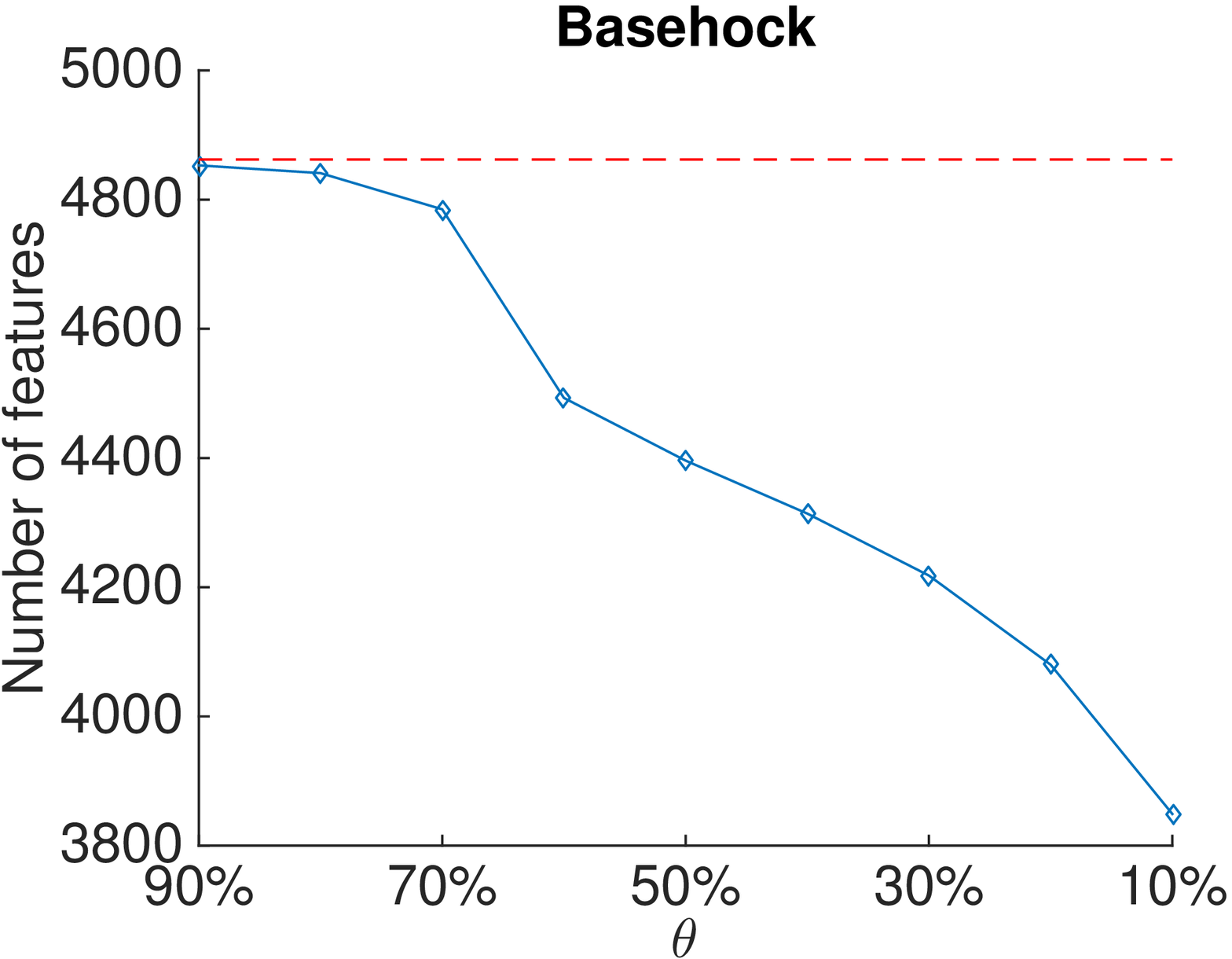} 	
	\includegraphics[width=0.225\textwidth]{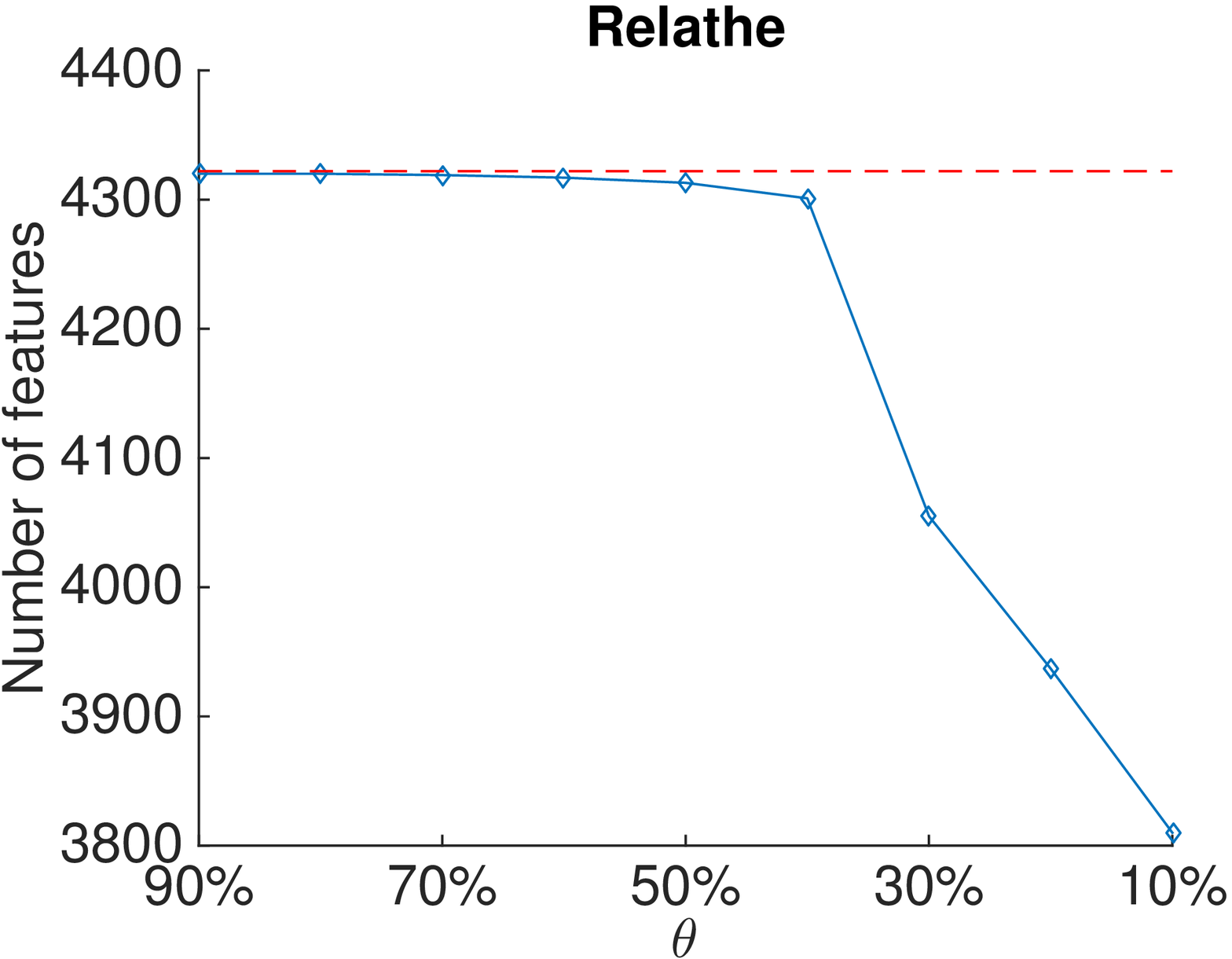} 
	\includegraphics[width=0.225\textwidth]{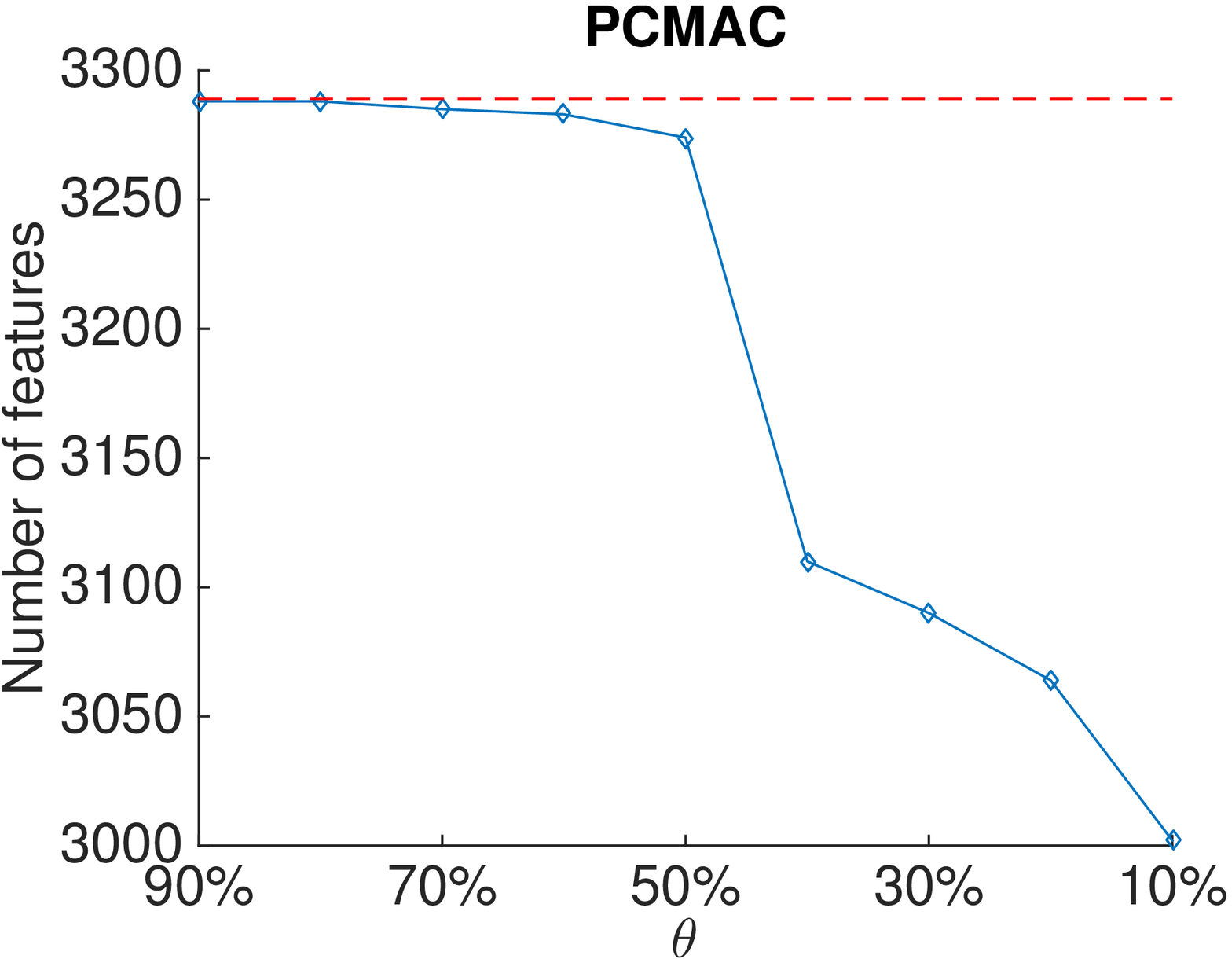}
	\includegraphics[width=0.225\textwidth]{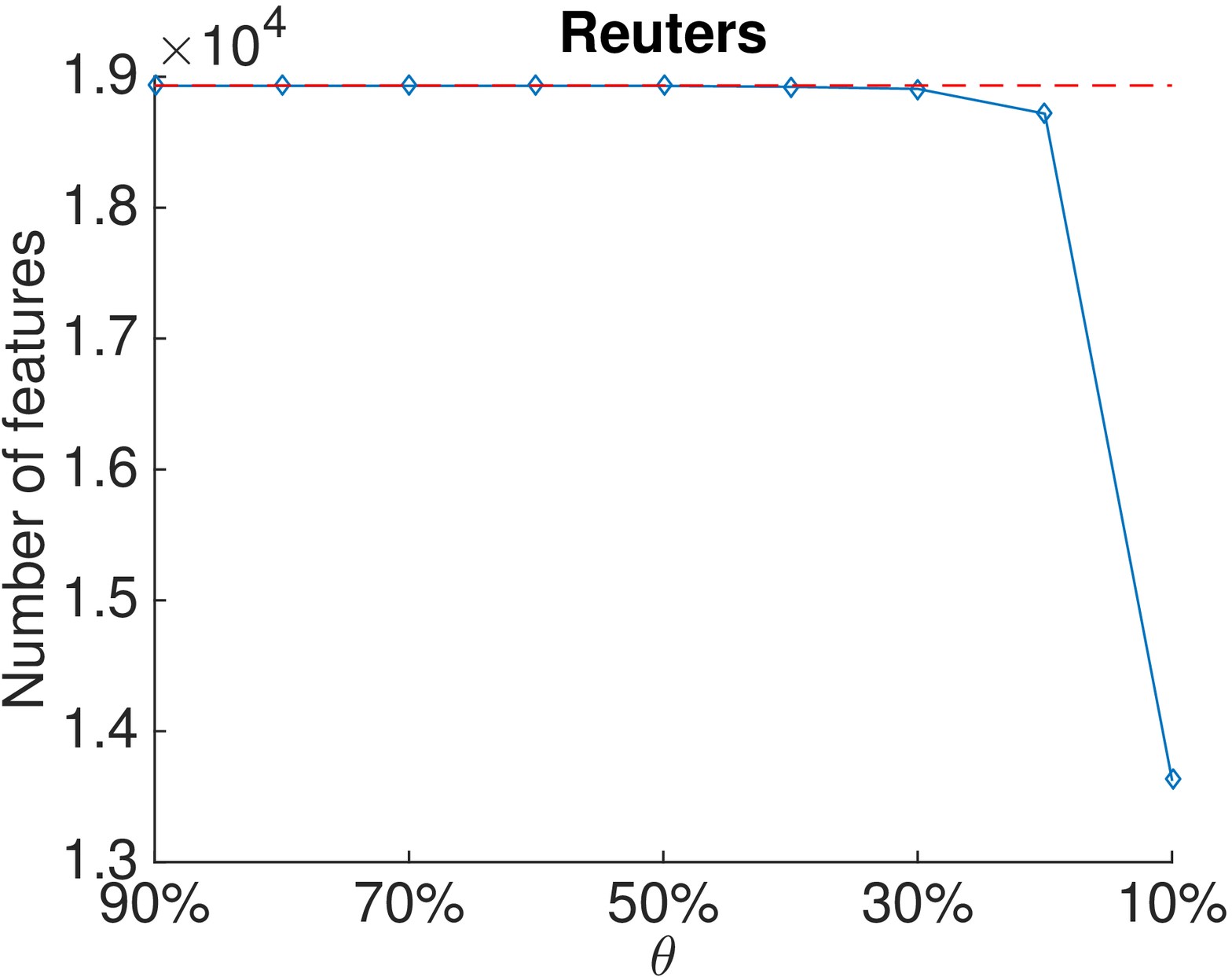} \\
	\caption{Spectral clustering performance of Text datasets with different parameter $\theta$. Top row: NMI; Middle row: ACC; Bottom row: number of features, the red dash line means the size of raw dataset.}
	\label{fig:tkdd_sc_text}
\end{figure}
\begin{figure}
	\centering
	\includegraphics[width=0.225\textwidth]{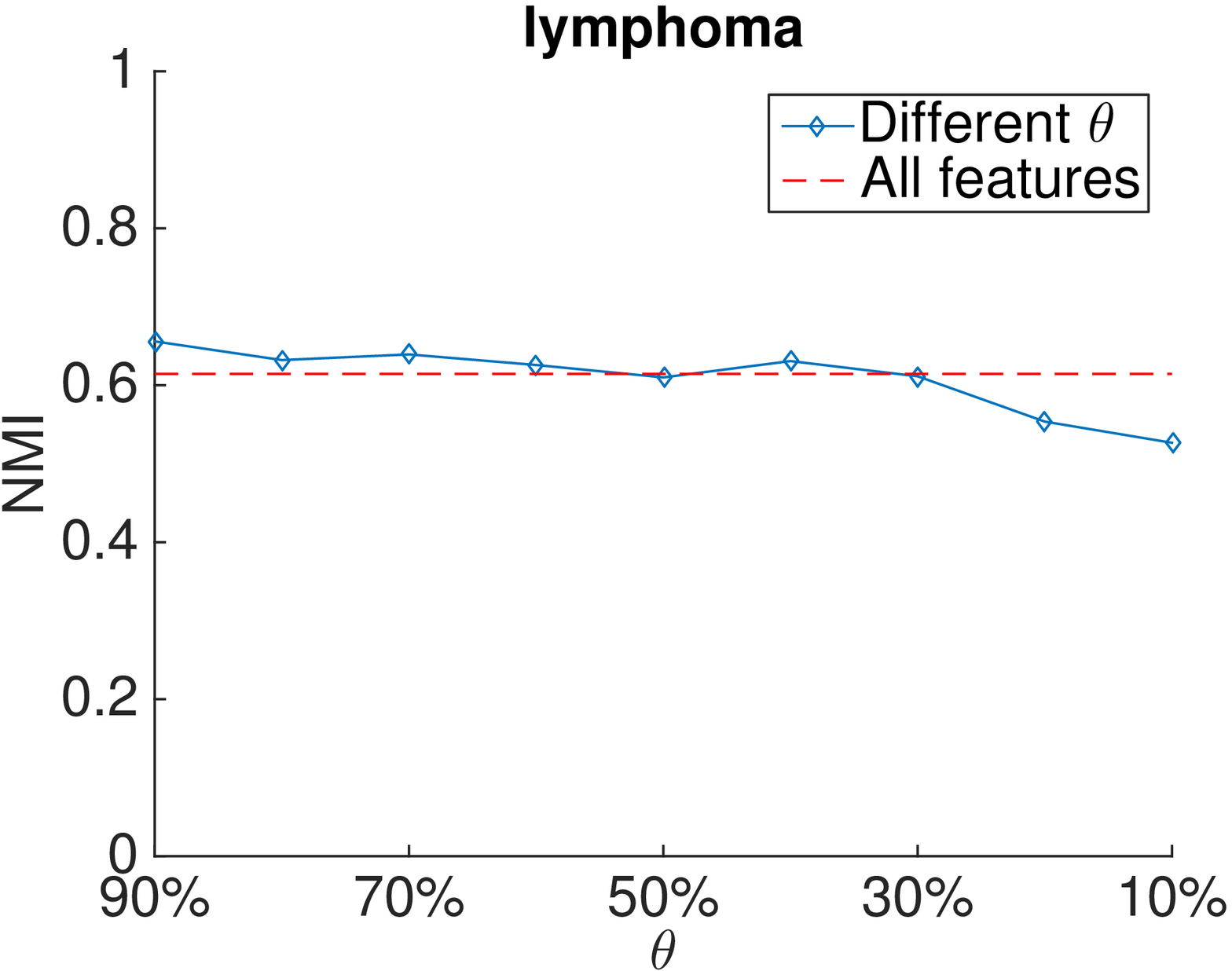} 	
	\includegraphics[width=0.225\textwidth]{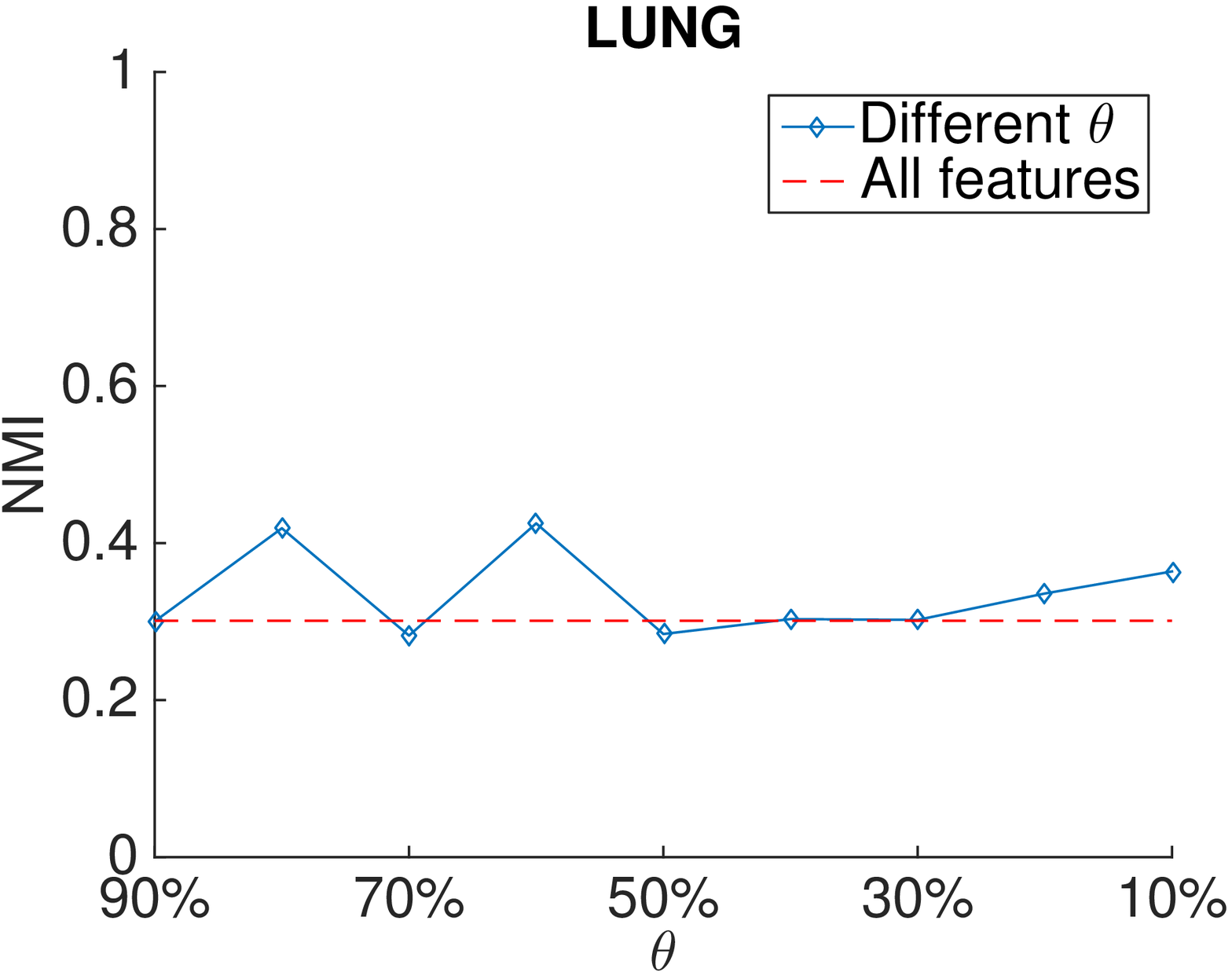} 
	\includegraphics[width=0.225\textwidth]{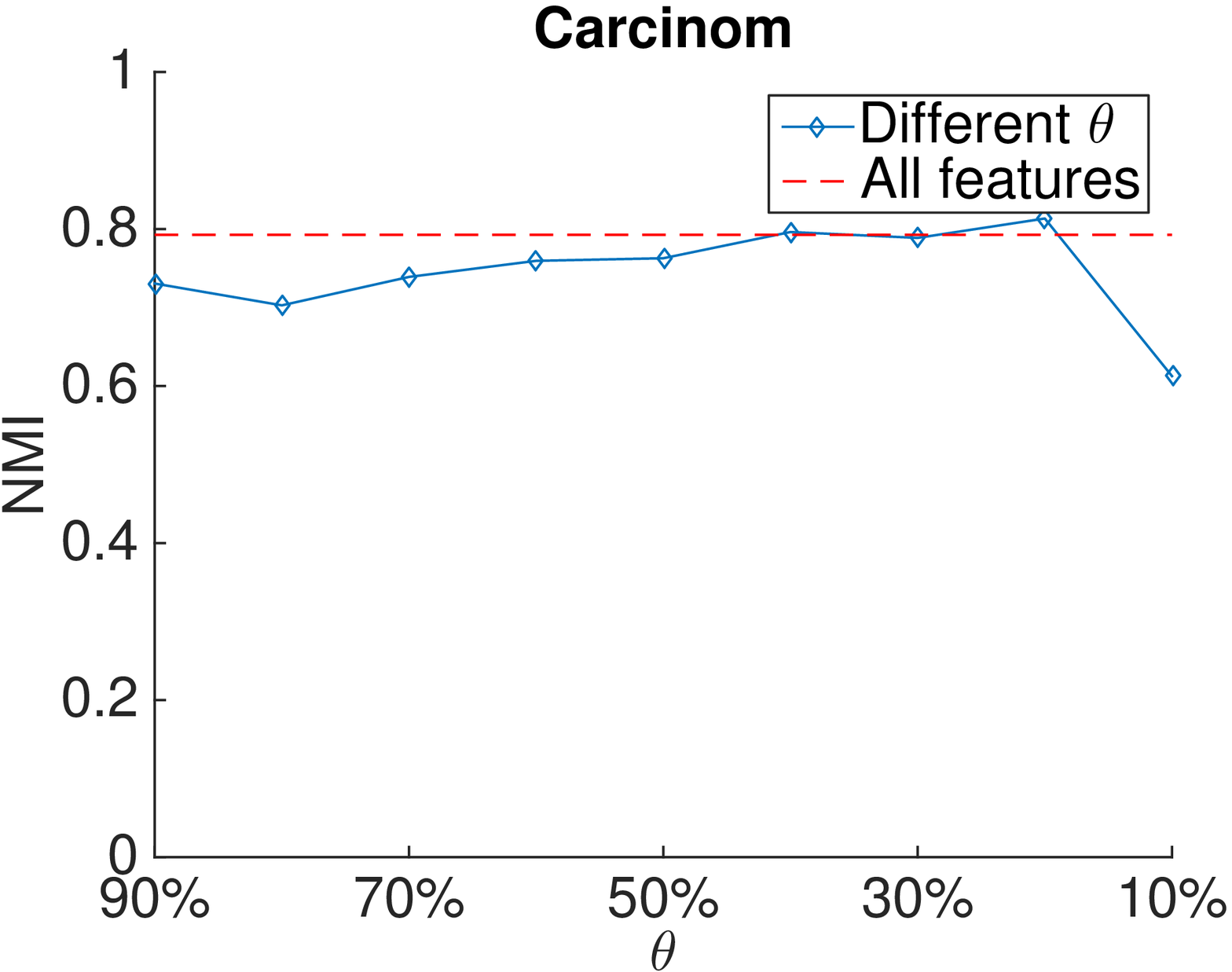}
	\includegraphics[width=0.225\textwidth]{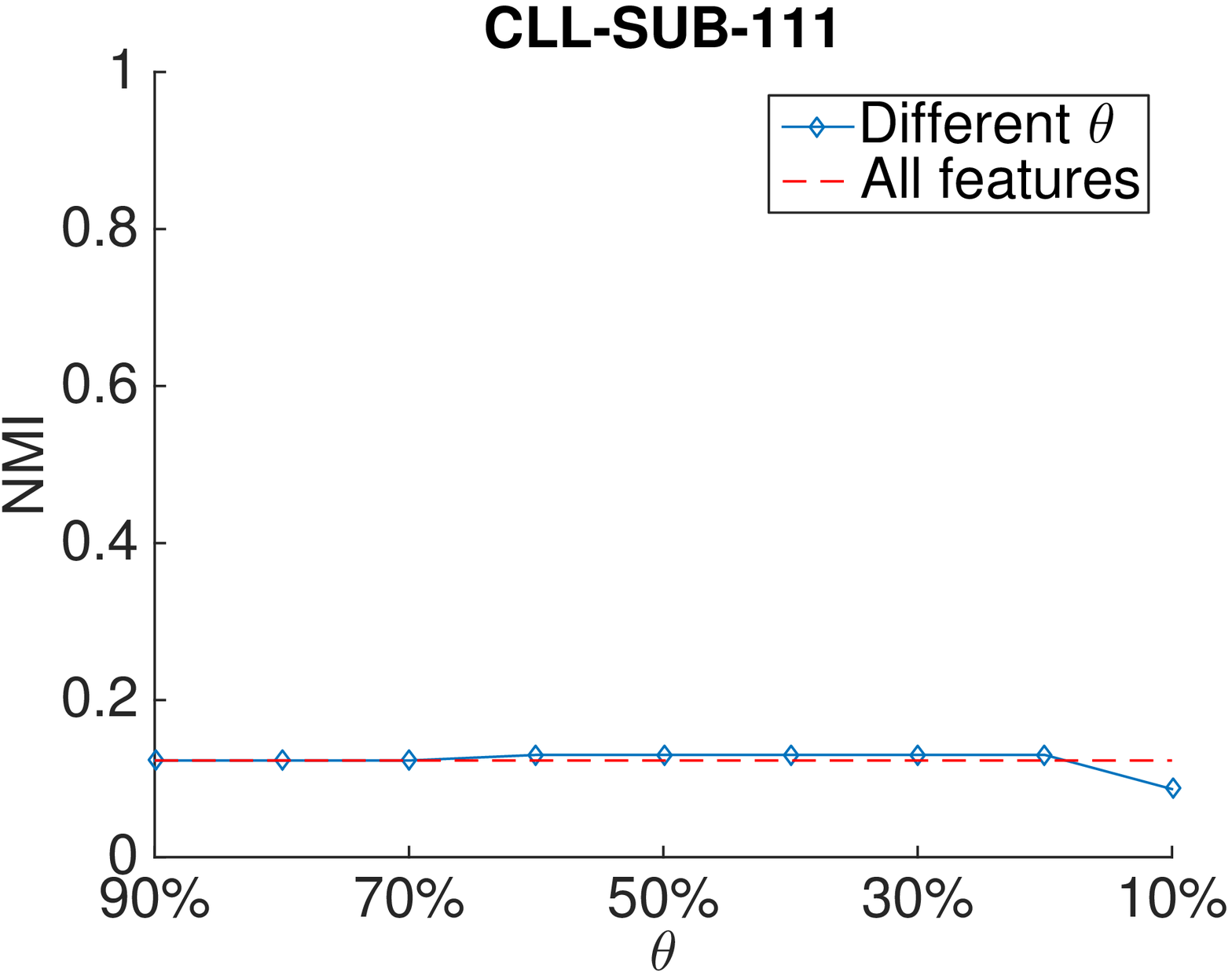} \\
	\includegraphics[width=0.225\textwidth]{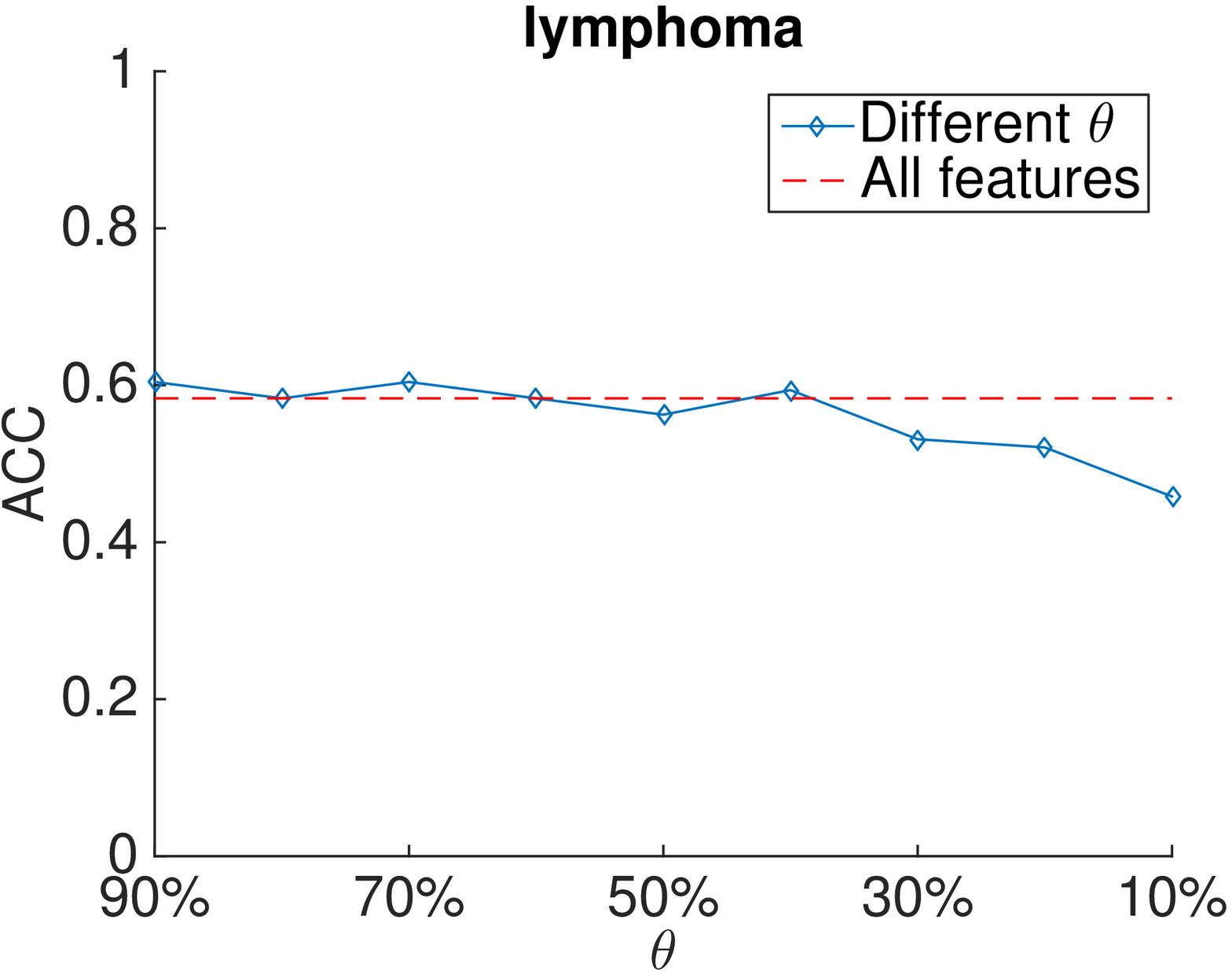} 	
	\includegraphics[width=0.225\textwidth]{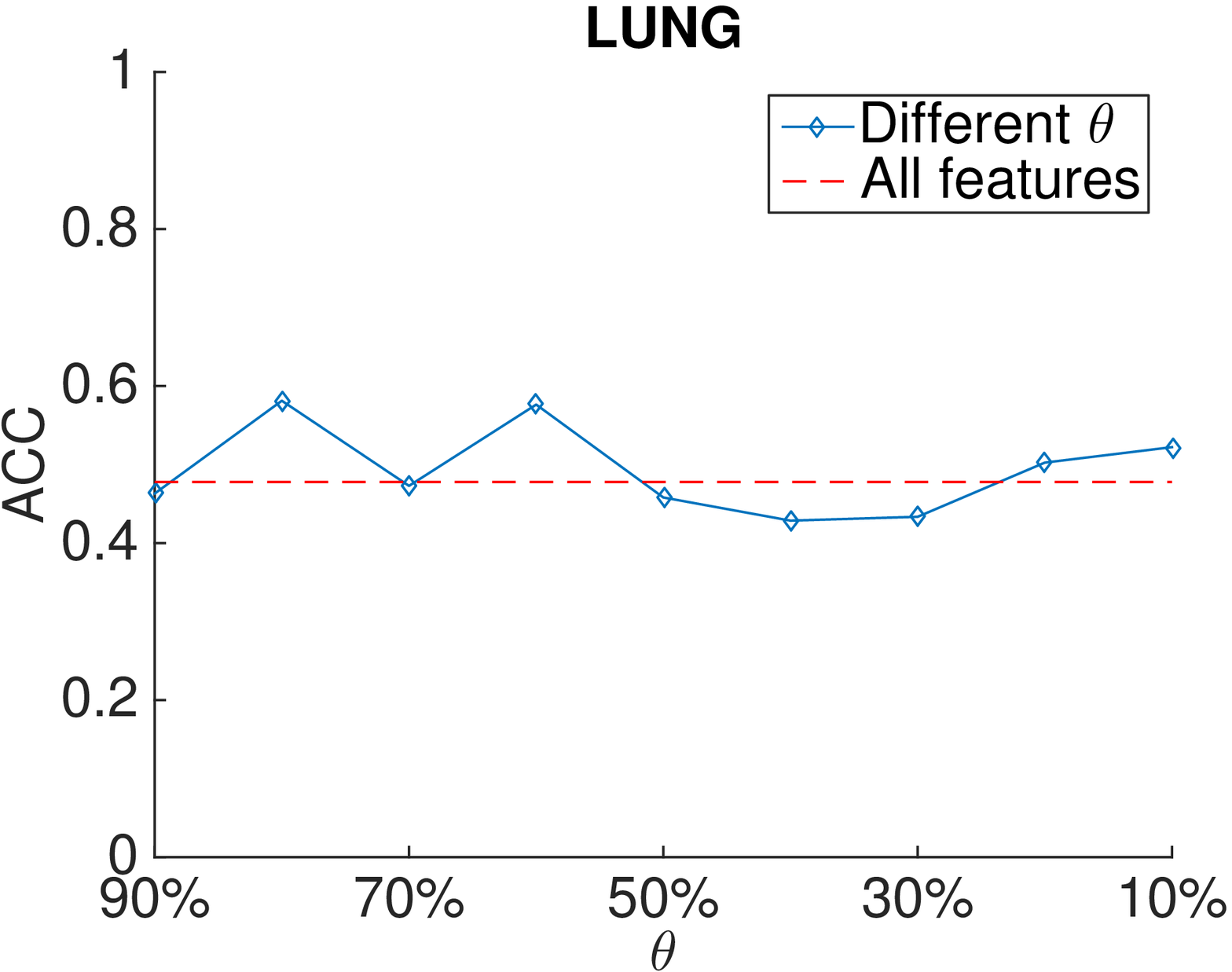} 
	\includegraphics[width=0.225\textwidth]{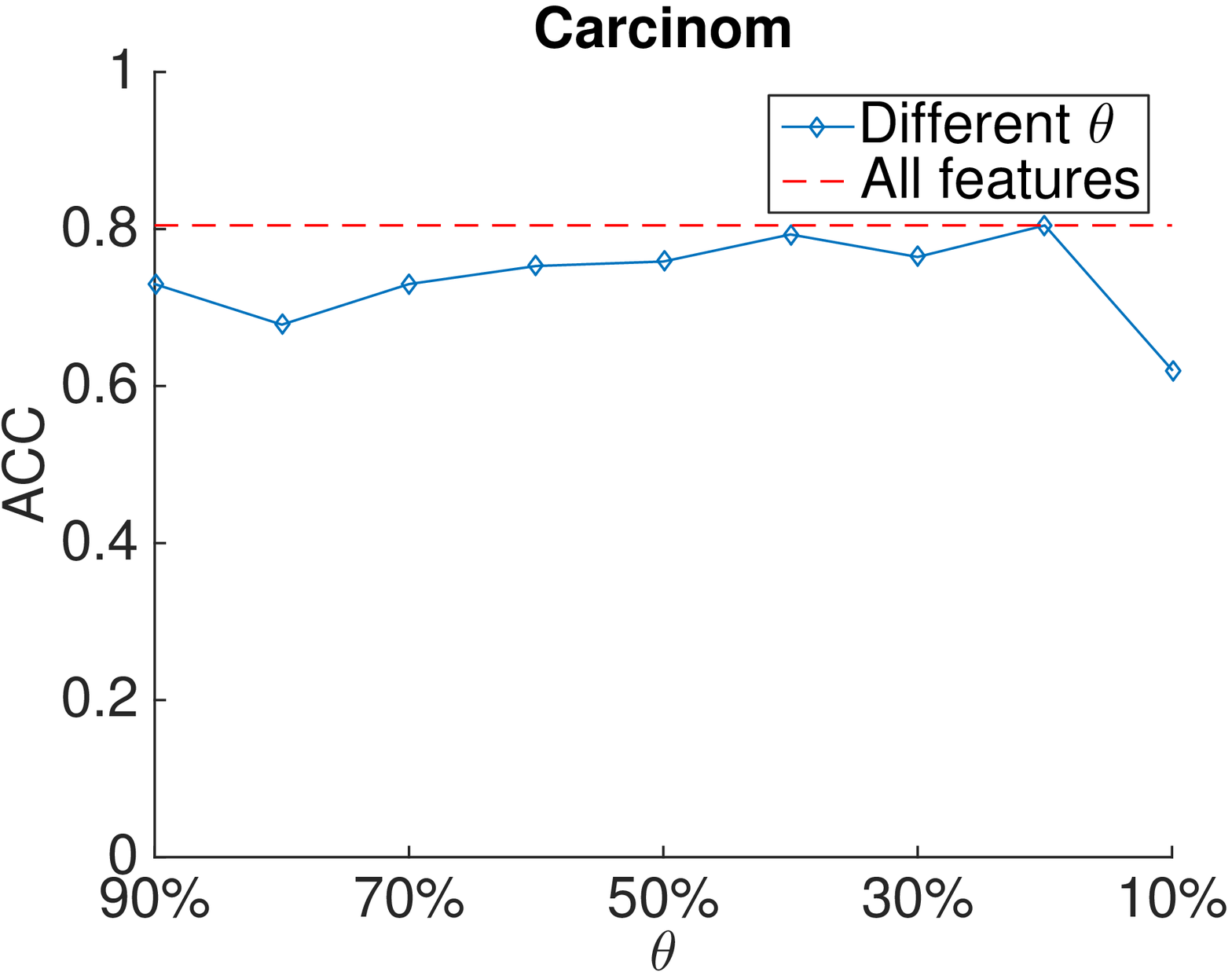}
	\includegraphics[width=0.225\textwidth]{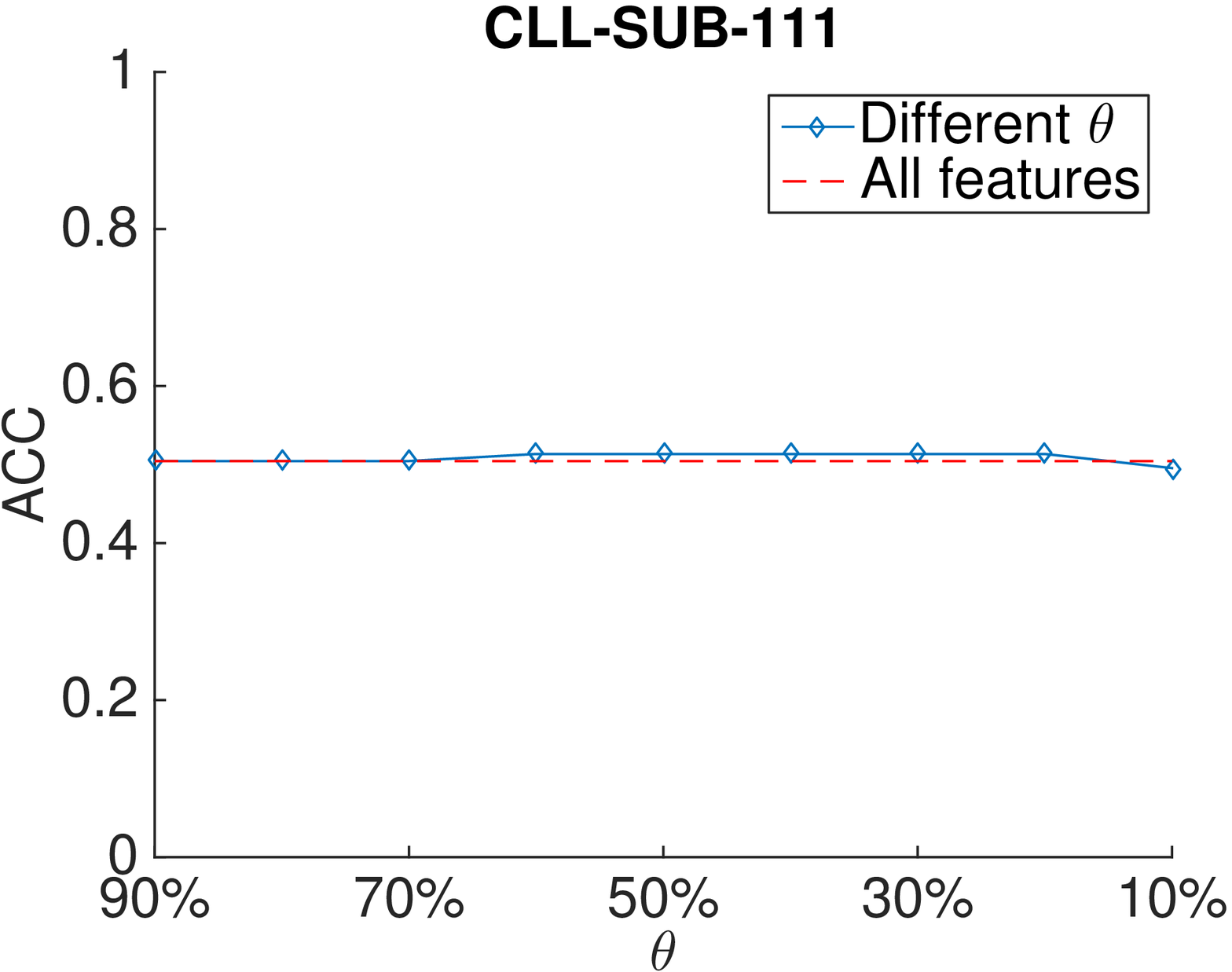} \\
	\includegraphics[width=0.225\textwidth]{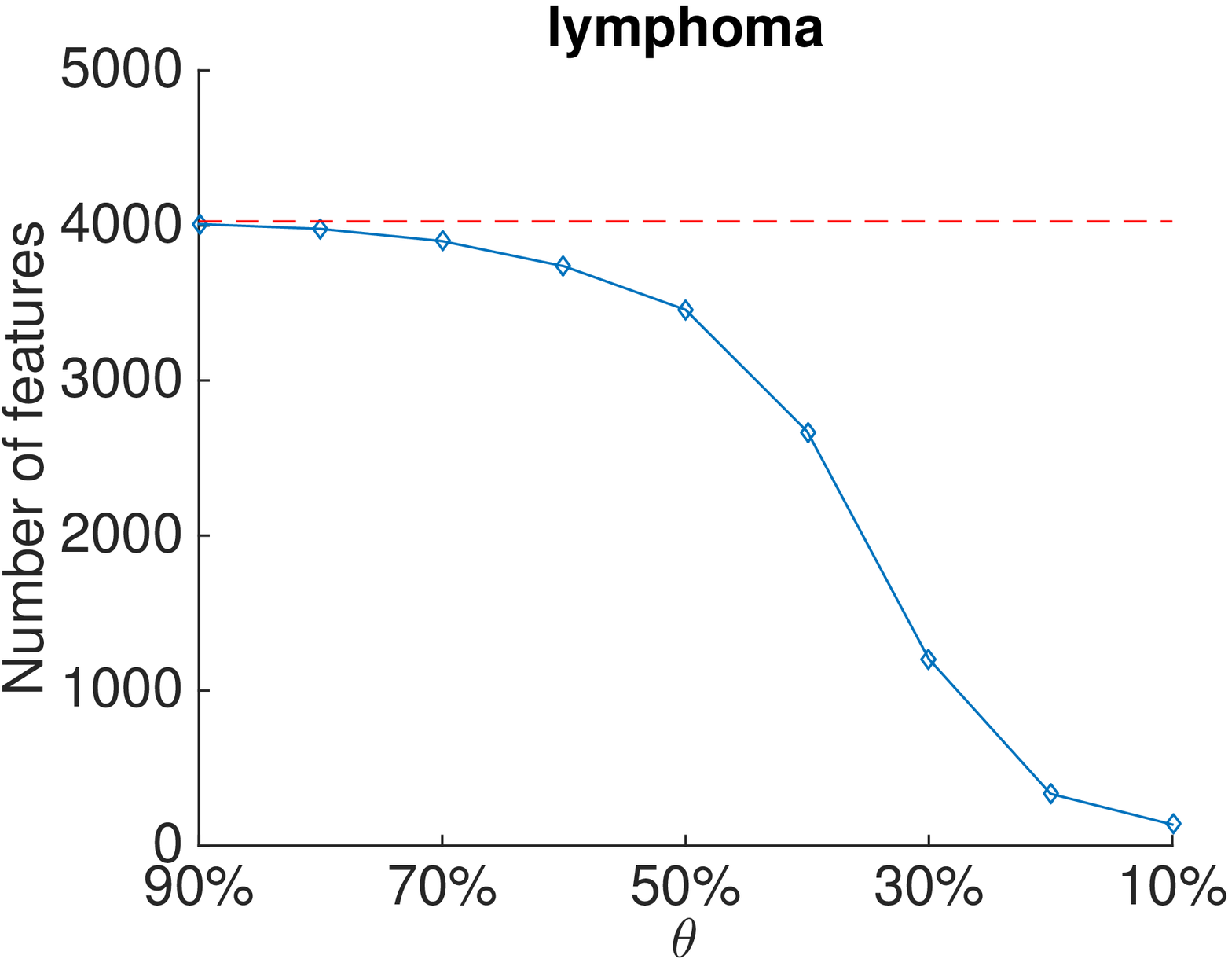} 	
	\includegraphics[width=0.225\textwidth]{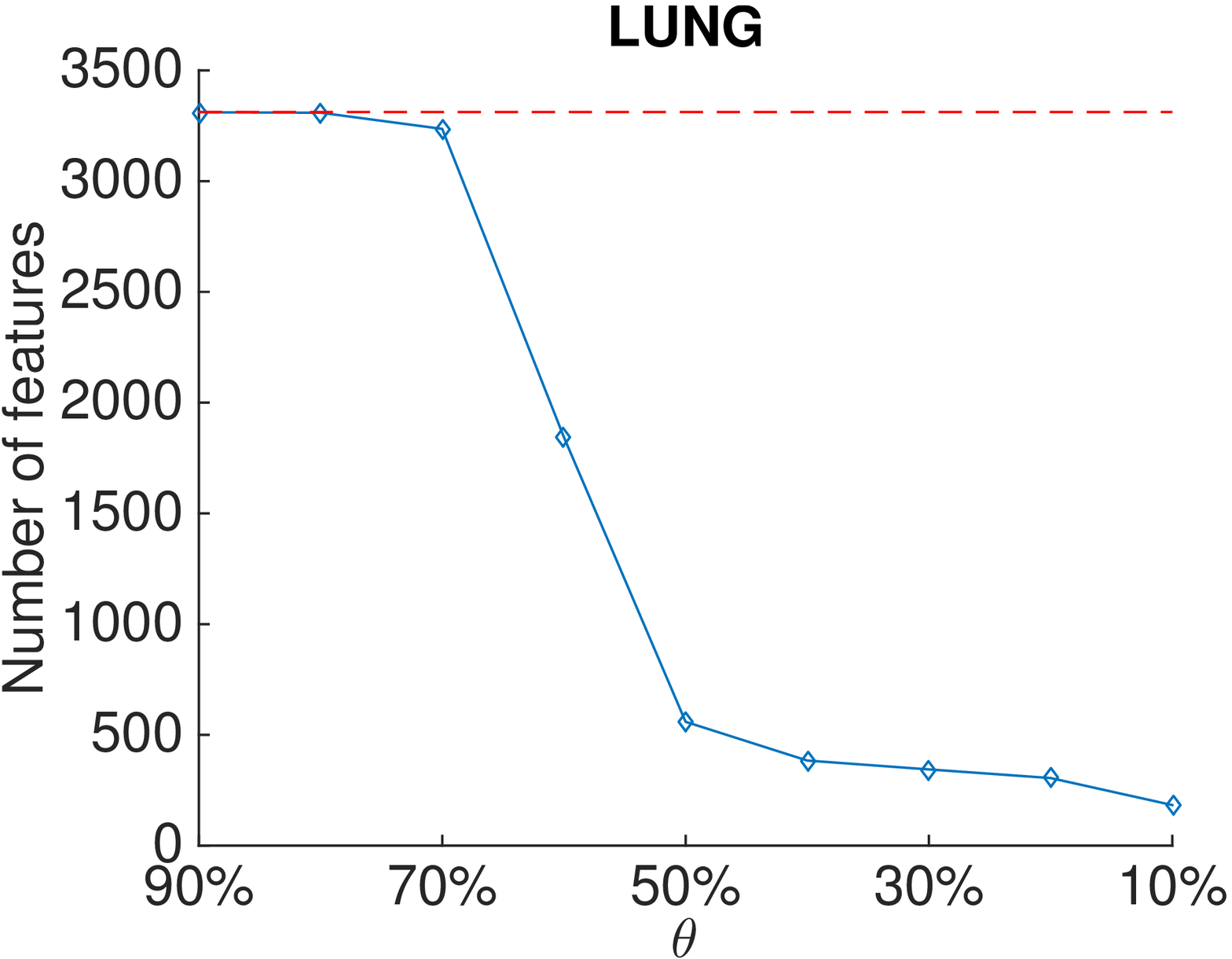} 
	\includegraphics[width=0.225\textwidth]{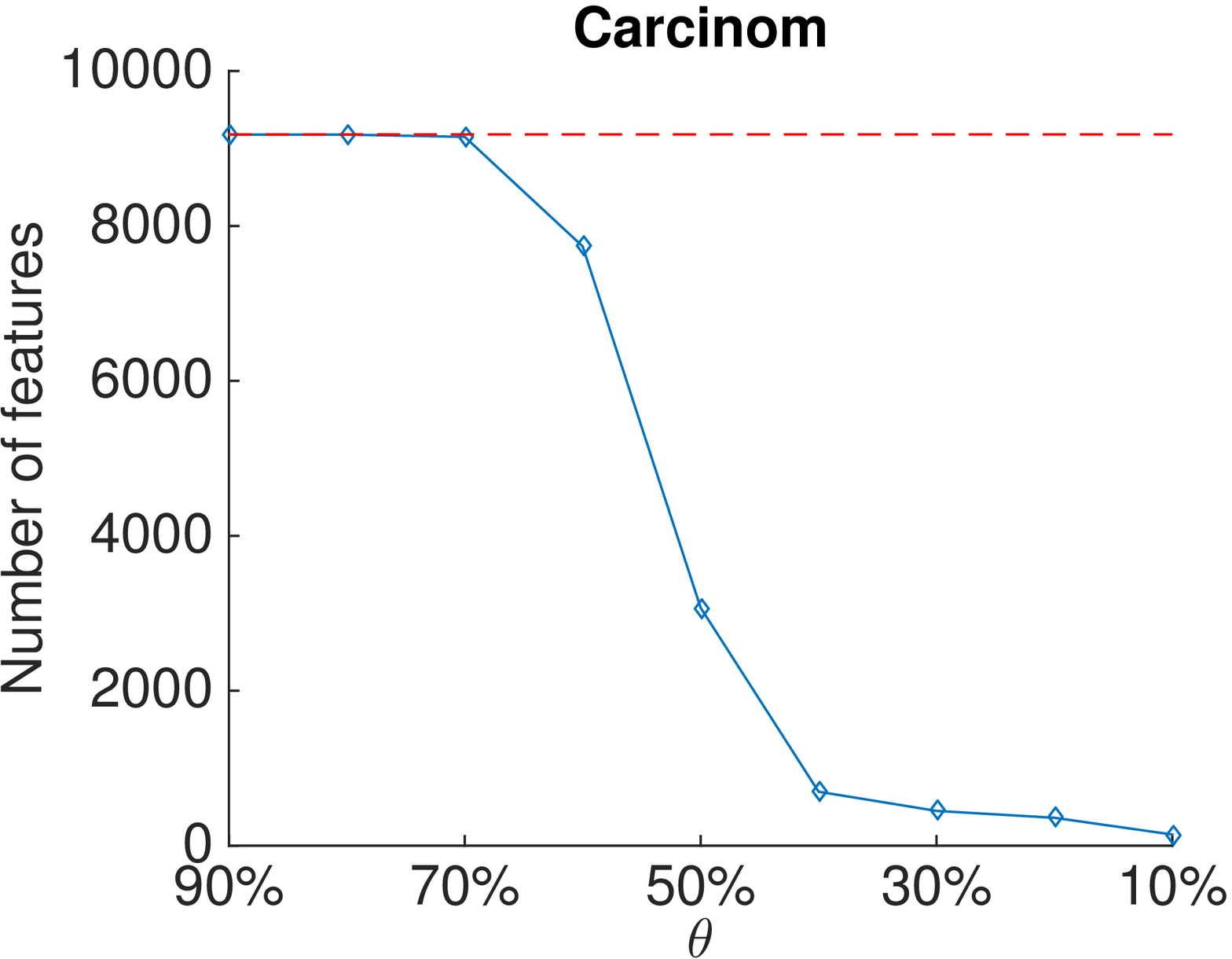}
	\includegraphics[width=0.225\textwidth]{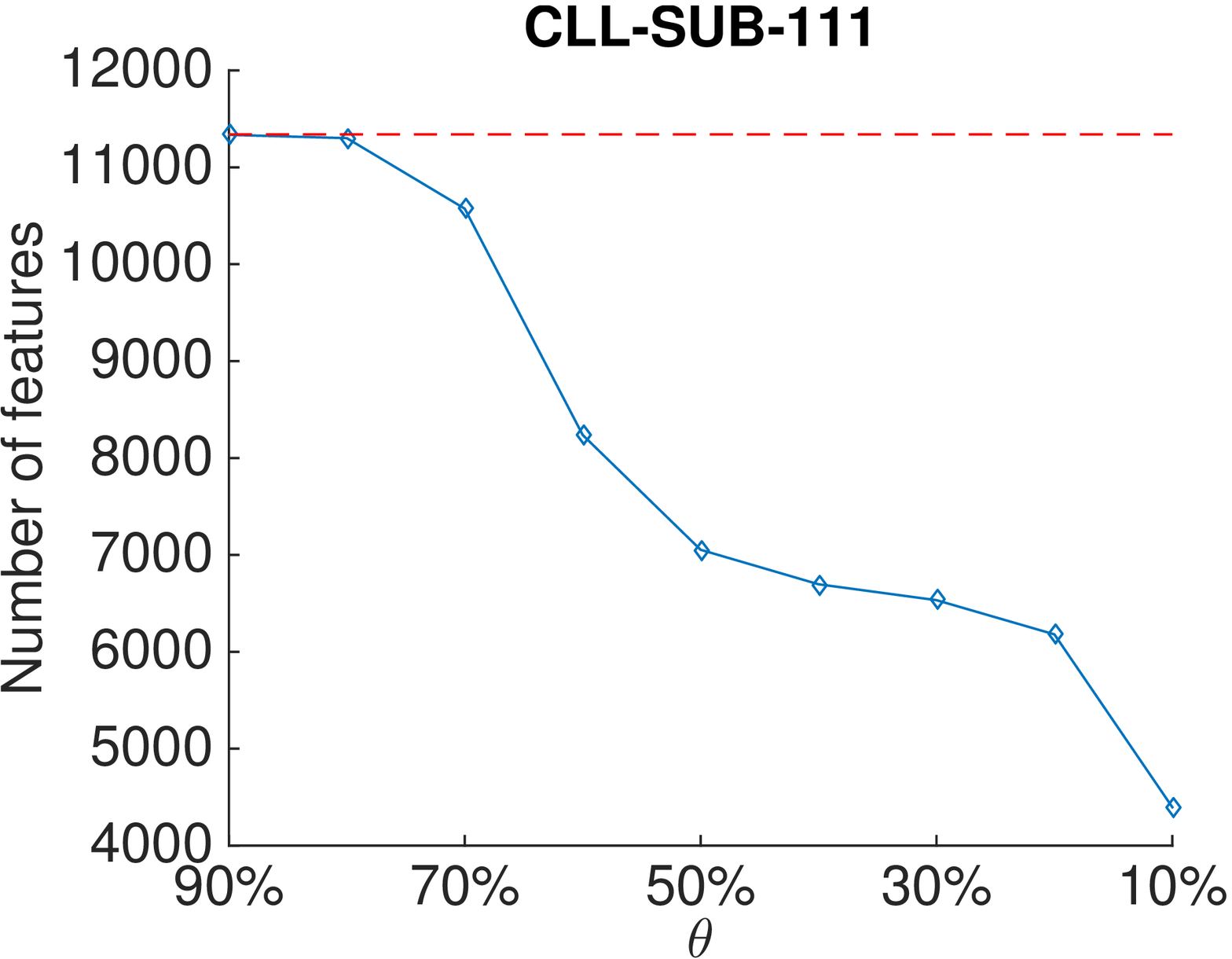} \\
	\caption{Spectral clustering performance of Biomedical datasets with different parameter $\theta$. Top row: NMI; Middle row: ACC; Bottom row: number of features, the red dash line means the size of raw dataset.}
	\label{fig:tkdd_sc_bio}
\end{figure}

We present our experiment results for image datasets, text datasets, and biological datasets in Figure~\ref{fig:tkdd_sc_image}, Figure~\ref{fig:tkdd_sc_text} and Figure~\ref{fig:tkdd_sc_bio} respectively. For each dataset, we show the NMI, ACC performance with different $\theta$ and comparing with original spectral clustering performance by using all features. From the experimental results, we can read that: \textbf{Even when $\theta$ is reduced to $30\%$, the NMI and ACC values are staying in same level as original data.} When $\theta$ equals to $30\%$, it means the edges of SFG that with weights (absolute value) in the highest $70\%$ value range are removed. (It does not mean that $70\%$ of top weights edges are removed). This observation validate the correctness of our proposed algorithm.   

\subsection{Performance of MCFS}
Our proposed algorithm is targeting for unsupervised feature selection. And the quality of indication vectors (or the spectral clustering performance based on eigenvectors) is an important factor evaluate the effectiveness of our proposed algorithm. In this section, we evaluate the MCFS performance over the redundant feature removed data, and comparing with the raw data that without any feature removal. 

The spectral clustering performance is measured for different input data from original whole feature data to processed ones by our proposed algorithm with different $\theta$. We report the experiment results over image datasets and biological datasets in this section. For text datasets, the feature vectors of them are very sparse, and our eigen decomposition process are always failed and we only can collect partial results. For fair evaluation, we omit the experiment results of text datasets in this work. The result of MCFS performance shows from Table~\ref{tab:tkdd_orl} to Table~\ref{tab:tkdd_cll}.

For each dataset, we set the number of selected features ranging from $[5,10,15,\cdots,60]$, which has 11 different sizes in total. The parameter $\theta$ is configured from $0.9$ to $0.1$ with stepsize equals to $0.1$. 

We report the experimental results in tables (from Table~\ref{tab:tkdd_orl} to Table~\ref{tab:tkdd_cll}). For each table, the first row means the number of features that used as input of MCFS. The first column is the number of selected features by MCFS algorithm. The baseline is in the second column, which is the testing result of MCFS algorithm with raw data. The hyphens in the tables means the number of selected features is larger than the feature size of input data, which means invalid test. To show the effectiveness of our algorithm, we also mark those NMI and ACC scores that larger or equals to baseline in bold text.
\begin{table}[!ht]
 	\centering
 	\setlength\tabcolsep{4pt}
 	\begin{minipage}{0.49\textwidth}
 		\centering
 	\resizebox{\textwidth}{!}{%
 	\begin{tabular}{|l||c||*{9}c|}
 		\hline 
 \#$f$ & 1024 & 913 & 620 & 535 & 469 & 327 & 160 & 104 & 58 & 33 \\ \hline \hline
 10 & 0.63 & 0.51 & 0.60 & 0.56 & 0.53 & 0.62 & 0.61 & \textbf{0.65} & 0.60 & 0.62 \\
 15 & 0.66 & 0.56 & 0.63 & 0.60 & 0.58 & \textbf{0.67 }& 0.62 & 0.60 & 0.63 & 0.58 \\
 20 & 0.67 & 0.59 & 0.65 & 0.64 & 0.59 & 0.64 & 0.63 & 0.61 & 0.64 & 0.56 \\
 25 & 0.67 & 0.59 & 0.66 & 0.64 & 0.63 & 0.65 & 0.66 & 0.64 & 0.65 & 0.58 \\
 30 & 0.68 & 0.63 & 0.66 & 0.65 & 0.66 & 0.67 & 0.65 & 0.67 & 0.65 & 0.59 \\
 35 & 0.69 & 0.64 & \textbf{0.70} & 0.66 & 0.65 & 0.67 & 0.67 & 0.68 & 0.65 & - \\
 40 & 0.70 & 0.67 & \textbf{0.71} & 0.68 & 0.67 & 0.68 & \textbf{0.70} & \textbf{0.70 }& 0.66 & - \\
 45 & 0.70 & 0.69 & \textbf{0.70} & 0.69 & 0.66 & 0.69 & \textbf{0.70} & 0.69 & 0.65 & - \\
 50 & 0.73 & 0.71 & 0.72 & 0.68 & 0.66 & 0.70 & 0.72 & 0.69 & 0.66 & - \\
 55 & 0.71 &\textbf{ 0.74} & 0.70 & 0.68 & 0.67 & 0.71 & 0.71 & 0.71 & 0.66 & - \\
 60 & 0.71 & \textbf{0.74} & \textbf{0.71} & \textbf{0.72} & \textbf{0.71} & 0.69 & \textbf{0.72 }& 0.71 & - & - \\
\hline
 	\end{tabular}}%
 	\caption{NMI results of ``ORL" dataset}%
 	\label{tab:tkdd_orl}
\end{minipage}%
 \hfill
 \begin{minipage}{0.49\textwidth}
 	\centering
 	\resizebox{\textwidth}{!}{%
 \begin{tabular}{|l||c||*{9}c|}
 	\hline 
 	\#$f$ & 1024 & 913 & 620 & 535 & 469 & 327 & 160 & 104 & 58 & 33 \\ \hline \hline
 10 & 0.38 & 0.28 & 0.36 & 0.31 & 0.28 & \textbf{0.39} & \textbf{0.39} & \textbf{0.46} & \textbf{0.39} & \textbf{0.41} \\
 15 & 0.45 & 0.33 & 0.41 & 0.40 & 0.34 & 0.43 & 0.40 & 0.38 & 0.42 & 0.36 \\
 20 & 0.47 & 0.34 & 0.43 & 0.43 & 0.35 & 0.43 & 0.41 & 0.39 & 0.43 & 0.32 \\
 25 & 0.48 & 0.35 & 0.45 & 0.44 & 0.37 & 0.42 & 0.47 & 0.41 & 0.45 & 0.34 \\
 30 & 0.47 & 0.40 & 0.42 & 0.42 & 0.43 & 0.47 & 0.43 & 0.45 & 0.42 & 0.35 \\
 35 & 0.49 & 0.41 & 0.48 & 0.46 & 0.44 & 0.44 & 0.47 & 0.47 & 0.42 & - \\
 40 & 0.51 & 0.46 & \textbf{0.53} & 0.48 & 0.46 & 0.45 & 0.48 & \textbf{0.51} & 0.43 & - \\
 45 & 0.49 & 0.47 & \textbf{0.51} & \textbf{0.51} & 0.44 & 0.48 & \textbf{0.49} & \textbf{0.49} & 0.43 & - \\
 50 & 0.55 & 0.51 & 0.52 & 0.47 & 0.47 & 0.50 & 0.52 & 0.48 & 0.46 & - \\
 55 & 0.53 & \textbf{0.53} & 0.51 & 0.46 & 0.45 & 0.48 & 0.50 & \textbf{0.53} & 0.46 & - \\
 60 & 0.51 & \textbf{0.55} & \textbf{0.52} & \textbf{0.54} & 0.51 & 0.47 & \textbf{0.54} & 0.51 & - & - \\
 	\hline
 \end{tabular}}%
 \caption{ACC results of ``ORL" dataset.}%
\end{minipage}%
\hfill
\begin{minipage}{0.49\textwidth}
 		\centering
 		\resizebox{\textwidth}{!}{%
 			\begin{tabular}{|l||c||*{9}c|}
 				\hline 
 				\#$f$ &1024 & 1023 & 964 & 654 & 525 & 427 & 271 & 152 & 83 & 34 \\ \hline \hline
 10 & 0.48 & 0.43 & 0.43 & 0.45 & 0.42 & 0.46 & 0.45 & 0.46 & 0.47 & 0.44 \\
 15 & 0.49 & 0.47 & 0.46 & \textbf{0.51} & \textbf{0.49} & 0.48 & 0.45 & 0.47 & \textbf{0.50} & 0.43 \\
 20 & 0.49 & 0.48 & 0.46 & \textbf{0.55} & 0.48 & \textbf{0.51} & 0.47 & 0.47 & \textbf{0.51} & 0.41 \\
 25 & 0.51 & 0.49 & 0.49 & \textbf{0.52} & \textbf{0.52} & \textbf{0.52} & 0.45 & 0.49 & \textbf{0.54} & 0.41 \\
 30 & 0.51 & \textbf{0.51} & 0.49 & \textbf{0.54} & 0.50 & \textbf{0.51} & \textbf{0.51} & 0.49 & \textbf{0.50} & 0.39 \\
 35 & 0.53 & 0.49 & 0.50 & \textbf{0.54} & \textbf{0.53} & 0.52 & 0.52 & 0.48 & 0.50 & - \\
 40 & 0.49 & \textbf{0.50} & \textbf{0.51} & \textbf{0.53} & \textbf{0.58} & \textbf{0.55} & \textbf{0.55} & 0.48 & \textbf{0.51} & - \\
 45 & 0.48 & \textbf{0.51} & \textbf{0.51} & \textbf{0.56} & \textbf{0.59} & \textbf{0.57} & \textbf{0.52} & \textbf{0.52} & \textbf{0.49} & - \\
 50 & 0.52 & 0.50 & 0.47 & \textbf{0.53} & \textbf{0.59} & \textbf{0.53} & \textbf{0.53} & \textbf{0.56} & 0.49 & - \\
 55 & 0.54 & 0.51 & 0.52 & \textbf{0.55} & 0.50 & 0.51 & 0.51 & 0.51 & 0.49 & - \\
 60 & 0.54 & 0.49 & 0.51 & 0.49 & \textbf{0.54} & 0.50 & 0.51 & 0.46 & 0.52 & - \\
 				\hline
 			\end{tabular}}%
 			\caption{NMI results of ``Yale" dataset}%
\end{minipage}%
 		\hfill
 		\begin{minipage}{0.49\textwidth}
 			\centering
 			\resizebox{\textwidth}{!}{%
 				\begin{tabular}{|l||c||*{9}c|}
 					\hline 
 					\#$f$ & 1024 & 1023 & 964 & 654 & 525 & 427 & 271 & 152 & 83 & 34 \\ \hline \hline
  10 & 0.39 & 0.36 & 0.37 & 0.36 & 0.33 & 0.38 & \textbf{0.41} & \textbf{0.40} & \textbf{0.41} & 0.36 \\
  15 & 0.43 & 0.41 & 0.42 & \textbf{0.44} & 0.41 & 0.41 & 0.39 & 0.41 & \textbf{0.46} & 0.39 \\
  20 & 0.44 & 0.42 & 0.41 & \textbf{0.48} & \textbf{0.44} & \textbf{0.44} & 0.43 & 0.42 & \textbf{0.44} & 0.35 \\
  25 & 0.45 & \textbf{0.45} & 0.44 & \textbf{0.46} & \textbf{0.47} & \textbf{0.45} & 0.41 & 0.43 & \textbf{0.49} & 0.33 \\
  30 & 0.48 & 0.44 & 0.42 & 0.47 & 0.47 & 0.45 & 0.45 & 0.40 & 0.47 & 0.33 \\
  35 & 0.48 & \textbf{0.48} & 0.44 & \textbf{0.50} & 0.47 & 0.46 & 0.47 & 0.41 & 0.44 & - \\
  40 & 0.42 & \textbf{0.44} & \textbf{0.45} & \textbf{0.50} & \textbf{0.55} & \textbf{0.48} & \textbf{0.53} & 0.41 & \textbf{0.44} & - \\
  45 & 0.41 & \textbf{0.48} & \textbf{0.46} & \textbf{0.51} & \textbf{0.53} & \textbf{0.54} & \textbf{0.49} & \textbf{0.47} & \textbf{0.42} & - \\
  50 & 0.46 & 0.41 & 0.42 & \textbf{0.48} & \textbf{0.56} & \textbf{0.50} & \textbf{0.46} & \textbf{0.52} & 0.41 & - \\
  55 & 0.48 & 0.44 & \textbf{0.48} & \textbf{0.48} & 0.43 & 0.45 & \textbf{0.49} & 0.47 & 0.42 & - \\
  60 & 0.50 & 0.42 & 0.44 & 0.40 & \textbf{0.50} & 0.41 & 0.46 & 0.42 & 0.43 & - \\
 					\hline
 				\end{tabular}}%
 				\caption{ACC results of ``Yale" dataset.}%
\end{minipage}%
\hfill
\begin{minipage}{0.49\textwidth}
	\centering
	\resizebox{\textwidth}{!}{%
		\begin{tabular}{|l||c||*{9}c|}
			\hline 
			\#$f$ & 2420 & 2409 & 1871 & 793 & 698 & 662 & 654 & 630 & 566 & 324 \\
			 \hline \hline
 10 & 0.44 & \textbf{0.48} & \textbf{0.55} & \textbf{0.53} & \textbf{0.58} & \textbf{0.56} & \textbf{0.54} & \textbf{0.61} & \textbf{0.50} & 0.38 \\
 15 & 0.44 & \textbf{0.61} & \textbf{0.57} & \textbf{0.50} & \textbf{0.58} & \textbf{0.58} & \textbf{0.55} & \textbf{0.59} & \textbf{0.53} & 0.39 \\
 20 & 0.43 & \textbf{0.56} & \textbf{0.61} & \textbf{0.59} & \textbf{0.60} & \textbf{0.56} & \textbf{0.62} & \textbf{0.59} & \textbf{0.56} & 0.41 \\
 25 & 0.52 & \textbf{0.61} & \textbf{0.61} & \textbf{0.64} & \textbf{0.61} & \textbf{0.60} & \textbf{0.58} & \textbf{0.58} & \textbf{0.54} & \textbf{0.43} \\
 30 & 0.53 & \textbf{0.61} & \textbf{0.62} & \textbf{0.57} & \textbf{0.62} & \textbf{0.62} & \textbf{0.60} & \textbf{0.53} & \textbf{0.63} & 0.41 \\
 35 & 0.59 & \textbf{0.60} & \textbf{0.59} & \textbf{0.60} & \textbf{0.63} & \textbf{0.61} & \textbf{0.60} & \textbf{0.62} & \textbf{0.64} & 0.43 \\
 40 & 0.53 & \textbf{0.60} & \textbf{0.58} & \textbf{0.57} & \textbf{0.66} & \textbf{0.62} & \textbf{0.59} & \textbf{0.62} & \textbf{0.69} & 0.42 \\
 45 & 0.55 & \textbf{0.61} & \textbf{0.61} & \textbf{0.62} & \textbf{0.60} & \textbf{0.64} & 0.60 & \textbf{0.64} & \textbf{0.65} & 0.43 \\
 50 & 0.56 & \textbf{0.63} & \textbf{0.62} & \textbf{0.68} & \textbf{0.64} & \textbf{0.62} & \textbf{0.58} & \textbf{0.63} & \textbf{0.66} & 0.37 \\
 55 & 0.61 & 0.60 & \textbf{0.62} & \textbf{0.69} & \textbf{0.62} & 0.60 & 0.57 & \textbf{0.65} & 0.58 & 0.39 \\
 60 & 0.55 & \textbf{0.64} & \textbf{0.63} & \textbf{0.64} & \textbf{0.60} & \textbf{0.63} & 0.54 & \textbf{0.63} & 0.51 & 0.39 \\
			\hline
		\end{tabular}}%
		\caption{NMI results of ``PIE10P" dataset}%
	\end{minipage}%
	\hfill
	\begin{minipage}{0.49\textwidth}
		\centering
		\resizebox{\textwidth}{!}{%
			\begin{tabular}{|l||c||*{9}c|}
				\hline 
				\#$f$ & 2420 & 2409 & 1871 & 793 & 698 & 662 & 654 & 630 & 566 & 324 \\ \hline \hline
 10 & 0.39 & \textbf{0.45} & \textbf{0.48} & \textbf{0.50} & \textbf{0.56} & \textbf{0.50} & \textbf{0.53} & \textbf{0.59} & \textbf{0.46} & 0.39 \\
 15 & 0.39 & \textbf{0.58} & \textbf{0.51} & \textbf{0.49} & \textbf{0.51} & \textbf{0.55} & \textbf{0.56} & \textbf{0.60} & \textbf{0.50} & \textbf{0.41} \\
 20 & 0.36 & \textbf{0.51} & \textbf{0.53} & \textbf{0.53} & \textbf{0.55} & \textbf{0.56} & \textbf{0.60} & \textbf{0.54} & \textbf{0.50} & \textbf{0.38} \\
 25 & 0.45 & \textbf{0.59} & \textbf{0.53} & \textbf{0.60} & \textbf{0.54} & \textbf{0.59} & \textbf{0.60} & \textbf{0.56} & \textbf{0.52} & 0.40 \\
 30 & 0.50 & \textbf{0.58} & \textbf{0.56} & \textbf{0.58} & \textbf{0.59} & \textbf{0.60} & \textbf{0.59} & \textbf{0.49} & \textbf{0.60} & 0.40 \\
 35 & 0.48 & \textbf{0.57} & \textbf{0.51} & \textbf{0.59} & \textbf{0.61} & \textbf{0.53} & \textbf{0.54} & \textbf{0.62} & \textbf{0.61} & \textbf{0.37} \\
 40 & 0.42 & \textbf{0.52} & \textbf{0.53} & \textbf{0.56} & \textbf{0.63} & \textbf{0.59} & \textbf{0.53} & \textbf{0.60} & \textbf{0.64} & 0.38 \\
 45 & 0.44 & \textbf{0.52} & \textbf{0.52} & \textbf{0.58} & \textbf{0.51} & \textbf{0.63} & \textbf{0.54} & \textbf{0.62} & \textbf{0.60} & \textbf{0.41} \\
 50 & 0.44 & \textbf{0.61} & \textbf{0.52} & \textbf{0.64} & \textbf{0.60} & \textbf{0.59} & \textbf{0.55} & \textbf{0.62} & \textbf{0.61} & 0.37 \\
 55 & 0.46 & \textbf{0.54} & \textbf{0.53} & \textbf{0.67} & \textbf{0.58} & \textbf{0.57} & \textbf{0.57} & \textbf{0.63} & \textbf{0.54} & 0.37 \\
 60 & 0.49 & \textbf{0.60} & \textbf{0.61} & \textbf{0.61} & \textbf{0.57} & \textbf{0.61} & \textbf{0.51} & \textbf{0.61} & \textbf{0.46} & 0.35 \\
				\hline
			\end{tabular}}%
			\caption{ACC results of ``PIE10P" dataset.}%
		\end{minipage}%
\hfill
\begin{minipage}{0.49\textwidth}
	\centering
	\resizebox{\textwidth}{!}{%
		\begin{tabular}{|l||c||*{9}c|}
			\hline 
\#$f$ & 10304 & 10302 & 8503 & 3803 & 3408 & 3244 & 3030 & 2822 & 2638 & 2175 \\
			\hline \hline
 10 & 0.65 & \textbf{0.78} & \textbf{0.77} & \textbf{0.76} & \textbf{0.77} & \textbf{0.80} & \textbf{0.74} & \textbf{0.72} & \textbf{0.75} & \textbf{0.73} \\
 15 & 0.72 & \textbf{0.82} & \textbf{0.79} & \textbf{0.78} & \textbf{0.81} & \textbf{0.83} & \textbf{0.79} & \textbf{0.81} & \textbf{0.75} & \textbf{0.79} \\
 20 & 0.76 & \textbf{0.81} & 0.74 & \textbf{0.78} & \textbf{0.84} & \textbf{0.83} & \textbf{0.81} & \textbf{0.76} & \textbf{0.80} & \textbf{0.78} \\
 25 & 0.79 & \textbf{0.84} & 0.74 & 0.73 & \textbf{0.82} & \textbf{0.86} & \textbf{0.88} & \textbf{0.83} & \textbf{0.86} & \textbf{0.81} \\
 30 & 0.75 & \textbf{0.77} & \textbf{0.82} & 0.74 & \textbf{0.88} & \textbf{0.82} & \textbf{0.83} & \textbf{0.83} & \textbf{0.86} & \textbf{0.86} \\
 35 & 0.81 & \textbf{0.81} & 0.80 & \textbf{0.83} & \textbf{0.85} & \textbf{0.83} & 0.80 & \textbf{0.82} & \textbf{0.85} & \textbf{0.85} \\
 40 & 0.83 & \textbf{0.88} & \textbf{0.84} & \textbf{0.84} & \textbf{0.90} & \textbf{0.86} & 0.81 & \textbf{0.93} & \textbf{0.84} & \textbf{0.87} \\
 45 & 0.84 & 0.93 & 0.83 & \textbf{0.85} & \textbf{0.91} & \textbf{0.86} & 0.83 & \textbf{0.88} & \textbf{0.84} & \textbf{0.86} \\
 50 & 0.78 & \textbf{0.88} & \textbf{0.88} & \textbf{0.87} & \textbf{0.89} & \textbf{0.86} & \textbf{0.82} & \textbf{0.90} & \textbf{0.84} & \textbf{0.83} \\
 55 & 0.84 & \textbf{0.89} & \textbf{0.86} & \textbf{0.89} & \textbf{0.91} & \textbf{0.89} & \textbf{0.88} & \textbf{0.86} & \textbf{0.84} & \textbf{0.86} \\
 60 & 0.85 & \textbf{0.88} & \textbf{0.86} & 0.84 & \textbf{0.85} & \textbf{0.91} & \textbf{0.85} & \textbf{0.88} & \textbf{0.86} & \textbf{0.85} \\
			\hline
		\end{tabular}}%
		\caption{NMI results of ``ORL10P" dataset}%
	\end{minipage}%
	\hfill
	\begin{minipage}{0.49\textwidth}
		\centering
		\resizebox{\textwidth}{!}{%
			\begin{tabular}{|l||c||*{9}c|}
				\hline 
\#$f$ & 10304 & 10302 & 8503 & 3803 & 3408 & 3244 & 3030 & 2822 & 2638 & 2175 \\ \hline \hline
 10 & 0.66 & \textbf{0.74} & \textbf{0.81} & \textbf{0.75} & \textbf{0.75} & \textbf{0.69} & \textbf{0.72} & \textbf{0.70} & \textbf{0.69} & \textbf{0.67} \\
 15 & 0.69 & \textbf{0.85} & \textbf{0.76} & \textbf{0.78} & \textbf{0.78} & \textbf{0.86} & \textbf{0.80} & \textbf{0.75} & \textbf{0.73} & \textbf{0.75} \\
 20 & 0.77 & \textbf{0.84} & 0.74 & 0.76 & \textbf{0.80} & \textbf{0.80} & \textbf{0.78} & 0.69 & 0.75 & 0.74 \\
 25 & 0.71 & \textbf{0.79} & 0.68 & \textbf{0.74} & \textbf{0.78} & \textbf{0.86} & \textbf{0.84} & \textbf{0.82} & \textbf{0.82} & \textbf{0.74} \\
 30 & 0.71 & \textbf{0.71} & \textbf{0.77} & 0.68 & \textbf{0.86} & \textbf{0.77} & \textbf{0.81} & \textbf{0.77} & \textbf{0.82} & \textbf{0.81} \\
 35 & 0.74 & \textbf{0.74} & \textbf{0.74} & \textbf{0.76} & \textbf{0.81} & \textbf{0.77} & 0.73 & \textbf{0.76} & \textbf{0.82} & \textbf{0.78} \\
 40 & 0.80 & \textbf{0.85} & 0.74 & 0.77 & \textbf{0.87} & \textbf{0.80} & 0.75 & \textbf{0.89} & \textbf{0.80} & \textbf{0.83} \\
 45 & 0.82 & \textbf{0.89} & 0.73 & 0.81 & \textbf{0.88} & 0.78 & 0.77 & \textbf{0.86} & 0.80 & 0.79 \\
 50 & 0.73 & \textbf{0.80} & \textbf{0.80} & \textbf{0.74} & \textbf{0.86} & \textbf{0.79} & \textbf{0.74} & \textbf{0.88} & \textbf{0.81} & \textbf{0.77} \\
 55 & 0.79 & \textbf{0.85} & \textbf{0.82} & \textbf{0.86} & \textbf{0.89} & \textbf{0.87} & \textbf{0.80} & \textbf{0.82} & \textbf{0.81} & \textbf{0.79} \\
 60 & 0.82 & \textbf{0.84} & 0.77 & 0.75 & \textbf{0.82} & \textbf{0.89} & 0.77 & \textbf{0.84} & \textbf{0.82} & \textbf{0.82} \\
				\hline
			\end{tabular}}%
			\caption{ACC results of ``ORL10P" dataset.}%
		\end{minipage}%
\end{table}
 
\begin{table}[!ht]
 	\centering
 	\setlength\tabcolsep{4pt}
 	\begin{minipage}{0.49\textwidth}
 		\centering
 	\resizebox{\textwidth}{!}{%
 	\begin{tabular}{|l||c||*{9}c|}
 		\hline 
 \#$f$ & 4026 & 4009 & 3978 & 3899 & 3737 & 3456 & 2671 & 1203 & 334 & 136 \\ \hline \hline
 10 & 0.51 & \textbf{0.59} & \textbf{0.58} & \textbf{0.52} & 0.50 & 0.50 & \textbf{0.51} & 0.50 & 0.50 & 0.49 \\ 
 15 & 0.55 & \textbf{0.60} & \textbf{0.62} & \textbf{0.56} & \textbf{0.58} & \textbf{0.58} & \textbf{0.58} & \textbf{0.56} & 0.47 & 0.52 \\ 
 20 & 0.60 & \textbf{0.61} & \textbf{0.60} & 0.57 & \textbf{0.62} & \textbf{0.62} & \textbf{0.64} & 0.58 & 0.58 & \textbf{0.60} \\ 
 25 & 0.63 & 0.59 & \textbf{0.64} & 0.60 & \textbf{0.63} & 0.58 & \textbf{0.66} & 0.57 & 0.56 & 0.53 \\ 
 30 & 0.59 & \textbf{0.61} & \textbf{0.62} & \textbf{0.60} & \textbf{0.62} & \textbf{0.64} & \textbf{0.65} & \textbf{0.60} & \textbf{0.60} & \textbf{0.59} \\ 
 35 & 0.61 & \textbf{0.66} & \textbf{0.62} & 0.60 & \textbf{0.65} & \textbf{0.62} & \textbf{0.61} & \textbf{0.62} & 0.56 & 0.53 \\ 
 40 & 0.64 & 0.60 & \textbf{0.66} & 0.63 & 0.61 & 0.63 & \textbf{0.66} & 0.61 & 0.58 & 0.55 \\ 
 45 & 0.58 & \textbf{0.63} & \textbf{0.62} & \textbf{0.62} & \textbf{0.58} & \textbf{0.61} & \textbf{0.63} & \textbf{0.64} & \textbf{0.60} & 0.57 \\ 
 50 & 0.65 & 0.60 & 0.61 & 0.61 & 0.56 & 0.63 & 0.61 & 0.63 & 0.58 & 0.54 \\ 
 55 & 0.63 & 0.60 & 0.61 & 0.62 & 0.60 & 0.60 & \textbf{0.63} & 0.60 & 0.58 & 0.58 \\ 
 60 & 0.60 & \textbf{0.60} & \textbf{0.63} & \textbf{0.61} & \textbf{0.63} & 0.59 & \textbf{0.65} & 0.59 & 0.57 & 0.57 \\ 
\hline
 	\end{tabular}}%
 	\caption{NMI results of ``Lymphoma" dataset}%
\end{minipage}%
 \hfill
 \begin{minipage}{0.49\textwidth}
 	\centering
 	\resizebox{\textwidth}{!}{%
 \begin{tabular}{|l||c||*{9}c|}
 	\hline 
 	\#$f$ &  4026 & 4009 & 3978 & 3899 & 3737 & 3456 & 2671 & 1203 & 334 & 136 \\ \hline \hline
 10 & 0.50 & \textbf{0.57} & \textbf{0.56} & \textbf{0.53} & 0.49 & \textbf{0.51} & \textbf{0.51} & 0.48 & \textbf{0.50} & \textbf{0.50} \\ 
 15 & 0.53 & \textbf{0.62} & \textbf{0.59} & \textbf{0.58} & \textbf{0.56} & \textbf{0.59} & \textbf{0.58} & \textbf{0.55} & 0.50 & \textbf{0.53} \\ 
 20 & 0.59 & 0.56 & 0.55 & 0.56 & 0.56 & \textbf{0.59} & \textbf{0.59} & 0.54 & 0.55 & \textbf{0.59} \\ 
 25 & 0.60 & 0.57 & \textbf{0.62} & 0.56 & \textbf{0.62} & 0.58 & \textbf{0.64} & 0.56 & 0.52 & 0.50 \\ 
 30 & 0.56 & \textbf{0.60} & \textbf{0.58} & \textbf{0.58} & \textbf{0.59} & \textbf{0.61} & \textbf{0.65} & \textbf{0.59} & \textbf{0.57} & 0.55 \\ 
 35 & 0.55 & \textbf{0.62} & \textbf{0.59} & \textbf{0.58} & \textbf{0.61} & \textbf{0.60} & \textbf{0.57} & \textbf{0.59} & \textbf{0.55} & 0.53 \\ 
 40 & 0.66 & 0.57 & 0.61 & 0.61 & 0.61 & 0.59 & 0.60 & 0.58 & 0.59 & 0.54 \\ 
 45 & 0.54 & \textbf{0.60} & \textbf{0.60} & \textbf{0.58} & \textbf{0.55} & \textbf{0.60} & \textbf{0.62} & \textbf{0.59} & \textbf{0.56} & \textbf{0.54} \\ 
 50 & 0.65 & 0.62 & 0.58 & 0.64 & 0.52 & 0.59 & 0.56 & 0.59 & 0.53 & 0.53 \\ 
 55 & 0.57 & \textbf{0.60} & \textbf{0.65} & \textbf{0.60} & 0.54 & \textbf{0.57} & \textbf{0.65} & \textbf{0.59} & 0.54 & \textbf{0.59} \\ 
 60 & 0.56 & \textbf{0.58} & \textbf{0.64} & \textbf{0.58} & \textbf{0.61} & \textbf{0.57} & \textbf{0.67} & \textbf{0.56} & 0.53 & \textbf{0.57} \\ 
 	\hline
 \end{tabular}}%
 \caption{ACC results of ``Lymphoma" dataset.}%
\end{minipage}%
\hfill
\begin{minipage}{0.49\textwidth}
 		\centering
 		\resizebox{\textwidth}{!}{%
 			\begin{tabular}{|l||c||*{9}c|}
 				\hline 
 				\#$f$ & 3312 & 3311 & 3309 & 3236 & 1844 & 559 & 384 & 344 & 305 & 183 \\ \hline \hline
10 & 0.42 & \textbf{0.42} & \textbf{0.43} & \textbf{0.49} & \textbf{0.52} & \textbf{0.53} & \textbf{0.43} & \textbf{0.46} & \textbf{0.43} & 0.25 \\ 
15 & 0.54 & \textbf{0.54} & 0.53 & 0.51 & 0.51 & 0.51 & 0.45 & 0.52 & 0.38 & 0.21 \\ 
20 & 0.51 & \textbf{0.51} & \textbf{0.52} & \textbf{0.53} & 0.41 & 0.49 & 0.36 & \textbf{0.52} & 0.38 & 0.20 \\ 
25 & 0.51 & \textbf{0.51} & \textbf{0.53} & 0.48 & 0.42 & \textbf{0.52} & 0.40 & 0.48 & 0.35 & 0.26 \\ 
30 & 0.47 & \textbf{0.48} & \textbf{0.52} & \textbf{0.49} & 0.41 & 0.37 & \textbf{0.49} & \textbf{0.48} & 0.41 & 0.24 \\ 
35 & 0.46 & 0.38 & \textbf{0.46} & \textbf{0.48} & 0.39 & \textbf{0.52} & \textbf{0.49} & 0.38 & 0.35 & 0.27 \\ 
40 & 0.49 & \textbf{0.49} & \textbf{0.50} & 0.46 & 0.43 & 0.40 & 0.38 & 0.35 & 0.40 & 0.29 \\ 
45 & 0.36 & \textbf{0.42} & 0.33 & \textbf{0.47} & \textbf{0.40} & 0.33 & \textbf{0.38} & 0.35 & 0.35 & 0.31 \\ 
50 & 0.45 & \textbf{0.45} & \textbf{0.47} & \textbf{0.49} & \textbf{0.52} & 0.32 & 0.40 & 0.36 & 0.35 & 0.31 \\ 
55 & 0.44 & \textbf{0.44} & \textbf{0.44} & \textbf{0.49} & \textbf{0.51} & 0.33 & \textbf{0.49} & 0.31 & 0.30 & 0.31 \\ 
60 & 0.47 & 0.46 & 0.45 & \textbf{0.51} & \textbf{0.49} & 0.33 & 0.39 & 0.32 & 0.31 & 0.35 \\ 
 				\hline
 			\end{tabular}}%
 			\caption{NMI results of ``LUNG" dataset}%
\end{minipage}%
 		\hfill
 		\begin{minipage}{0.49\textwidth}
 			\centering
 			\resizebox{\textwidth}{!}{%
 				\begin{tabular}{|l||c||*{9}c|}
 					\hline 
 					\#$f$ & 3312 & 3311 & 3309 & 3236 & 1844 & 559 & 384 & 344 & 305 & 183 \\ \hline \hline
10 & 0.71 & \textbf{0.72} & \textbf{0.73} & \textbf{0.77} & \textbf{0.77} & \textbf{0.75} & 0.68 & 0.65 & 0.66 & 0.56 \\ 
15 & 0.81 & \textbf{0.81} & 0.79 & 0.72 & 0.73 & 0.72 & 0.67 & 0.65 & 0.58 & 0.48 \\ 
20 & 0.71 & \textbf{0.73} & \textbf{0.74} & \textbf{0.72} & 0.69 & 0.69 & 0.61 & 0.60 & 0.58 & 0.39 \\ 
25 & 0.71 & \textbf{0.71} & \textbf{0.74} & 0.67 & 0.69 & 0.68 & 0.59 & 0.61 & 0.56 & 0.49 \\ 
30 & 0.66 & \textbf{0.66} & \textbf{0.67} & \textbf{0.71} & \textbf{0.68} & 0.56 & 0.59 & 0.59 & 0.61 & 0.43 \\ 
35 & 0.64 & 0.60 & 0.63 & \textbf{0.68} & \textbf{0.66} & 0.60 & 0.58 & 0.56 & 0.53 & 0.49 \\ 
40 & 0.65 & \textbf{0.65} & \textbf{0.66} & \textbf{0.65} & 0.64 & 0.57 & 0.54 & 0.54 & 0.56 & 0.46 \\ 
45 & 0.60 & \textbf{0.63} & 0.57 & \textbf{0.65} & \textbf{0.61} & 0.52 & 0.54 & 0.52 & 0.52 & 0.49 \\ 
50 & 0.65 & \textbf{0.65} & 0.63 & \textbf{0.65} & \textbf{0.65} & 0.48 & 0.57 & 0.53 & 0.53 & 0.52 \\ 
55 & 0.61 & \textbf{0.61} & 0.59 & \textbf{0.65} & \textbf{0.62} & 0.48 & 0.59 & 0.48 & 0.49 & 0.49 \\ 
60 & 0.64 & 0.63 & 0.63 & \textbf{0.64} & 0.62 & 0.51 & 0.55 & 0.49 & 0.48 & 0.51 \\ 
 					\hline
 				\end{tabular}}%
 				\caption{ACC results of ``LUNG" dataset.}%
\end{minipage}%
\hfill
\begin{minipage}{0.49\textwidth}
	\centering
	\resizebox{\textwidth}{!}{%
		\begin{tabular}{|l||c||*{9}c|}
			\hline 
		\#$f$ &  9182 & 9180 & 9179 & 9150 & 7736 & 3072 & 697 & 449 & 360 & 144 \\
			 \hline \hline
10 & 0.70 & \textbf{0.70} & \textbf{0.70} & 0.69 & 0.67 & 0.64 & 0.66 & 0.65 & 0.66 & 0.47 \\ 
15 & 0.71 & 0.70 & \textbf{0.73} & \textbf{0.73} & \textbf{0.74} & 0.66 & 0.67 & 0.70 & 0.66 & 0.52 \\ 
20 & 0.77 & \textbf{0.78} & \textbf{0.77} & 0.72 & 0.75 & 0.72 & 0.73 & 0.71 & 0.73 & 0.54 \\ 
25 & 0.74 & \textbf{0.77} & \textbf{0.77} & \textbf{0.75} & \textbf{0.74} & 0.71 & \textbf{0.79} & \textbf{0.75} & \textbf{0.74} & 0.53 \\ 
30 & 0.69 & \textbf{0.71} & \textbf{0.72} & \textbf{0.70} & \textbf{0.74} & \textbf{0.75} & \textbf{0.77} & \textbf{0.79} & \textbf{0.73} & 0.54 \\ 
35 & 0.77 & 0.76 & 0.76 & 0.76 & 0.74 & \textbf{0.77} & \textbf{0.78} & \textbf{0.78} & \textbf{0.78} & 0.60 \\ 
40 & 0.75 & 0.74 & \textbf{0.76} & \textbf{0.77} & 0.74 & \textbf{0.79} & \textbf{0.76} & \textbf{0.78} & \textbf{0.75} & 0.59 \\ 
45 & 0.77 & 0.76 & 0.74 & \textbf{0.78} & 0.74 & \textbf{0.82} & \textbf{0.78} & \textbf{0.80} & \textbf{0.79} & 0.57 \\ 
50 & 0.79 & 0.76 & 0.75 & 0.75 & \textbf{0.79} & 0.76 & \textbf{0.79} & \textbf{0.84} & \textbf{0.83} & 0.58 \\ 
55 & 0.75 & \textbf{0.76} & \textbf{0.76} & 0.74 & \textbf{0.75} & \textbf{0.79} & \textbf{0.79} & \textbf{0.83} & \textbf{0.83} & 0.59 \\ 
60 & 0.74 & 0.72 & \textbf{0.76} & 0.73 & \textbf{0.76} & \textbf{0.82} & \textbf{0.84} & \textbf{0.82} & \textbf{0.78} & 0.62 \\ 
			\hline
		\end{tabular}}%
		\caption{NMI results of ``Carcinom" dataset}%
	\end{minipage}%
	\hfill
	\begin{minipage}{0.49\textwidth}
		\centering
		\resizebox{\textwidth}{!}{%
			\begin{tabular}{|l||c||*{9}c|}
				\hline 
\#$f$ & 9182 & 9180 & 9179 & 9150 & 7736 & 3072 & 697 & 449 & 360 & 144 \\ \hline \hline
10 & 0.63 & \textbf{0.66} & 0.62 & 0.61 & \textbf{0.67} & 0.60 & 0.60 & 0.59 & \textbf{0.64} & 0.48 \\ 
15 & 0.67 & 0.57 & \textbf{0.70} & 0.66 & \textbf{0.68} & 0.63 & 0.57 & \textbf{0.67} & 0.64 & 0.53 \\ 
20 & 0.70 & 0.68 & \textbf{0.74} & 0.66 & \textbf{0.71} & \textbf{0.71} & 0.64 & \textbf{0.73} & \textbf{0.74} & 0.56 \\ 
25 & 0.70 & \textbf{0.72} & \textbf{0.75} & 0.69 & \textbf{0.75} & 0.64 & \textbf{0.75} & \textbf{0.72} & \textbf{0.76} & 0.51 \\ 
30 & 0.61 & \textbf{0.64} & \textbf{0.70} & \textbf{0.69} & \textbf{0.67} & \textbf{0.71} & \textbf{0.74} & \textbf{0.76} & \textbf{0.71} & 0.52 \\ 
35 & 0.76 & 0.74 & 0.74 & 0.74 & 0.70 & 0.75 & 0.70 & \textbf{0.76} & \textbf{0.77} & 0.57 \\ 
40 & 0.72 & \textbf{0.72} & \textbf{0.73} & \textbf{0.75} & 0.69 & \textbf{0.76} & 0.66 & \textbf{0.78} & 0.71 & 0.56 \\ 
45 & 0.75 & 0.74 & 0.70 & \textbf{0.75} & 0.74 & \textbf{0.79} & 0.72 & \textbf{0.79} & \textbf{0.76} & 0.55 \\ 
50 & 0.74 & \textbf{0.74} & 0.70 & 0.72 & \textbf{0.74} & 0.66 & \textbf{0.74} & \textbf{0.83} & \textbf{0.79} & 0.56 \\ 
55 & 0.73 & \textbf{0.74} & \textbf{0.74} & 0.72 & 0.71 & 0.72 & 0.72 & \textbf{0.82} & \textbf{0.80} & 0.56 \\ 
60 & 0.70 & 0.61 & \textbf{0.71} & 0.66 & \textbf{0.72} & \textbf{0.75} & \textbf{0.82} & \textbf{0.80} & \textbf{0.77} & 0.55 \\ 
				\hline
			\end{tabular}}%
			\caption{ACC results of ``Carcinom" dataset.}%
		\end{minipage}%
\hfill
\begin{minipage}{0.49\textwidth}
	\centering
	\resizebox{\textwidth}{!}{%
		\begin{tabular}{|l||c||*{9}c|}
			\hline 
\#$f$ & 11340 & 11335 & 11301 & 10573 & 8238 & 7053 & 6697 & 6533 & 6180 & 4396 \\ \hline \hline
10 & 0.16 & \textbf{0.16} & 0.15 & \textbf{0.26} & \textbf{0.18} & \textbf{0.22} & \textbf{0.20} & \textbf{0.20} & \textbf{0.20} & \textbf{0.21} \\ 
15 & 0.14 & \textbf{0.14} & \textbf{0.15} & \textbf{0.26} & \textbf{0.18} & \textbf{0.28} & 0.09 & \textbf{0.24} & 0.07 & 0.06 \\ 
20 & 0.16 & \textbf{0.16} & 0.15 & 0.08 & 0.14 & \textbf{0.21} & 0.04 & \textbf{0.31} & \textbf{0.16} & 0.11 \\ 
25 & 0.14 & \textbf{0.14} & \textbf{0.15} & 0.09 & 0.08 & \textbf{0.22} & \textbf{0.23} & 0.10 & 0.09 & 0.11 \\ 
30 & 0.13 & \textbf{0.13} & \textbf{0.13} & 0.08 & 0.07 & \textbf{0.18} & 0.03 & \textbf{0.14} & 0.10 & 0.11 \\ 
35 & 0.17 & \textbf{0.17} & 0.13 & 0.03 & 0.07 & 0.12 & 0.10 & 0.01 & 0.08 & 0.10 \\ 
40 & 0.14 & \textbf{0.14} & \textbf{0.14} & 0.07 & 0.08 & 0.13 & 0.12 & 0.05 & \textbf{0.14} & 0.09 \\ 
45 & 0.09 & \textbf{0.09} & \textbf{0.18} & 0.08 & \textbf{0.11} & \textbf{0.10} & \textbf{0.13} & 0.07 & \textbf{0.12} & \textbf{0.09} \\ 
50 & 0.15 & 0.14 & \textbf{0.15} & 0.08 & 0.11 & 0.11 & 0.12 & 0.12 & 0.13 & 0.09 \\ 
55 & 0.15 & \textbf{0.15} & 0.14 & \textbf{0.21} & 0.08 & 0.13 & 0.13 & 0.12 & 0.13 & 0.07 \\ 
60 & 0.10 & \textbf{0.10} & \textbf{0.14} & \textbf{0.15} & 0.08 & \textbf{0.10} & \textbf{0.12} & \textbf{0.12} & \textbf{0.14} & 0.07 \\ 
			\hline
		\end{tabular}}%
		\caption{NMI results of ``CLL-SUB-111" dataset}%
	\end{minipage}%
	\hfill
	\begin{minipage}{0.49\textwidth}
		\centering
		\resizebox{\textwidth}{!}{%
			\begin{tabular}{|l||c||*{9}c|}
				\hline 
\#$f$ & 11340 & 11335 & 11301 & 10573 & 8238 & 7053 & 6697 & 6533 & 6180 & 4396  \\ \hline \hline
10 & 0.51 & \textbf{0.51} & 0.50 & \textbf{0.54} & \textbf{0.59} & \textbf{0.57} & \textbf{0.58} & \textbf{0.55} & \textbf{0.51} & 0.50 \\ 
15 & 0.51 & \textbf{0.51} & 0.50 & \textbf{0.57} & \textbf{0.55} & \textbf{0.62} & 0.47 & \textbf{0.59} & 0.45 & 0.43 \\ 
20 & 0.50 & \textbf{0.50} & 0.48 & 0.46 & \textbf{0.50} & \textbf{0.54} & 0.40 & \textbf{0.59} & \textbf{0.54} & \textbf{0.50} \\ 
25 & 0.48 & \textbf{0.48} & \textbf{0.51} & 0.44 & 0.46 & \textbf{0.54} & \textbf{0.57} & \textbf{0.50} & 0.46 & \textbf{0.50} \\ 
30 & 0.49 & \textbf{0.49} & \textbf{0.49} & 0.44 & 0.44 & \textbf{0.53} & 0.42 & \textbf{0.51} & 0.48 & 0.48 \\ 
35 & 0.51 & \textbf{0.51} & 0.49 & 0.42 & 0.44 & 0.49 & 0.49 & 0.41 & 0.44 & 0.48 \\ 
40 & 0.51 & \textbf{0.51} & 0.50 & 0.43 & 0.45 & 0.50 & 0.49 & 0.43 & 0.48 & 0.47 \\ 
45 & 0.46 & 0.45 & \textbf{0.52} & 0.44 & \textbf{0.46} & \textbf{0.47} & \textbf{0.51} & 0.45 & \textbf{0.47} & \textbf{0.47} \\ 
50 & 0.51 & 0.50 & \textbf{0.51} & 0.45 & 0.46 & 0.50 & 0.49 & 0.49 & 0.49 & 0.48 \\ 
55 & 0.49 & \textbf{0.49} & \textbf{0.50} & \textbf{0.54} & 0.46 & \textbf{0.50} & \textbf{0.50} & \textbf{0.49} & \textbf{0.49} & 0.45 \\ 
60 & 0.49 & \textbf{0.49} & \textbf{0.50} & \textbf{0.53} & 0.43 & 0.48 & \textbf{0.49} & \textbf{0.49} & \textbf{0.50} & 0.44 \\ 
				\hline
			\end{tabular}}%
			\caption{ACC results of ``CLL-SUB-111" dataset.}%
			\label{tab:tkdd_cll}
		\end{minipage}%
\end{table}

\subsection{Sparse Representation Errors}
\begin{figure}[ht!]
	\centering
	\includegraphics[width=0.225\textwidth]{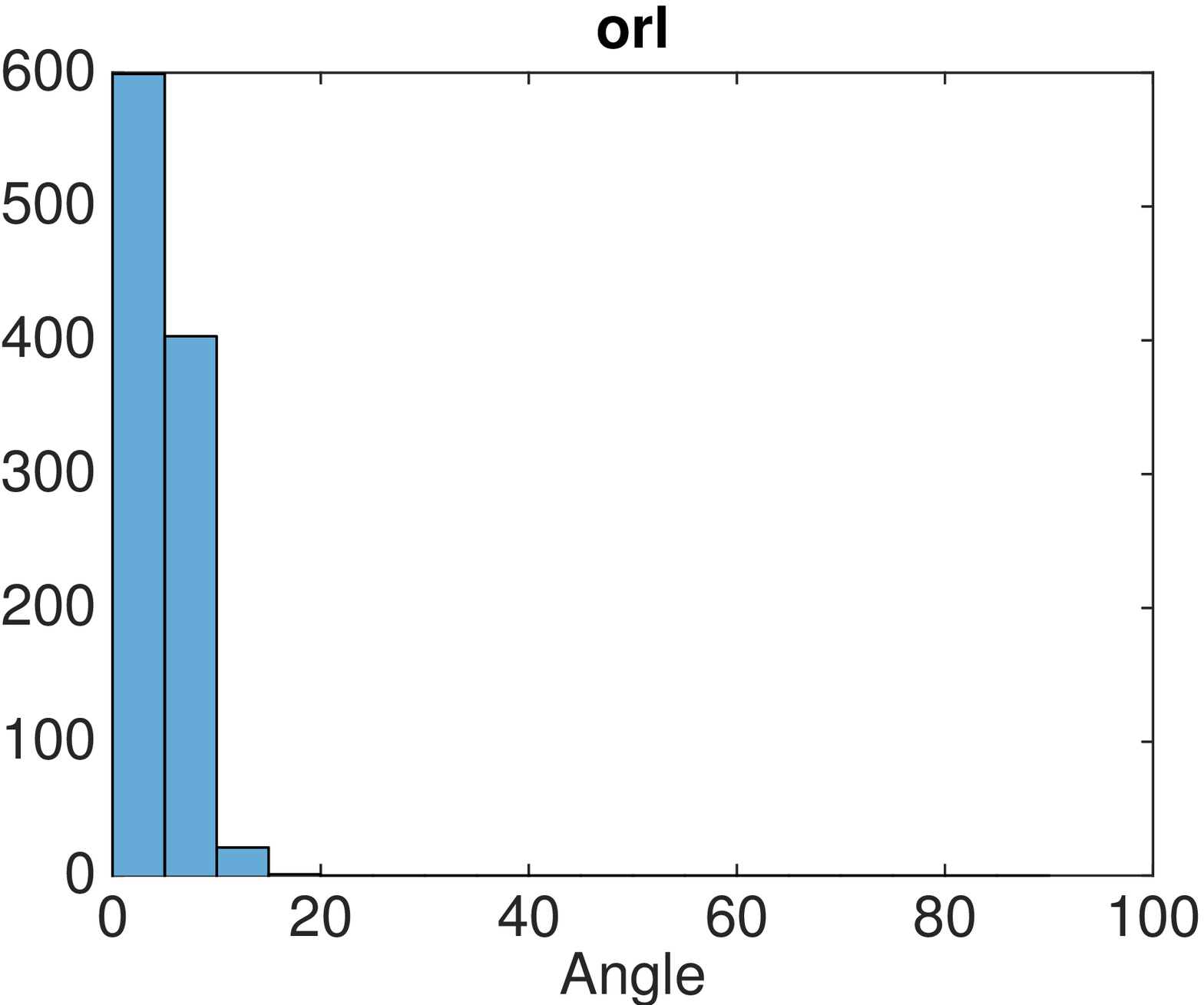} 	
	\includegraphics[width=0.225\textwidth]{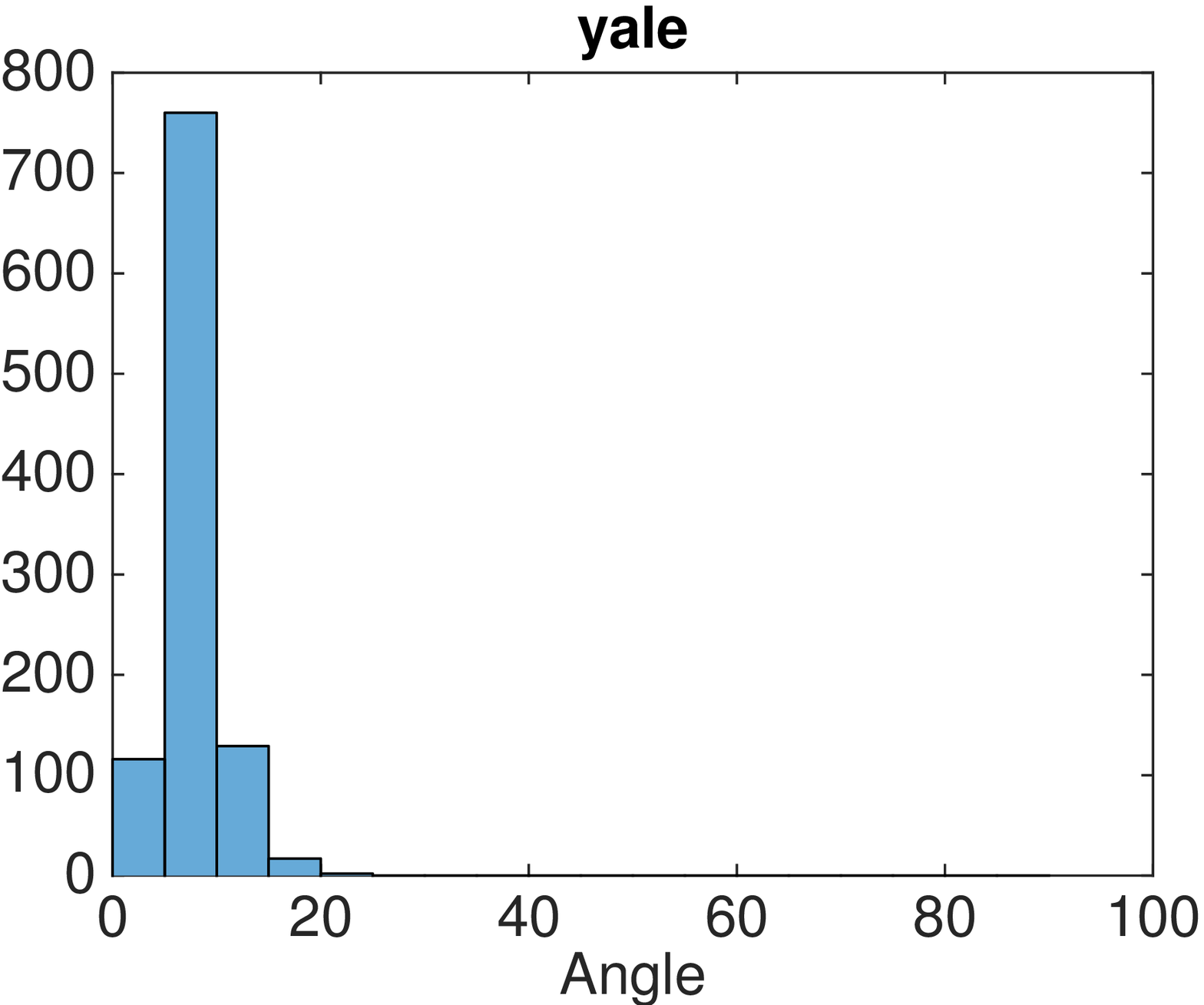} 
	\includegraphics[width=0.225\textwidth]{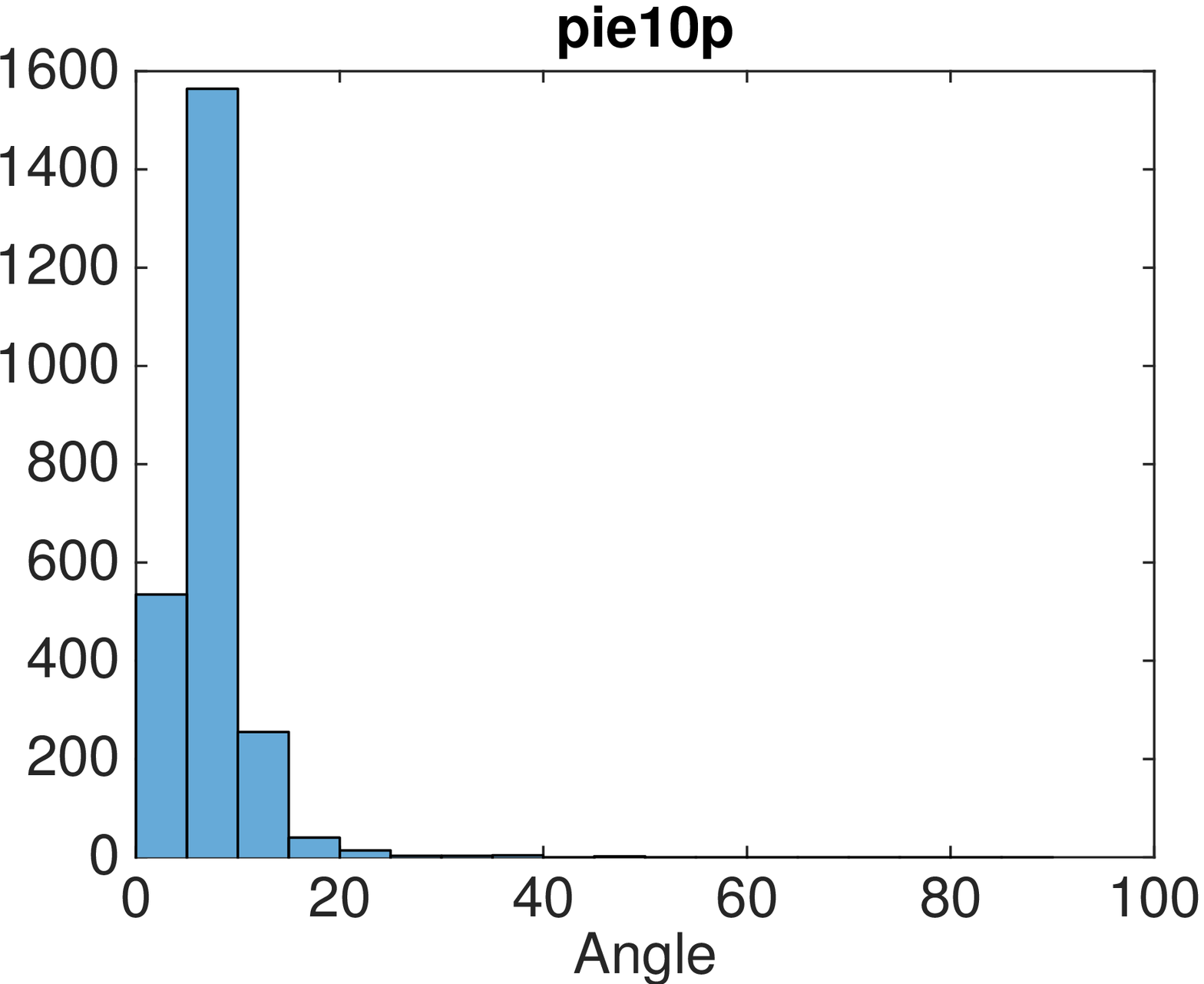}
	\includegraphics[width=0.225\textwidth]{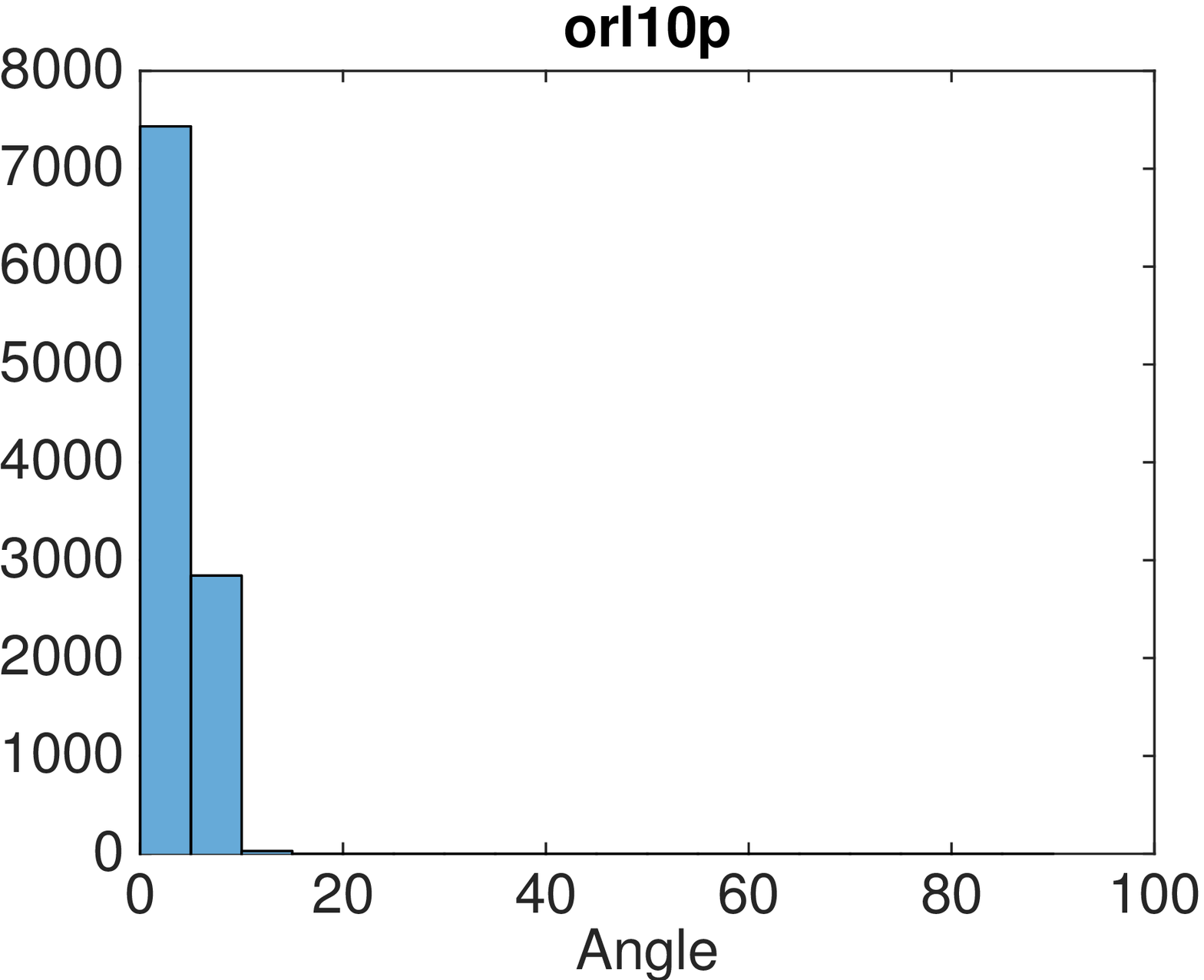} \\
	\includegraphics[width=0.225\textwidth]{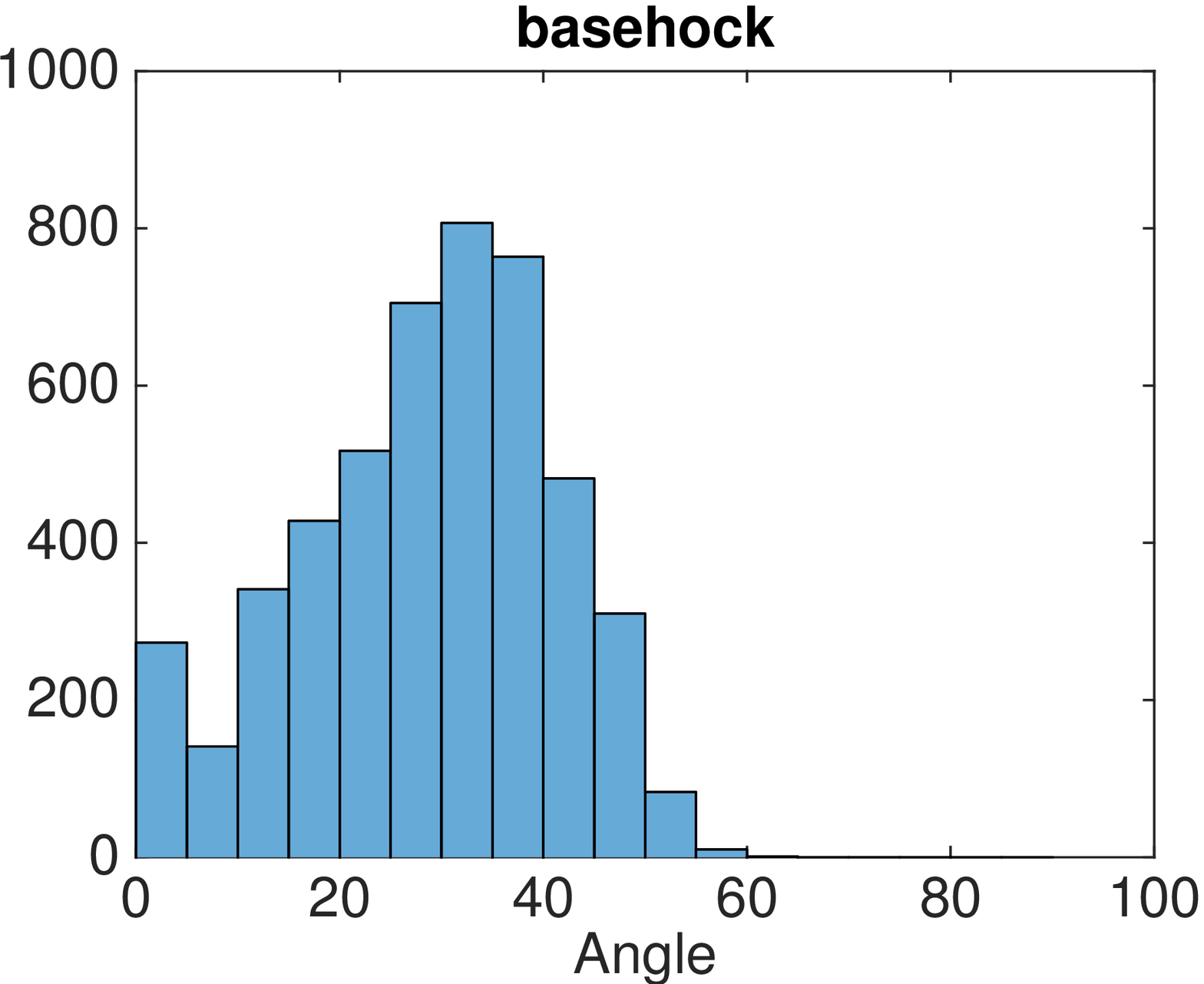} 	
	\includegraphics[width=0.225\textwidth]{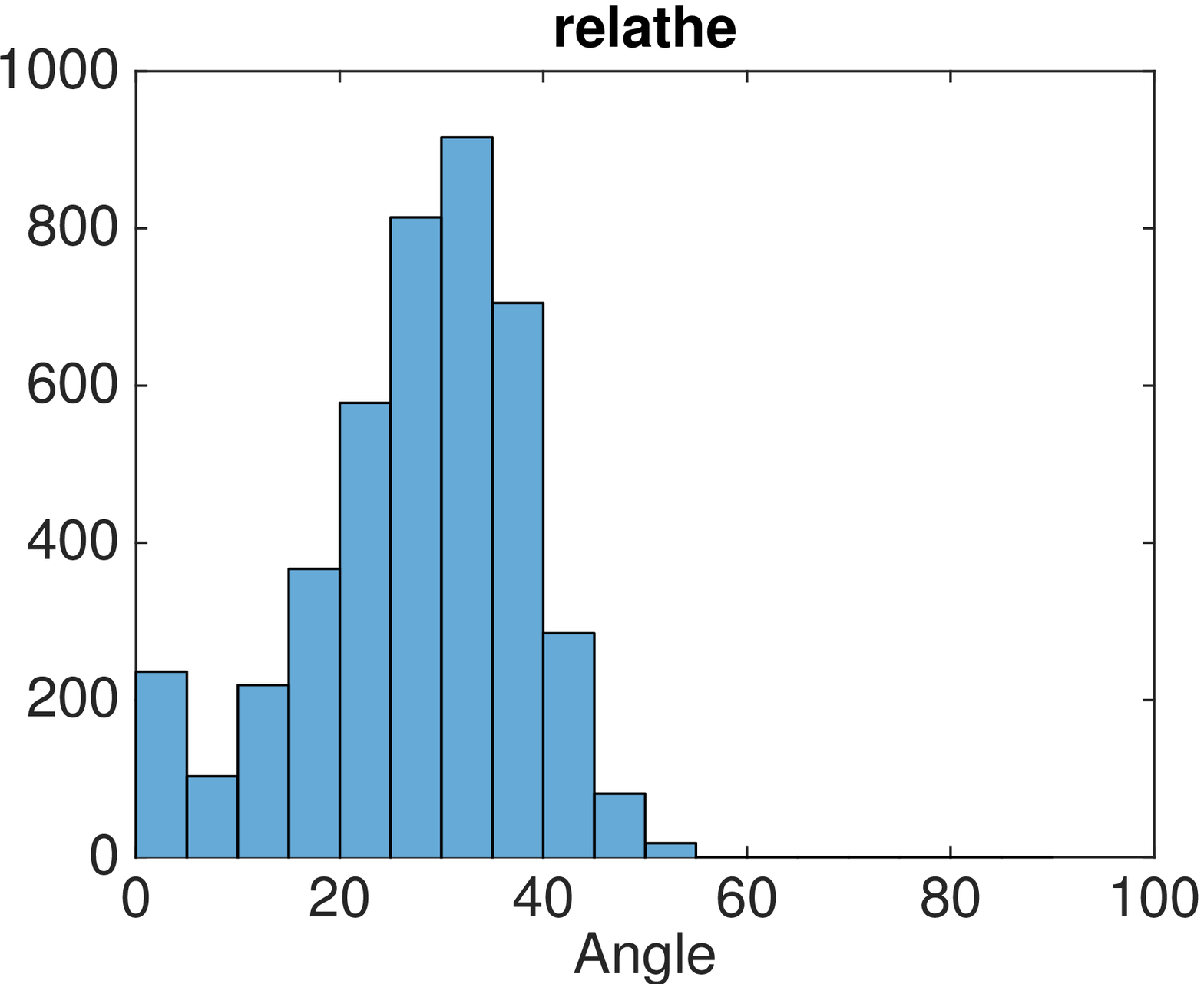} 
	\includegraphics[width=0.225\textwidth]{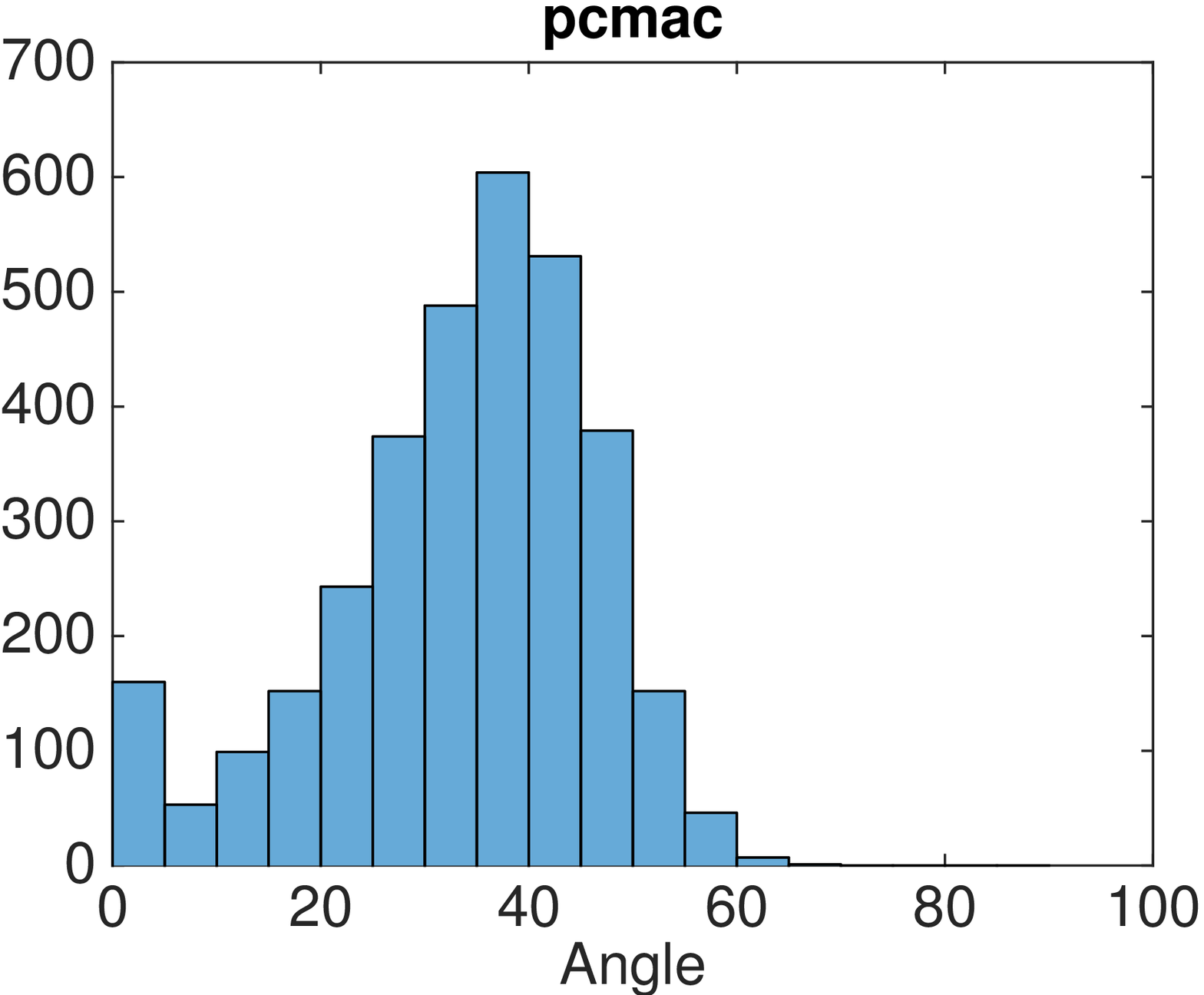}
	\includegraphics[width=0.225\textwidth]{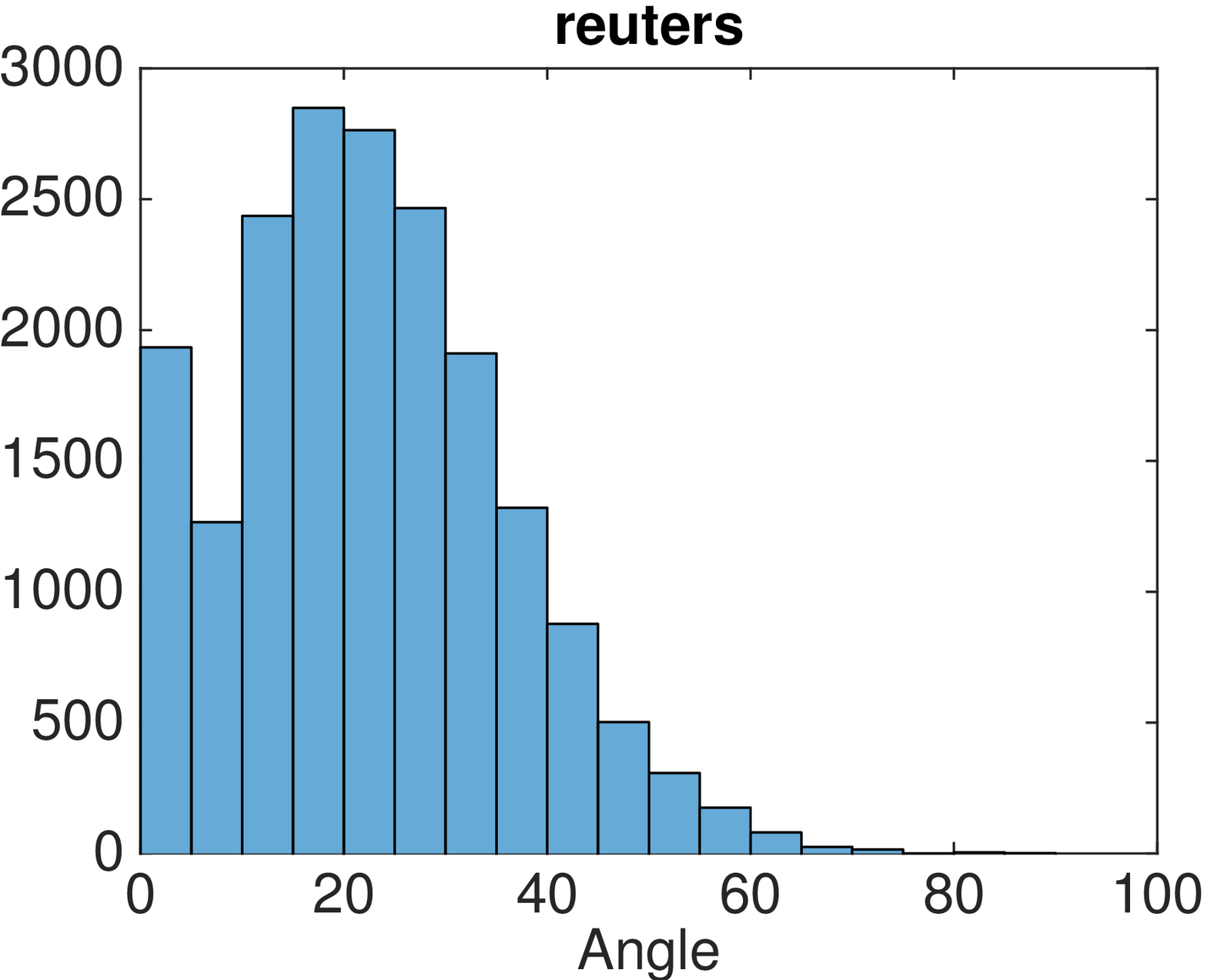} \\
	\includegraphics[width=0.225\textwidth]{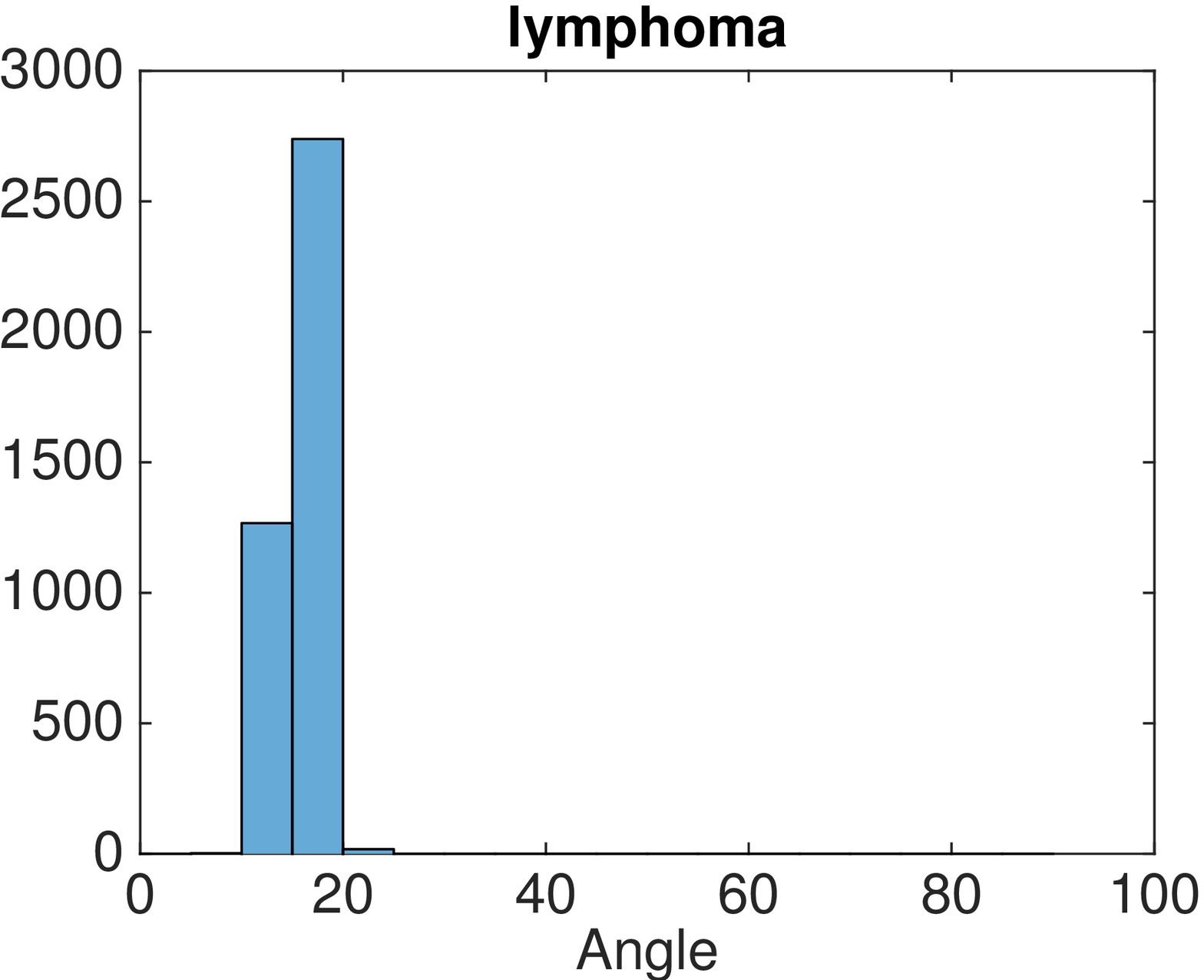} 	
	\includegraphics[width=0.225\textwidth]{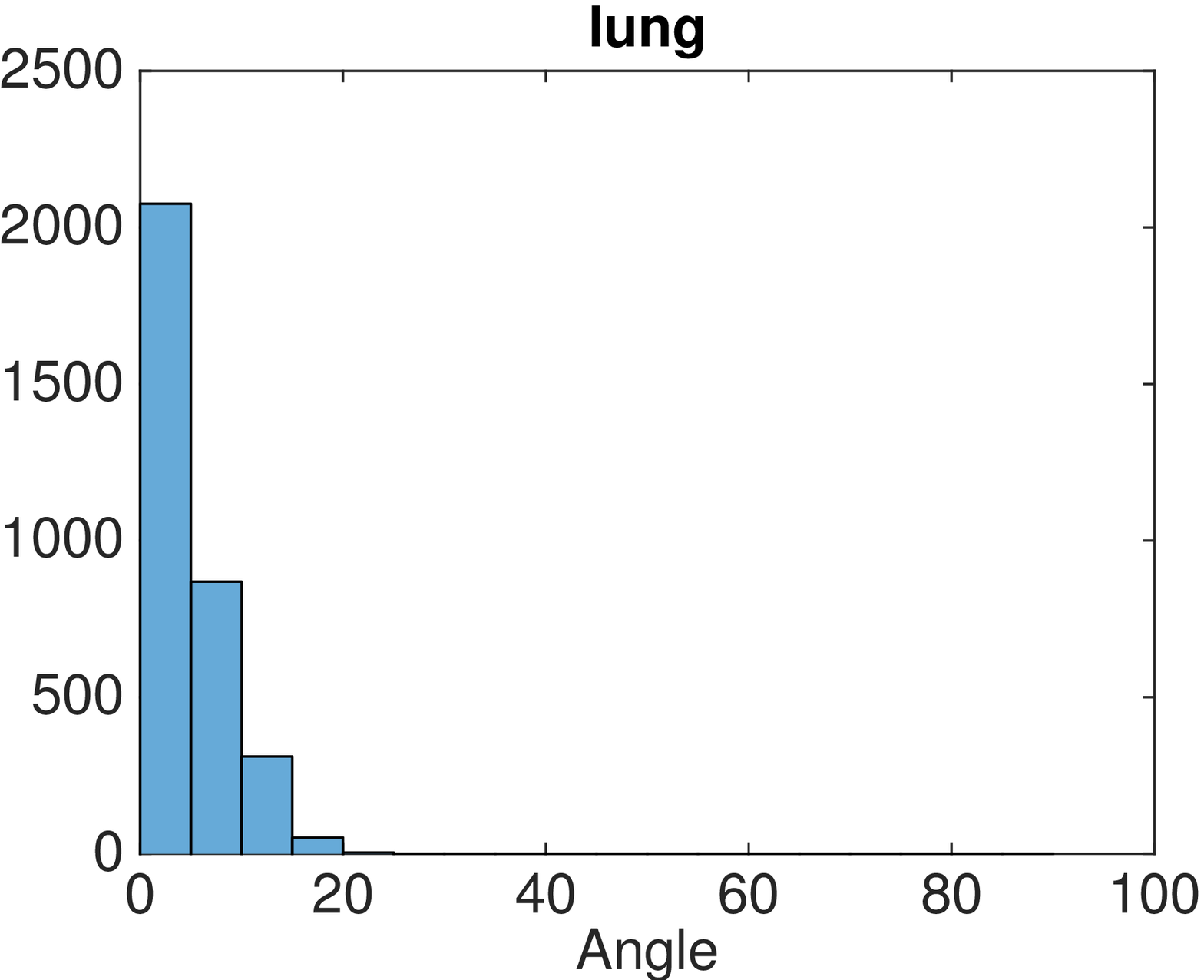} 
	\includegraphics[width=0.225\textwidth]{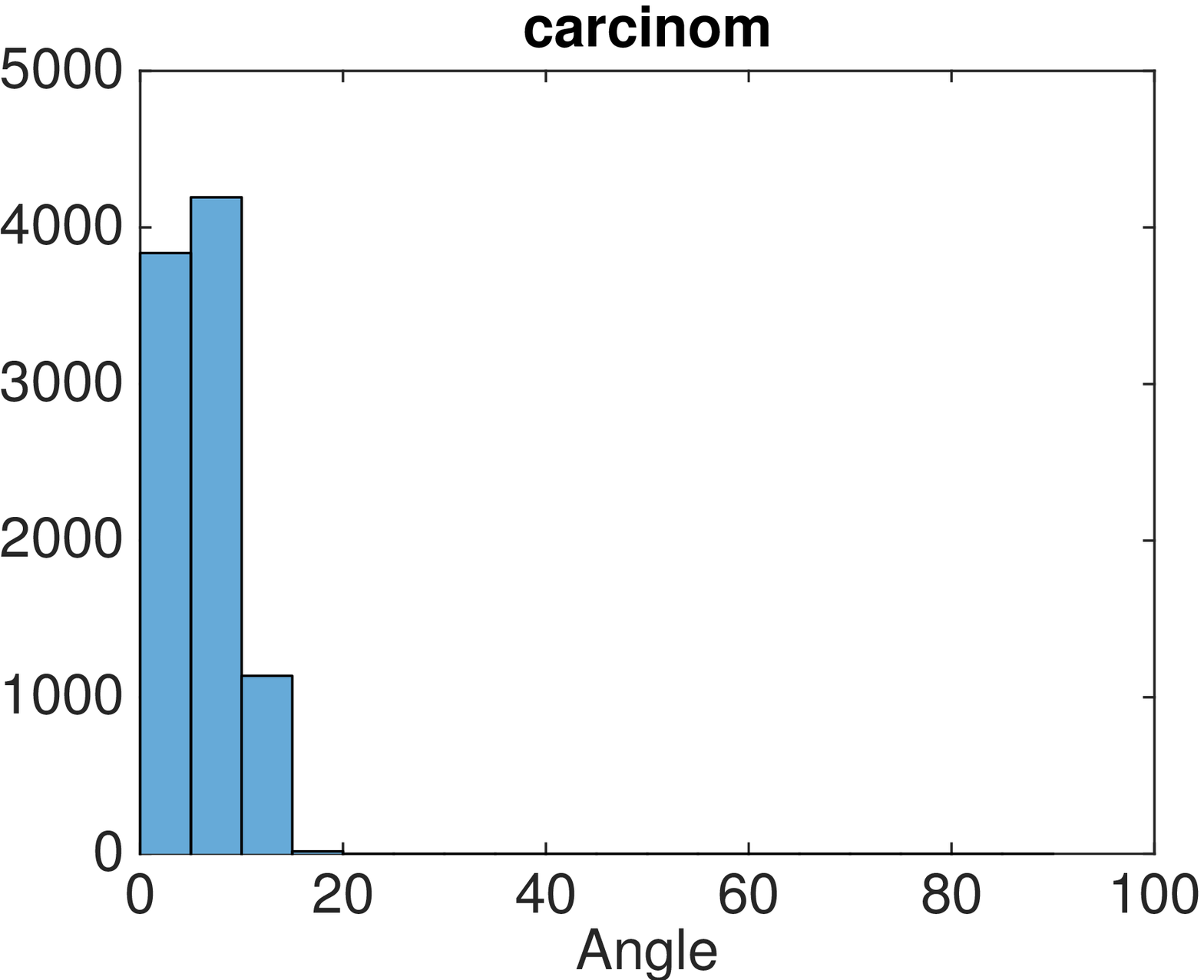}
	\includegraphics[width=0.225\textwidth]{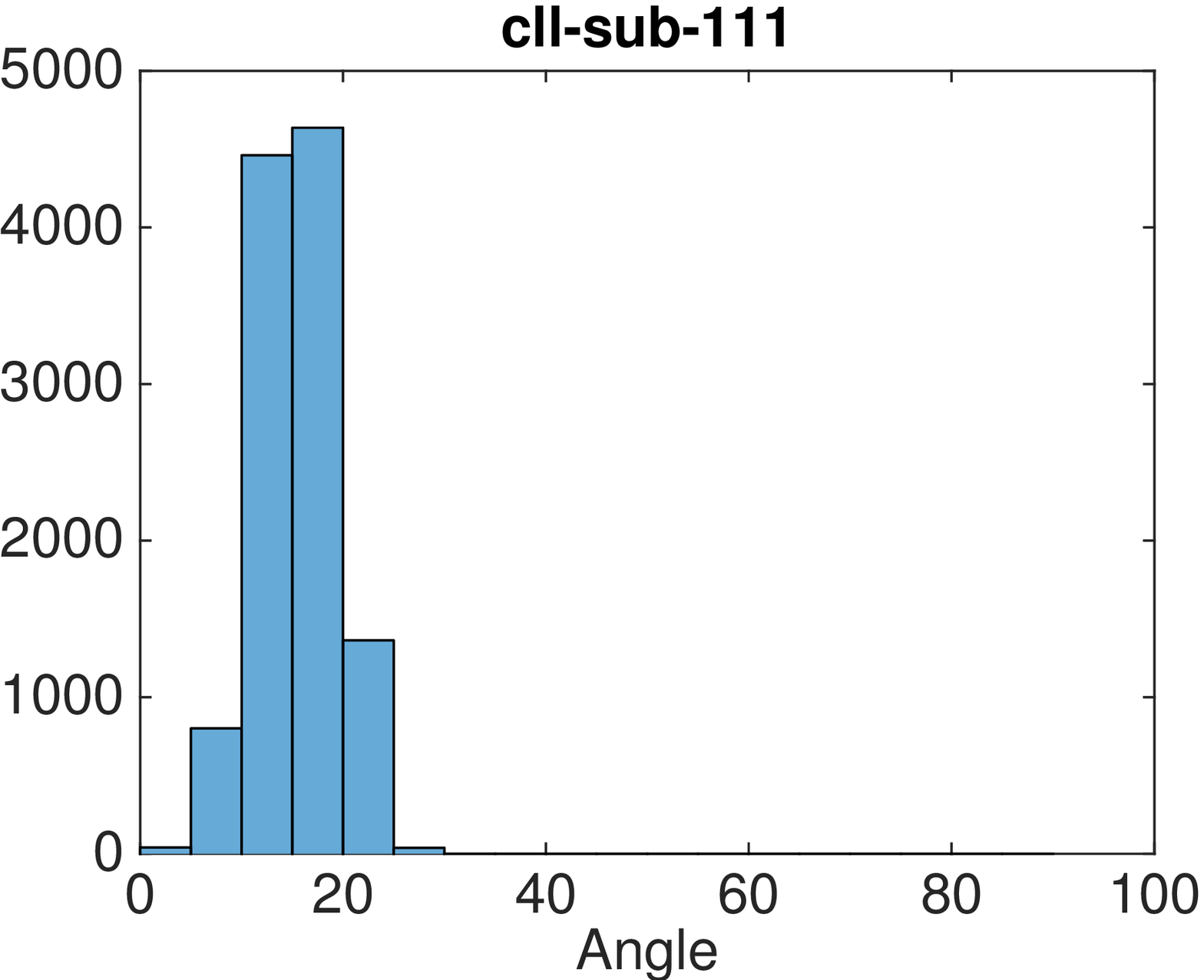} \\
	\caption{The distribution of angle between original feature vector and its sparse representation.}
	\label{fig:tkdd_sp_err}
\end{figure}

With the design of our modified OMP solvers, there will be failed/wrong sparse representations existing in generated sparse feature graph. The meaning of these edge connections and edge weights are invalid. And they should be removed from the SFG since wrong connections will deteriorate the accuracy of feature redundancy relationship. To validate the sparse representation, we check the angle between original feature vector and the linear weighted summation resulted vector (or recover signal from sparse coding point of view) from its sparse representation. If the angle lower than a threshold, we remove all out-edges from the generated sparse feature graph. To specify the threshold, we learn it from the empirical results of our selected twelve datasets. The distribution (or histogram) result of angle values is presented in figure~\ref{fig:tkdd_sp_err}.

\section{Related Works}
Remove redundant features is an important step for feature selection algorithms. Prestigious works include ~\cite{yu2004efficient}  which gives a formal definition of redundant features. Peng et al.~\cite{peng2005feature} propose a greedy algorithm (named as mRMR) to select features with minimum redundancy and maximum dependency. Zhao et al.~\cite{zhao2010efficient} develop an efficient spectral feature selection algorithm to minimize the redundancy within the selected feature subset through $\mathcal{L}_{2,1}$ norm. Recently, researchers pay attention to unsupervised feature selection with global minimized redundancy~\cite{xufeature}~\cite{wang2015feature}. Several graph based approaches are proposed in~\cite{mairal2013supervised},~\cite{song2013fast}. The most closed research work to us is~\cite{liu2012sparsity} which build a sparse graph at feature side and ranking features by approximation errors.
\section{Conclusion}
In this study, we propose sparse feature graph to model both one-to-one feature redundancy and one-to-many features redundancy. By separate whole features into different redundancy feature group through local compressible subgraphs, we reduce the dimensionality of data by only select one representative feature from each group. One advantage of our algorithm is that it does not need to calculate the pairwise distance which is always not accurate for high dimensional datasets. The experiment results shows that our algorithm is an effective way to obtain accurate data structure information which is demanding for unsupervised feature selection algorithms.

\bibliographystyle{plain}
\bibliography{BIB/RF.bib}

\end{document}